\pdfoutput=1

\documentclass[10pt,twocolumn,letterpaper]{article}

\usepackage[pagenumbers]{cvpr}   %

\usepackage{iftex}\ifXeTeX\usepackage[T1]{fontenc}\fi

\usepackage{currfile} %

\usepackage{caption} %

\usepackage{microtype}

\renewcommand{\paragraph}[1]{\vspace{.5em}\noindent\textbf{#1}}

\usepackage{makecell}

\newcommand{\vz}{\boldsymbol{z}}
\newcommand{\va}{\boldsymbol{a}}
\newcommand{\vsk}{\boldsymbol{s}}
\newcommand{\vq}{\boldsymbol{q}}
\newcommand{\vp}{\boldsymbol{p}}
\newcommand{\vF}{\boldsymbol{F}}
\newcommand{\vc}{\boldsymbol{c}}
\newcommand{\vP}{\boldsymbol{P}}
\newcommand{\vJ}{\boldsymbol{J}}
\newcommand{\vb}{\boldsymbol{b}}
\newcommand{\vr}{\boldsymbol{r}}
\newcommand{\vg}{\boldsymbol{g}}
\newcommand{\vS}{\boldsymbol{S}}
\newcommand{\vu}{\boldsymbol{u}}
\newcommand{\vx}{\boldsymbol{x}}
\newcommand{\vtheta}{\boldsymbol{\theta}}
\newcommand{\veps}{\boldsymbol{\varepsilon}}
\newcommand{\vDelta}{\boldsymbol{\Delta}}
\newcommand{\hvz}{\hat{\boldsymbol{z}}}
\newcommand{\hvp}{\hat{\boldsymbol{p}}}

\makeatletter
\def\ewm@b{b}\def\ewm@bx{bx}
\DeclareRobustCommand{\ewm}{\mbox{\edef\ewm@s{\f@series}\edef\ewm@f{\fontname\font}\@expandtwoargs\in@{\detokenize{ptmb}}{\ewm@f}%
  \ifin@\else\ifx\ewm@s\ewm@bx\in@true\else\ifx\ewm@s\ewm@b\in@true\fi\fi\fi
  \ifin@\ensuremath{\boldsymbol{\varepsilon}}\else\ensuremath{\varepsilon}\fi-WM}}
\makeatother
\AtBeginDocument{\ifdefined\pdfstringdefDisableCommands\pdfstringdefDisableCommands{\def\ewm{ε-WM}}\fi}

\usepackage{array} %
\makeatletter
\def\FloatBarrier{\par\begingroup \let\@elt\relax
 \edef\@tempa{\@botlist\@deferlist\@dbldeferlist}%
 \ifx\@tempa\@empty
 \else
  \ifx\@fltovf\relax
   \if@firstcolumn \clearpage
   \else \null\newpage\FloatBarrier
   \fi
  \else
   \newpage \let\@fltovf\relax
   \FloatBarrier
 \fi\fi \endgroup
 \suppressfloats[t]}
\makeatother
\newlength{\exmaxheight}
\newsavebox{\excapbox}

\definecolor{cvprblue}{rgb}{0.21,0.49,0.74}
\usepackage[pagebackref,breaklinks,colorlinks,allcolors=cvprblue]{hyperref}

\title{\ewm: Error-Driven Dynamics Adaptation\\across Fast--Slow Timescales for World Modeling}

\author{%
{\normalsize
Haoyi Zhang$^1$ \quad \quad Qi Zhang$^{1,2}$ \quad \quad Yunfan Ren$^{1,2}$ \quad \quad
Yi Bin$^1$ \quad \quad Lei Zhu$^{1,2}$ \quad \quad Heng Tao Shen}$^1$ \vspace{0.2cm}\\
$^{1}$Tongji University  \quad $^{2}$WorldMind\\
}

\makeatletter
\patchcmd{\@maketitle}{\vskip .375in}{\vskip 12pt}{}{\PackageError{main}{title-block patch 1 failed}{}}
\patchcmd{\@maketitle}{\vspace*{24pt}}{\vspace*{14pt}}{}{\PackageError{main}{title-block patch 2 failed}{}}
\patchcmd{\@maketitle}{\vspace*{12pt}}{\vspace*{1pt}}{}{\PackageError{main}{title-block patch 3 failed}{}}
\makeatother

\begin{document}
\maketitle
\begin{abstract}
A core challenge for world models is adapting to sudden, persistent, yet initially unknown changes in robot dynamics. If the world model fails to adapt, its prediction errors accumulate over the open-loop rollout and ultimately mislead the robot's actions. Existing world models, however, struggle to balance three key requirements for dynamics adaptation: responsiveness, rapidly correcting sudden changes; stability, avoiding lasting model updates from transient disturbances; and persistence, retaining adapted dynamics during the observation-free open-loop rollout. Inspired by multi-timescale predictive coding in the human brain, we propose \ewm{}, a world model that adapts to unknown dynamics through error-driven corrections across fast and slow timescales. 
\ewm{} incorporates three novel features: 1) Multiple internal experts (MiE) emulate neural pathway reweighting, each combining a fixed base expert with fast--slow adaptation networks. 2) Fast--slow timescale adaptation captures distinct adaptation dynamics: fast states respond rapidly and decay, whereas slow states accumulate corrections and persist. 3) Predictive correction uses prediction errors to update fast and slow states and reweight experts without gradients. During open-loop rollout, each expert feeds back its own predictions, and \ewm{} aggregates their outputs through adaptive weighting. Extensive experiments show \ewm{} has a consistently lower multi-step mean error than all ten baselines on six datasets, verifying the effectiveness of each feature.
\end{abstract}

\section{Introduction}
\label{sec:intro}

\begin{figure*}[t]
\centering
\includegraphics[width=0.867\textwidth,trim=0 1.9 0 0.6,clip]{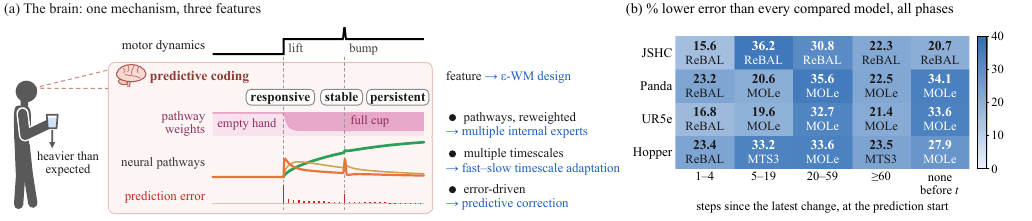}
\caption{\textbf{(a)}~Predictive coding (boxes: the three requirements). \textbf{(b)}~\ewm{}'s error reduction over the closest compared model (\cref{sec:accuracy}).}
\label{fig:teaser}
\vspace{-3mm}
\end{figure*}

In robotics and embodied AI, a world model predicts how a robot's state responds to actions, for planning and control~\cite{ha2018world,chua2018pets,hafner2025dreamerv3,hansen2024tdmpc2}. A core challenge is adapting predictions to sudden, persistent, yet initially unknown changes in robot dynamics during interaction~\cite{alegre2021mbcd,luo2022escp}. For instance, when a robotic arm grasps an object with an unknown payload or when a joint or engine fails, the robot dynamics change suddenly~\cite{nagabandi2019grbal,keipour2021alfa}. Such unknown dynamics require world models to adapt online through prediction errors~\cite{nagabandi2019grbal,nagabandi2019mole}. Without timely adaptation, prediction errors accumulate during open-loop rollouts, leading to misguided robot actions.

Intuitively, adapting to unknown dynamics imposes three fundamental requirements on world models. 1)~Responsiveness: Sudden dynamics changes demand rapid adaptation, as prediction errors can accumulate when model-generated states are recursively fed back during open-loop rollouts~\cite{venkatraman2015improving,janner2019mbpo}. 2)~Stability: Adaptation must distinguish persistent changes from transient disturbances, such as impulses or impacts, which may initially appear similar but should not induce lasting model updates~\cite{widmer1996learning,gama2014survey}. 3)~Persistence: Since dynamics changes may persist without further observations during open-loop rollouts, adapted dynamics must be retained to prevent subsequent predictions from reverting to outdated dynamics. Obviously, these requirements introduce an inherent single-timescale trade-off: faster adaptation increases sensitivity to transient disturbances~\cite{widrow1976stationary,haykin2002adaptive}, whereas slower adaptation suppresses such disturbances at the cost of retaining outdated dynamics or neglecting persistent dynamics changes throughout open-loop rollouts.

However, existing world models for robotics can be broadly categorized into three classes, each satisfying only a subset of these requirements. General-purpose world models~\cite{hafner2025dreamerv3,hansen2024tdmpc2} support diverse tasks but lack explicit mechanisms for online dynamics adaptation, making their multi-step predictions vulnerable to persistent dynamics changes. Online-adaptive world models~\cite{nagabandi2019grbal,nagabandi2019mole,alegre2021mbcd} update their dynamics estimates using recent observations, but typically operate at a single adaptation timescale. Consequently, rapidly adapting methods, such as ReBAL, offer responsiveness but struggle to retain adapted dynamics throughout open-loop rollouts, whereas methods with more persistent adaptation, such as MOLe, respond less promptly to sudden changes (\cref{sec:why}). Multi-timescale world models~\cite{shaj2023mts3} introduce slowly evolving latent variables to capture persistent dynamics. However, their interval-based inference delays adaptation to sudden changes until the next update. These limitations highlight the need for a world model that simultaneously achieves responsiveness, stability, and persistence in adapting to unknown dynamics.

The human brain offers a promising framework through the mechanism of multi-timescale predictive coding~\cite{rao1999predictive,friston2010free,keller2018predictive}. It maintains multiple internal `models' specialized for different conditions~\cite{wolpert1998multiple}, with adaptation driven by discrepancies between predicted and observed sensory outcomes~\cite{wolpert1995internal,tseng2007sensory,shadmehr2010error}. \Cref{fig:teaser}a illustrates this mechanism: lifting an unexpectedly heavy cup induces a persistent change in dynamics, whereas an incidental bump introduces a transient disturbance. First, neural pathways are reweighted by prediction errors~\cite{wolpert1998multiple,haruno2001mosaic}, shifting responsibility from the empty hand to the full cup while preserving the former intact. Second, adaptation proceeds across neural pathways with multiple timescales~\cite{smith2006interacting,kording2007dynamics}, ranging from fast-decaying (orange) to slow-persistent (green) responses. Thus, the reflexive response to the bump decays, while that to the extra weight persists. Third, neural adaptation is error‑driven and therefore begins as soon as a prediction fails. %

\begin{figure*}[t]
  \centering
  \includegraphics[width=\textwidth]{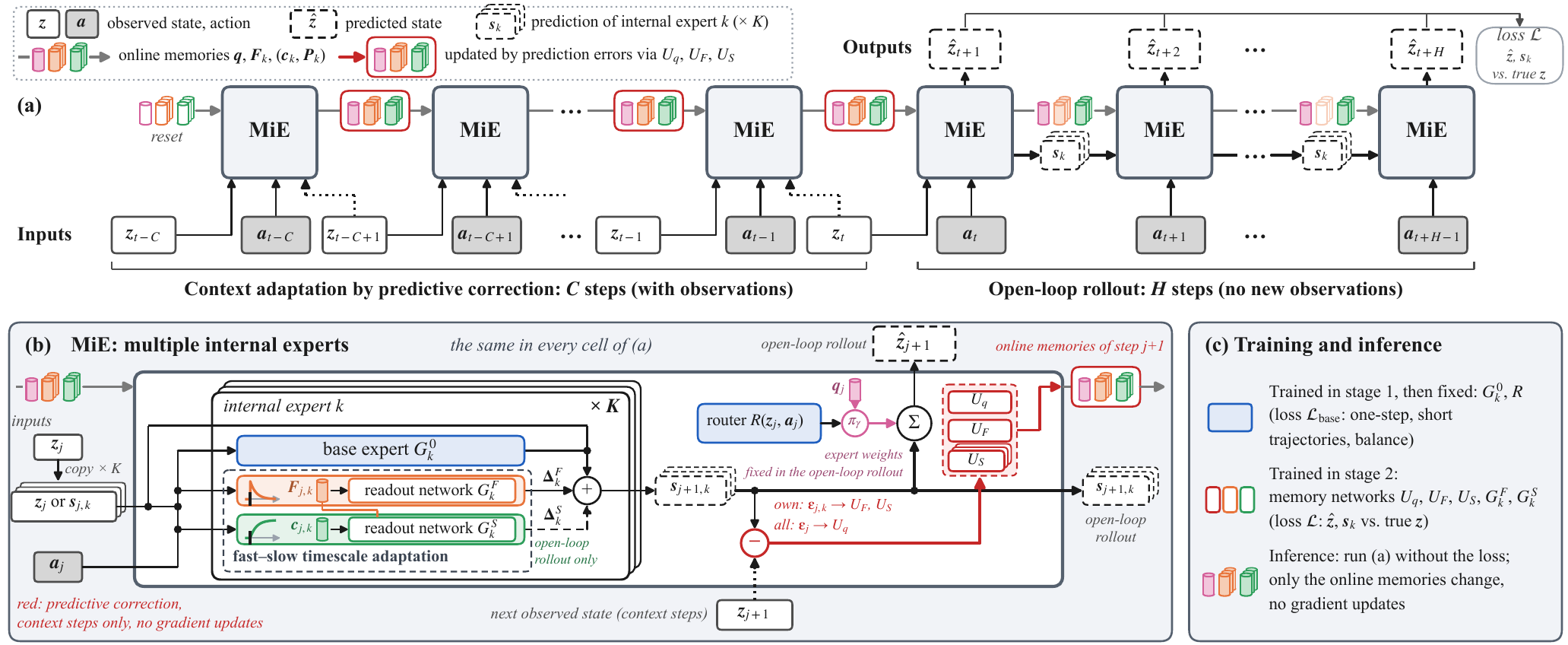}
  \caption{\ewm{} architecture. \textbf{(a)}~The same MiE runs at every step. In context adaptation, all prediction errors update the routing memory $\vq$, and then each internal expert's own prediction error updates its fast state $\vF_k$ and slow state $\vc_k$. In the open-loop rollout, each internal expert feeds back its own prediction. \textbf{(b)}~One MiE at step $j$ (notation as in \cref{eq:onestep}), enlarged, with the same inputs and outputs as each cell of (a). The observed state $\vz_j$ is copied to all $K$ internal experts, and each internal expert uses its own $\vsk_{j,k}$ in the open-loop rollout. $G^F_k$ and $G^S_k$ read out the fast and slow corrections $\vDelta^F_k$ and $\vDelta^S_k$ from the current $\vF_{j,k}$ and $\vc_{j,k}$. Red marks predictive correction, which compares each prediction $\vsk_{j+1,k}$ with the next observed state $\vz_{j+1}$ (dotted arrows: context steps only). Each internal expert's own prediction error $\veps_{j,k}$ updates its fast and slow states through $U_F$ and $U_S$, and all prediction errors $\veps_j$ update $\vq$ through $U_q$, giving the online memories of step $j{+}1$. The small curves sketch the decaying fast state and the persisting slow state. \textbf{(c)}~Training and inference (supp.\ Secs.~\hyperref[sec:supp-method]{B.2} and \hyperref[sec:supp-stage1]{B.3}).}
  \label{fig:framework}
  \vspace{-3mm}
\end{figure*}

Inspired by the mechanism, we propose \ewm{} (\cref{fig:framework}), a world model that adapts to unknown dynamics through error-driven corrections across fast and slow timescales (\cref{fig:teaser}a). First, \textbf{multiple internal experts} (\textbf{MiE}) emulate neural pathway reweighting through a mixture of experts~\cite{jacobs1991adaptive}. A router assigns expert weights based on the current state and action, refined by routing memory. Each internal expert comprises a fixed base expert and fast–slow adaptation networks. Second, \textbf{fast–slow timescale adaptation} corrects base predictions with fast and slow states: a fast state that decays at a fixed rate and a slow state that accumulates prediction errors through a Kalman-like gain and has no decay factor. Third, \textbf{predictive correction} drives error-driven adaptation: each expert's prediction error updates its fast and slow states, while the errors across experts update routing memory to favor those aligned with the new dynamics. During open-loop rollout, the experts predict future states autoregressively, with their outputs aggregated through adaptive weighting. At inference, only the fast and slow states and the routing memory are updated, without gradients.
Extensive experiments on six simulation and real-world datasets demonstrate that \ewm{} outperforms ten baselines in multi-step prediction. It consistently achieves higher accuracy across all dynamics-change phases in simulation (\cref{fig:teaser}b), and \cref{sec:mechanism,sec:ablation} confirm each design's contribution. Our key contributions are:

\begin{itemize}
\item We introduce a predictive coding-inspired world model \ewm{} that adapts to unknown dynamics through prediction errors across fast and slow timescales.
\item We design multiple internal experts with fast–slow timescale adaptation and predictive correction, enabling error-driven corrections aligned with unknown dynamics.
\item Extensive experiments demonstrate \ewm{}'s superior multi-step prediction accuracy, reducing errors by \mbox{16\%--34\%} over the strongest baseline in simulation. In-depth analyses further reveal its responsiveness, stability, and persistence under dynamics changes.
\end{itemize}

\section{Related Work}
\label{sec:related}

We review world models for robot dynamics in the three classes of \cref{sec:intro} and contrast each with \ewm{}.

\noindent\textbf{General-purpose world models} serve online planning or policy learning in imagined trajectories~\cite{ha2018world,hafner2019planet,hafner2025dreamerv3,hansen2024tdmpc2,chua2018pets,janner2019mbpo}. Some of these models, such as TD-MPC2 and PETS, predict only from the current state and therefore cannot detect a change. Recurrent models, such as DreamerV3 and RKN~\cite{becker2019rkn}, update a latent state with new observations (RKN via Kalman updates) but lack explicit mechanisms for online dynamics adaptation. In contrast, \ewm{} adds fast and slow corrections, driven by prediction errors, to fixed base experts.

\noindent\textbf{Online-adaptive world models} adapt to recent observations in four main ways. First, GrBAL and MOLe update parameters by gradient steps, and MOLe also adds models for new situations~\cite{nagabandi2019grbal,nagabandi2019mole}. Second, ReBAL, CaDM and HiP-RSSM~\cite{nagabandi2019grbal,lee2020cadm,shaj2022hiprssm} encode recent history into a context variable for the current dynamics, as do adaptive policies~\cite{yu2017uposi,luo2022escp,kumar2021rma}. Third, MBCD and GPMM~\cite{alegre2021mbcd,xu2020gpmm} detect a change or infer the current dynamics before switching models. Fourth, some methods fit only the last linear layer online with Kalman-style or least-squares rules, as in adaptive control~\cite{oconnell2022neuralfly,ingebrand2024zeroshot,levy2025metalearning}. More recent methods apply low-rank second-order weight updates~\cite{altawaitan2026adapting}, reweight world models by prototype similarity~\cite{jang2026tmow} or correct predictions with one feedback state~\cite{an2026feedback}. However, these methods typically adapt at a single timescale, set by a step size, history length, detection threshold, filter noise or feedback gain, and therefore face the trade-off between the three requirements (\cref{sec:intro}). In contrast, the fast--slow timescale adaptation of \ewm{} feeds prediction errors to both a fast state that decays and a slow state that persists. Moreover, transient disturbances never alter \ewm{}'s fixed base experts.

\noindent\textbf{Multi-timescale world models} maintain fast and slow latent states. Clockwork VAE updates higher layers more slowly for video prediction~\cite{saxena2021clockwork}, and THICK updates part of its state only sparsely for planning~\cite{gumbsch2024thick}. MTS3~\cite{shaj2023mts3} updates a fast latent state at every step. However, MTS3 infers its slow variable only at fixed intervals from an observation window, not from prediction errors. MTS3's slow variable therefore reflects outdated dynamics after a change until the next update. In contrast, our predictive correction feeds each prediction error to the slow state at each context step.

\section{Method}
\label{sec:method}

\ewm{} (\cref{fig:framework}) consists of multiple internal experts (MiE, \cref{sec:base}) with fast--slow timescale adaptation (\cref{sec:memory}) and predictive correction (\cref{sec:context}). The same MiE runs at every context step and at every step of the open-loop rollout (\cref{sec:rollout} and \cref{fig:framework}a). Training has two stages, and inference uses no gradient updates (\cref{sec:train} and \cref{fig:framework}c).

\subsection{Problem setting}
\label{sec:problem}

The next state depends on the state ${\vz_t\in\mathbb{R}^D}$ (\eg, joint angles and velocities), the action ${\va_t\in\mathbb{R}^A}$ and unobserved physical parameters $\vtheta_t$ (\eg, payload, friction and motor strength), as given by the robot dynamics ${\vz_{t+1}\sim p(\cdot\mid\vz_t,\va_t;\vtheta_t)}$.
A \emph{change} assigns $\vtheta_t$ a new, persistent value at an unknown time, so only recent transitions reveal it. A \emph{transient disturbance}, such as a pulse or an impact, also alters the dynamics but soon vanishes. The world model $f$ receives the \emph{context}~\cite{lee2020cadm,shaj2022hiprssm} at time $t$, \ie, the $C$ transitions up to $\vz_t$, and the actions of the \emph{horizon}, \ie, the next $H$ steps. The world model then predicts the states of the horizon in an \emph{open-loop rollout}, \ie, without new observations~\cite{hafner2019planet}:
\begin{equation*}
  \hvz_{t+1:t+H}=f\big(\{(\vz_j,\va_j,\vz_{j+1})\}_{j=t-C}^{t-1},\,\va_{t:t+H-1}\big),
\end{equation*}
with $C{+}1$ observed states and $C{+}H$ actions as inputs and $H$ predicted states as outputs. Adapting to unknown dynamics requires a low error at every predicted step, even immediately after a change, so that a planner can compare action sequences~\cite{chua2018pets}.

\subsection{Multiple internal experts}
\label{sec:base}

MiE (\cref{fig:framework}b) follows the mixture of experts~\cite{jacobs1991adaptive}, but each of its $K$ internal experts has its own feedforward and feedback path. Each internal expert predicts, compares its prediction with the next observed state and is corrected by its own prediction error. In contrast, a standard mixture only maps inputs to outputs and weights its experts by the input alone.

Following \cref{fig:framework}b from left to right, the observed state $\vz_j$ is copied to all $K$ internal experts in context adaptation (copy~$\times K$). In the open-loop rollout, each internal expert~$k$ instead takes its own prediction $\vsk_{j,k}$ (\cref{sec:rollout}). All internal experts receive the same action $\va_j$. These inputs enter base expert~$k$, an MLP ${G^0_k:\mathbb{R}^{D}\times\mathbb{R}^{A}\to\mathbb{R}^D}$ with three hidden layers of 256 units, trained in stage~1 and then fixed (\cref{sec:train}). Since the state and the action are low-dimensional vectors, base expert~$k$ takes their concatenation without an encoder. Base expert~$k$ predicts the state increment, ${\vz_{j+1}\approx\vz_j+G^0_k(\vz_j,\va_j)}$, so the state input is added to this increment (the top line into ``$+$'' in \cref{fig:framework}b). \emph{Internal expert}~$k$ then adds two corrections to this prediction. The first is the fast correction ${\vDelta^F_k\in\mathbb{R}^D}$ of its own \textbf{fast state} ${\vF_k\in\mathbb{R}^{d_F}}$. The second is the slow correction ${\vDelta^S_k\in\mathbb{R}^D}$ of its own \textbf{slow state} ${\vc_k\in\mathbb{R}^{d_c}}$ (\cref{sec:memory}), which is added in the open-loop rollout only. Hence, the prediction errors in the context are not reduced by the slow correction, and the slow state continues to accumulate them (\cref{sec:memory,sec:context}). The sum gives the prediction ${\vsk_{j+1,k}\in\mathbb{R}^D}$ of internal expert~$k$.

Finally, a router ${R:\mathbb{R}^{D}\times\mathbb{R}^{A}\to\mathbb{R}^K}$, an MLP with two hidden layers of 128 units, scores the internal experts. Like the base experts, the router is trained in stage~1 and then fixed. The router receives no prediction errors; its scores may thus not reflect a change. The \emph{routing memory}, a probability vector ${\vq\in\mathbb{R}^K}$ updated by the prediction errors (\cref{sec:context}), corrects these scores, giving the \emph{expert weights}~$\pi_\gamma$
\begin{equation}
  \pi_\gamma(\vz,\va;\vq)=\mathrm{softmax}\big(\gamma\,R(\vz,\va)+\alpha\log\vq\big)\in\mathbb{R}^K,
  \label{eq:router}
\end{equation}
where $\alpha$ sets the influence of $\vq$ and $\gamma>0$ is an inverse routing temperature. A smaller $\gamma$ makes the expert weights more uniform. In the open-loop rollout, MiE outputs the weighted sum of the internal experts' predictions ($\Sigma$ in \cref{fig:framework}b). In context adaptation, the prediction errors instead first update the routing memory, then the fast and slow states.

\subsection{Fast--slow timescale adaptation}
\label{sec:memory}

The fast and slow states of each internal expert, which reduce the multiple timescales of \cref{fig:teaser}a to two, are latent vectors updated at every context step from its prediction error ${\veps_{j,k}\in\mathbb{R}^D}$ (\cref{eq:onestep}). The fast state responds rapidly and decays by the factor $\rho=0.8$ at every step, a half-life of about 3 steps (supp.\ Sec.~\hyperref[sec:supp-hparams]{B.1}). In contrast, the slow state has no decay factor: it accumulates over several steps and persists. This difference serves stability: a transient disturbance should mainly move the decaying fast state, whereas a persistent change should also move the slow state. Readout networks $G^F_k$ and $G^S_k$, MLPs with two hidden layers, map the two states to the fast and slow corrections (orange and green rows of \cref{fig:framework}b),
\begin{equation*}
  \begin{aligned}
    \vDelta^F_k(\vz,\va;\vF_k)&=G^F_k(\vz,\va,\vF_k)-G^F_k(\vz,\va,\mathbf 0),\\
    \vDelta^S_k(\vz,\va;\vF_k,\vc_k)&=G^S_k(\vz,\va,\vF_k,\vc_k)-G^S_k(\vz,\va,\vF_k,\mathbf 0),
  \end{aligned}
\end{equation*}
so zero states yield zero corrections. The fast state of internal expert~$k$ is updated as
\begin{equation}
  \begin{aligned}
    \vF_{j+1,k}={}\rho\vF_{j,k}+(1-\rho)\tanh\big(U_F(\vu_{j,k})-U_F(\vu^0_{j,k})\big),
  \end{aligned}
  \label{eq:fast-update}
\end{equation}
where one MLP $U_F$ with two hidden layers of 128 units, shared by all internal experts, maps $\vu_{j,k}$ to $\mathbb{R}^{d_F}$. The input $\vu_{j,k}$ contains $\vz_j$, $\va_j$, $\vF_{j,k}$, $\veps_{j,k}$, the expert weight $p_{j,k}$ and the updated routing memory $q_{j+1,k}$ of \cref{eq:route-update}, and $\vu^0_{j,k}$ is $\vu_{j,k}$ with zero prediction error.

The slow-state update $U_S$ of each internal expert treats each prediction error as a noisy measurement of the slow state.\ $U_S$ consists of a measurement network $U^{\mathrm m}_S$ (an MLP with two hidden layers), which outputs the sensitivity ${\vJ_k\in\mathbb{R}^{D\times d_c}}$ and the bias ${\vb_k\in\mathbb{R}^D}$ of a linear measurement model, ${\veps_{j,k}\approx\vJ_k\vc_{j,k}+\vb_k}$, and a filter network $U^{\mathrm f}_S$ (one hidden layer), which outputs the measurement noise ${\vr_k\in\mathbb{R}^{d_c}}$. The variance ${\vP_{j,k}\in\mathbb{R}^{d_c}}$ measures the uncertainty of the slow state. With element-wise $\odot$ and $\oslash$, and with $j$ omitted from $\vJ_k$, $\vb_k$, $\vr_k$, $\vS_k$ and $\vg_k$, the slow state and its variance are updated as follows:
\begin{equation}
  \begin{gathered}
    (\vJ_k,\vb_k)=U^{\mathrm m}_S(\vz_j,\va_j,\vF_{j,k}),\\
    \vr_k=U^{\mathrm f}_S(\vz_j,\va_j,\vF_{j,k}),\\
  \begin{aligned}
    \vS_k&=\vP_{j,k}\odot\operatorname{diag}(\vJ_k^{\top}\vJ_k)+\vr_k,\\
    \vg_k&=\vP_{j,k}\oslash \vS_k,\\
    \vc_{j+1,k}&=\tanh\big(\vc_{j,k}+\vg_k\odot\tfrac1D \vJ_k^{\top}(\veps_{j,k}-\vb_k)\big),\\
    \vP_{j+1,k}&=\big(1-\tfrac1D\vg_k\odot\operatorname{diag}(\vJ_k^{\top}\vJ_k)\big)\odot\vP_{j,k}+\sigma_P^2 .
  \end{aligned}
  \end{gathered}
  \label{eq:slow-update}
\end{equation}
Here ${\vS_k\in\mathbb{R}^{d_c}}$ is the total variance of the prediction error per slow-state component, and the gain ${\vg_k\in\mathbb{R}^{d_c}}$ weights a new prediction error. The gain grows with the uncertainty and shrinks with the noise $\vr_k$, so noisy prediction errors move the slow state less. The process noise $\sigma_P^2$ keeps the uncertainty above zero, allowing the slow state to follow a new change. Since the update does not subtract the explained part $\vJ_k\vc_{j,k}$, a persistent prediction error accumulates in the slow state, whereas a transient disturbance is brief and adds little to it. This update and stage-2 training (\cref{sec:train}) distinguish the two, as \cref{sec:why} verifies.

\subsection{Predictive correction}
\label{sec:context}
Predictive correction (red in \cref{fig:framework}b) updates the \emph{online memories} from the context, one transition $j$ at a time, without gradient updates. The online memories are the routing memory and the fast state, slow state, and variance of each internal expert. Before each context, these memories are reset to ${\vq=\tfrac1K\mathbf 1}$, ${\vF_k=\vc_k=\mathbf 0}$ and the initial variance ${\vP_k=\sigma_0^2\mathbf 1}$, so \ewm{} infers a change from the context alone. At step~$j$, each internal expert starts from $\vz_j$ and the memories $\vq_j$, $\vF_{j,k}$, $\vc_{j,k}$ and $\vP_{j,k}$, and compares its prediction with the next observed state $\vz_{j+1}$ (dotted arrows in \cref{fig:framework}):
\begin{equation}
  \begin{aligned}
  \vsk_{j+1,k}&=\vz_j+G^0_k(\vz_j,\va_j)+\vDelta^F_k(\vz_j,\va_j;\vF_{j,k}),\\
    \veps_{j,k}&=\vz_{j+1}-\vsk_{j+1,k}.
  \end{aligned}
  \label{eq:onestep}
\end{equation}
This one-step prediction uses only the fast correction.

The prediction errors update the online memories in two steps. First, all prediction errors, ${\veps_j=(\veps_{j,1},\dots,\veps_{j,K})}$, update the routing memory. The current expert weights ${\vp_j=\pi_\gamma(\vz_j,\va_j;\vq_j)}$ are computed before the prediction errors arrive and then adjusted as
\begin{equation}
  \vq_{j+1}=\mathrm{softmax}\big(\log\vp_j+\tanh(U_q(\vx_j)-U_q(\vx^0_j))\big),
  \label{eq:route-update}
\end{equation}
where an MLP $U_q$ with two hidden layers of 128 units maps $\vx_j$ to $\mathbb{R}^K$. The input $\vx_j$ contains $\vz_j$, $\va_j$, $\vp_j$, $\veps_j$ and the fast states $\vF_{j,k}$ of all internal experts, and $\vx^0_j$ is $\vx_j$ with ${\veps_j=\mathbf 0}$. Second, each internal expert's own prediction error $\veps_{j,k}$ updates its fast and slow states by \cref{eq:fast-update,eq:slow-update}. The updated online memories are used at the next step (\cref{fig:framework}b). With zero prediction errors, the fast states only decay and ${\vq_{j+1}=\vp_j}$. The prediction errors thus reweight and correct the internal experts but never alter their fixed base experts.

\subsection{Open-loop rollout}
\label{sec:rollout}
The open-loop rollout starts from the online memories $\vq_t$, $\vF_{t,k}$ and $\vc_{t,k}$ after the context. The routing memory $\vq_t$ and the slow states $\vc_{t,k}$ remain fixed, because no new observations arrive. Likewise, the expert weights ${\hvp=\pi_\gamma(\vz_t,\va_t;\vq_t)}$ are computed once, and the fast states only decay, ${\vF_{t+h+1,k}=\rho\vF_{t+h,k}}$. Each internal expert~$k$ starts from ${\vsk_{t,k}=\vz_t}$ and feeds back its own prediction for $h=0,\dots,H-1$:
\begin{equation}
\vsk_{t+h+1,k}=\vsk_{t+h,k}+G^0_k(\vsk_{t+h,k},\va_{t+h})+\vDelta^F_k+\vDelta^S_k,
  \label{eq:rollout}
\end{equation}
where the corrections use $(\vsk_{t+h,k},\va_{t+h})$, $\vF_{t+h,k}$ and $\vc_{t,k}$. The prediction of \ewm{} is the weighted sum $\hvz_{t+h+1}=\sum_k\hat p_k\,\vsk_{t+h+1,k}$. As the fast state decays and the slow state stays fixed, the fast correction should dominate the early steps and the slow correction the late steps (\cref{sec:mechanism}).

\subsection{Training}
\label{sec:train}

\paragraph{Stage 1: training the base experts and the router.}
The base experts and the router are trained jointly on short training segments by minimizing
\begin{equation}
  \mathcal L_{\text{base}}=\mathcal L_{\text{step}}+\lambda_b\,\mathcal L_{\text{traj}}+\lambda_u\big(\mathcal L_{\text{bal}}+\mathcal L_{\text{use}}\big).
  \label{eq:stage1}
\end{equation}
Here $\lambda_b$ and $\lambda_u$ are loss weights. The terms $\mathcal L_{\text{step}}$ and $\mathcal L_{\text{traj}}$ are mixture-Gaussian negative log-likelihoods of the one-step predictions and of the open-loop trajectory of each base expert over the segment. The balance terms $\mathcal L_{\text{bal}}$ and $\mathcal L_{\text{use}}$ keep all base experts in use (supp.\ Sec.~\hyperref[sec:supp-stage1]{B.3}).

\paragraph{Stage 2: training the memory networks.}
We train the memory networks ($U_q$ and the fast--slow adaptation networks $U_F$, $U_S$, $G^F_k$ and $G^S_k$) as in \cref{fig:framework}a. On each sample, \ewm{} resets the online memories, reads $C$ context steps and runs an $H$-step open-loop rollout. Gradients flow through the whole context, enabling the model to infer changes from prediction errors. The loss is computed only over the $H$ steps of the horizon:
\begin{equation}
  \begin{aligned}
    \mathcal L&=\tfrac{1}{|\mathcal H|}\textstyle\sum_{n\in\mathcal H}\big(\ell_n+\lambda\,\mathrm{MSE}_n\big),\\
    \ell_n&=-\tfrac{1}{nD}\log\textstyle\sum_{k}\hat p_k\,e^{-E_{n,k}},\\
    E_{n,k}&=\tfrac{1}{2\sigma^2}\textstyle\sum_{h=1}^{n}\lVert \vsk_{t+h,k}-\vz_{t+h}\rVert^2,
  \end{aligned}
  \label{eq:loss}
\end{equation}
where $\mathcal H$ is the horizon set of the main metric (\cref{sec:setup}), $\lambda$ is a loss weight and $\sigma$ is a fixed standard deviation. The term $\ell_n$ keeps whole trajectories of highly weighted internal experts close to the truth. The weighted mean squared error $\mathrm{MSE}_n$ of $\hvz$ over steps 1 to $n$ (supp.\ Sec.~\hyperref[sec:supp-method]{B.2}) aligns the loss with the evaluation metric. At inference, \ewm{} updates only the online memories, without gradients.

\section{Experiments}
\label{sec:exp}

\begin{table*}[t]
\centering
\renewcommand{\arraystretch}{0.94}
\setlength{\tabcolsep}{2.6pt}
\resizebox{\textwidth}{!}{%
\begin{tabular}{lcccccccccc}
\toprule
Horizon: 40 steps & \multicolumn{2}{c}{JSHC} & \multicolumn{2}{c}{ESCP-HC} & \multicolumn{2}{c}{Panda} & \multicolumn{2}{c}{UR5e} & \multicolumn{2}{c}{Hopper} \\
\cmidrule(lr){2-3} \cmidrule(lr){4-5} \cmidrule(lr){6-7} \cmidrule(lr){8-9} \cmidrule(lr){10-11}
Model & multi-step mean & step 40 & multi-step mean & step 40 & multi-step mean & step 40 & multi-step mean & step 40 & multi-step mean & step 40 \\
\midrule
DreamerV3-style RSSM & 0.197$\pm$0.003 & 0.311$\pm$0.004 & 0.372$\pm$0.006 & 0.532$\pm$0.009 & 0.329$\pm$0.016 & 0.404$\pm$0.022 & 0.331$\pm$0.011 & 0.418$\pm$0.013 & 0.145$\pm$0.003 & 0.224$\pm$0.009 \\
TD-MPC2 & 0.192$\pm$0.002 & 0.370$\pm$0.012 & 0.284$\pm$0.008 & 0.599$\pm$0.033 & 0.189$\pm$0.013 & 0.411$\pm$0.026 & 0.226$\pm$0.014 & 0.547$\pm$0.064 & 0.102$\pm$0.004 & 0.204$\pm$0.010 \\
TD-MPC2-40 & 0.212$\pm$0.002 & \underline{0.279$\pm$0.009} & 0.332$\pm$0.002 & 0.499$\pm$0.012 & 0.244$\pm$0.007 & 0.409$\pm$0.010 & 0.253$\pm$0.010 & 0.450$\pm$0.022 & 0.111$\pm$0.002 & 0.219$\pm$0.013 \\
MLP & 0.195$\pm$0.001 & 0.306$\pm$0.009 & 0.279$\pm$0.002 & 0.486$\pm$0.009 & 0.0841$\pm$0.0024 & \underline{0.130$\pm$0.007} & 0.0846$\pm$0.0035 & \underline{0.116$\pm$0.003} & 0.0953$\pm$0.0011 & 0.177$\pm$0.004 \\
\midrule
GrBAL & 0.210$\pm$0.006 & 0.402$\pm$0.021 & 0.352$\pm$0.087 & 2.119$\pm$3.354 & 0.0889$\pm$0.0040 & 0.176$\pm$0.009 & 0.0836$\pm$0.0034 & 0.145$\pm$0.006 & 0.0817$\pm$0.0040 & 0.203$\pm$0.032 \\
ReBAL & \underline{0.166$\pm$0.007} & 0.341$\pm$0.008 & \underline{0.249$\pm$0.004} & 0.596$\pm$0.029 & 0.0728$\pm$0.0017 & 0.162$\pm$0.011 & \underline{0.0641$\pm$0.0035} & 0.130$\pm$0.018 & 0.0867$\pm$0.0126 & 0.247$\pm$0.029 \\
MOLe & 0.204$\pm$0.010 & 0.726$\pm$0.723 & 0.320$\pm$0.021 & 1.222$\pm$0.607 & \underline{0.0711$\pm$0.0013} & 0.149$\pm$0.007 & 0.0665$\pm$0.0012 & 0.123$\pm$0.005 & \underline{0.0752$\pm$0.0024} & 0.178$\pm$0.012 \\
MBCD & 0.245$\pm$0.006 & 0.388$\pm$0.022 & 0.365$\pm$0.024 & 0.641$\pm$0.034 & 0.318$\pm$0.412 & 6.801$\pm$14.710 & 1.951$\pm$3.957 & 172.477$\pm$384.954 & 0.111$\pm$0.021 & 0.434$\pm$0.480 \\
GPMM & 0.395$\pm$0.077 & 1.944$\pm$0.928 & 0.496$\pm$0.021 & 2.065$\pm$0.153 & 0.216$\pm$0.103 & 1.815$\pm$1.153 & 0.124$\pm$0.071 & 0.717$\pm$0.945 & 0.535$\pm$0.396 & 5.595$\pm$4.130 \\
\midrule
MTS3 & 0.208$\pm$0.001 & 0.290$\pm$0.004 & 0.274$\pm$0.004 & \underline{0.441$\pm$0.002} & 0.126$\pm$0.003 & 0.190$\pm$0.011 & 0.0861$\pm$0.0031 & 0.133$\pm$0.006 & 0.0796$\pm$0.0020 & \underline{0.157$\pm$0.004} \\
\midrule
\ewm{} (ours) & \textbf{0.124$\pm$0.001} & \textbf{0.266$\pm$0.008} & \textbf{0.208$\pm$0.003} & \textbf{0.428$\pm$0.008} & \textbf{0.0473$\pm$0.0007} & \textbf{0.0956$\pm$0.0021} & \textbf{0.0435$\pm$0.0014} & \textbf{0.0753$\pm$0.0037} & \textbf{0.0499$\pm$0.0027} & \textbf{0.124$\pm$0.003} \\
\ewm{} vs.\ strongest & \multicolumn{2}{c}{$-$25.7\% vs.\ ReBAL (5/5)} & \multicolumn{2}{c}{$-$16.4\% vs.\ ReBAL (5/5)} & \multicolumn{2}{c}{$-$33.5\% vs.\ MOLe (5/5)} & \multicolumn{2}{c}{$-$32.1\% vs.\ ReBAL (5/5)} & \multicolumn{2}{c}{$-$33.7\% vs.\ MOLe (5/5)} \\
\bottomrule
\end{tabular}}
\caption{Multi-step mean (horizons $n=1,5,10,20,40$) and step-40 errors (nMSE) on five simulated datasets, as mean $\pm$ standard deviation over 5 seeds. Rows follow the classes of \cref{sec:related}. Best in bold, second best underlined. One MBCD seed diverges on Panda and UR5e.}
\label{tab:main}
\end{table*}

\begin{figure*}[t]
\centering
\includegraphics[width=0.92\textwidth,trim=0 0 0 4.3,clip]{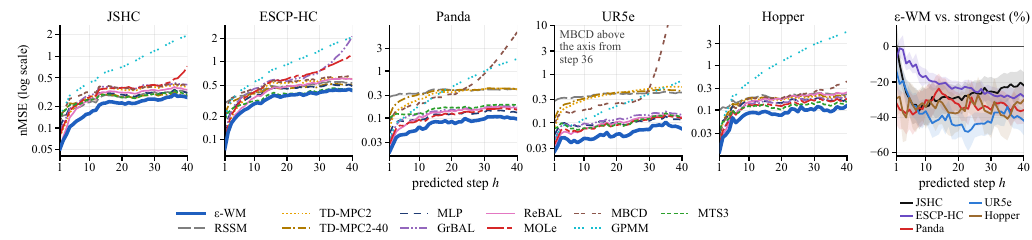}
\caption{Per-step error on the five simulated datasets: all models (left) and \ewm{} relative to the strongest compared model (right).}
\label{fig:perstep}
\vspace{-3mm}
\end{figure*}

\subsection{Experimental setup}
\label{sec:setup}

\noindent\textbf{Datasets.} Six datasets: five simulations in MuJoCo and one of real flights (supp.\ Sec.~\hyperref[sec:supp-a]{A}). Two \textbf{established datasets} follow prior change protocols. Joint-Switch HalfCheetah (JSHC) disables a random leg joint every 100 steps, as in \mbox{GrBAL} and MBCD~\cite{nagabandi2019grbal,alegre2021mbcd}, and \mbox{ESCP-HC} scales gravity, joint damping and torso mass~\cite{zhang2024dora} by new random factors every 100 steps, as in ESCP~\cite{luo2022escp}. Three \textbf{new datasets} add robot arms and transient disturbances, which these two lack. Built on standard robot models~\cite{todorov2012mujoco,towers2024gymnasium,zakka2022menagerie}, these datasets change the payload, joint friction and motor strength of the Panda and UR5e arms and the torso mass, ground friction and motor strength of Hopper, with pulses and impacts in between. The \textbf{real data} are public fixed-wing UAV flights with engine or control-surface faults from ALFA~\cite{keipour2021alfa}.

\noindent\textbf{Compared models and implementation.} We compare \ewm{} with ten baselines (\emph{compared models}) from the three classes of \cref{sec:related} (supp.\ Sec.~\hyperref[sec:supp-baselines]{B.4}). The \textbf{general-purpose} models are a DreamerV3-style recurrent state-space model (RSSM)~\cite{hafner2025dreamerv3}, TD-MPC2~\cite{hansen2024tdmpc2}, TD-MPC2-40 (trained with a 40-step loss) and an MLP. The \textbf{online-adaptive} models are GrBAL and ReBAL~\cite{nagabandi2019grbal}, MOLe~\cite{nagabandi2019mole}, MBCD~\cite{alegre2021mbcd} and GPMM~\cite{xu2020gpmm}. The \textbf{multi-timescale} model is MTS3~\cite{shaj2023mts3}. All neural models have about 0.70M parameters (GPMM is non-parametric), so model size cannot explain the differences. All models share the same training and test pipeline, data splits, context of $C=64$ steps and horizon of $H=40$ steps (2\,s). All settings, including $K=3$ internal experts for \ewm{}, are chosen without the test data (supp.\ Sec.~\hyperref[sec:supp-hparams]{B.1}).

\noindent\textbf{Task and metrics.} Each model receives the same context and the recorded actions of the 40 predicted steps, and predicts only the 40 states. Errors are measured on states rather than images, as in HiP-RSSM and MTS3~\cite{shaj2022hiprssm,shaj2023mts3}. The \emph{per-step error} (nMSE) is the mean squared error at one predicted step, with each state dimension standardized on the training set. Our main metric, the \emph{multi-step mean error}, averages the per-step error over steps 1 to $n$ and then over $n\in\mathcal H=\{1,5,10,20,40\}$, so that both short and long horizons contribute. The \emph{step-40 error} is the per-step error at the last step. Simulation results are means over 5 seeds (independent training runs), and shaded bands and error bars show 95\% bootstrap intervals (supp.\ Sec.~\hyperref[sec:supp-metrics]{C.1}).

\subsection{Prediction accuracy}
\label{sec:accuracy}

\begin{figure*}[t]
\centering
\captionsetup{skip=7pt}
\includegraphics[width=0.95\textwidth]{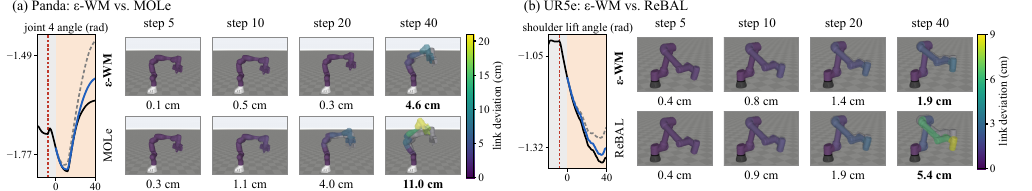}
\caption{Joint angles predicted by \ewm{} (blue) and the strongest compared model (gray, dashed) starting 8 steps after a change (red dashed line), with the truth in black. Arm poses overlay the true pose (gray), and each link is colored by its link deviation (distance from the truth), whose mean is given below each pose. Background: context (gray) and prediction (light orange).}
\label{fig:render}
\par\vspace{8pt}
\includegraphics[width=0.86\textwidth]{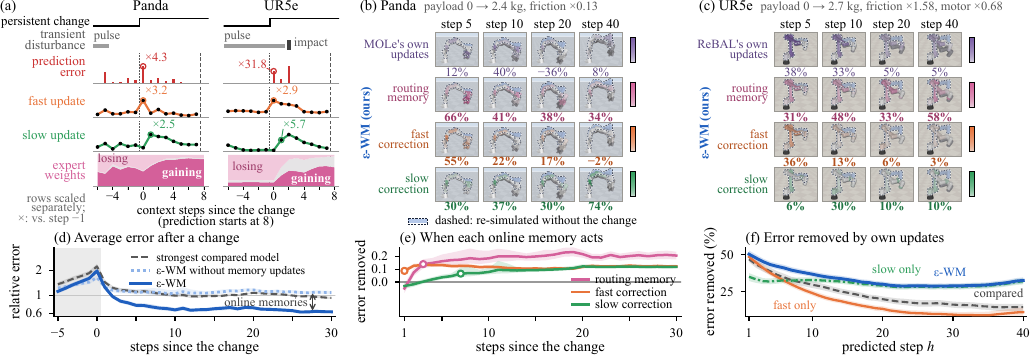}
\caption{Why \ewm{} leads after a change. (a--c)~Two examples. (d--f)~Means over 1,586 persistent changes (4 seeds on UR5e).}
\label{fig:mechviz}
\end{figure*}

\noindent\textbf{Main results.} On all five simulated datasets, \ewm{} has the lowest multi-step mean and step-40 errors (\cref{tab:main}). Its multi-step mean error is lower than that of every compared model in all 5 seeds, and the 95\% intervals of all 50 differences lie below zero. The \emph{strongest compared model} on a dataset is the compared model with the lowest multi-step mean error (ReBAL or MOLe, both online-adaptive). In the last row of \cref{tab:main}, \ewm{}'s multi-step mean error is 16.4\%--33.7\% below this model's (5/5: in all 5 seeds).

\noindent\textbf{Per-step errors.} \Cref{fig:perstep} (left, logarithmic scale) shows that \ewm{} has the lowest per-step error at every step on every simulated dataset except step 1 on JSHC, so its lead holds over almost the whole horizon. Relative to the strongest compared model (right, where negative values favor \ewm{}), all intervals lie below zero from step 4 onward, and at step 40, \ewm{}'s per-step error is still 21.9\%--41.9\% lower.

\begin{figure*}[t]
\centering
\captionsetup{skip=7pt}
\includegraphics[width=0.9\textwidth,trim=0 0 0 7.3,clip]{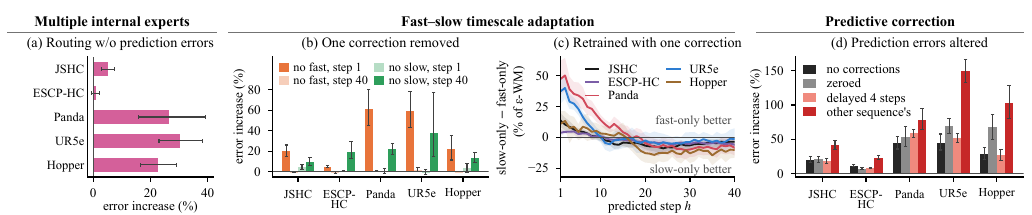}
\caption{Tests of the three designs: (a, b, d) alterations at inference, (c) retrained models. On UR5e, (a) and (b) use 4 seeds (supp.\ Sec.~\hyperref[sec:supp-mech]{C.7}).}
\label{fig:mechanism}
\vspace{-3mm}
\end{figure*}

\noindent\textbf{Change phases.} To examine how the error depends on the time since a change, we divide the test samples into five phases by the steps from the latest change to the prediction start~$t$. In \cref{fig:teaser}b, rows are datasets, columns are phases (the last being ``none before~$t$''), and each cell gives the percentage by which \ewm{}'s multi-step mean error is lower than that of the \emph{closest compared model}, the compared model with the lowest such error in that cell (named below the percentage). All cells are positive, so \ewm{} outperforms all ten compared models in every phase, as it does on \mbox{ESCP-HC} (supp.\ Sec.~\hyperref[sec:supp-E1]{E.1}). The reduction is at least 15.6\% for samples 1--4 steps after a change, which test responsiveness, and at least 28.1\% on Panda, UR5e and Hopper for samples with transient disturbances but no change in the context or horizon, which test stability (supp.\ Tab.~\hyperref[tab:supp-further]{S3}; not in \cref{fig:teaser}b).

\noindent\textbf{Robot-arm examples.} \Cref{fig:render} renders the joint angles predicted by \ewm{} and the strongest compared model as arm poses in two median cases (supp.\ Sec.~\hyperref[sec:supp-examples]{C.6}; other examples in Sec.~\hyperref[sec:supp-d]{D}): the Panda arm grasps a payload of about 1.4\,kg as its joint friction changes, and the UR5e arm releases its payload as its joint friction and motor strength return to normal. \ewm{}'s predicted joint angles follow the truth more closely, and at step 40, its mean link deviation is less than half the compared model's. Over all test samples, its step-40 error is lower on every joint of both arms (supp.\ Sec.~\hyperref[sec:supp-E2]{E.2}).

\subsection{Why \ewm{} works}
\label{sec:why}

Since the strongest compared models also adapt, accuracy alone cannot explain why \ewm{} works. We therefore test whether \ewm{}'s online memories provide responsiveness, stability, and persistence in two other examples on Panda and UR5e (\cref{fig:mechviz}a--c, more in supp.\ Sec.~\hyperref[sec:supp-E5]{E.5}) and on average over 1,586 simulated persistent changes (\cref{fig:mechviz}d--f). The \emph{error removed} by an update is the error without it minus the error with it, measured by disabling the update at inference.

In both examples, the arm grasps a payload 8 steps before the prediction starts while its joint friction (on UR5e, also its motor strength) changes. \Cref{fig:mechviz}a plots each signal against the context steps since the change (dotted line at the change, dashed line at the prediction start). At the change, \ewm{}'s one-step prediction errors jump to 4.3 times their previous value on Panda and to 31.8 times on UR5e, where an impact adds another jump at step 2. The fast-state update (``fast update'') grows at once. A step later, the slow-state update (``slow update'') grows, and the expert weights shift mainly from the internal expert marked ``losing'' to the one marked ``gaining'', which best predicts the new dynamics (gray band: the third internal expert). The prediction errors then fall.

\Cref{fig:mechviz}b,\,c show the arms at steps 5, 10, 20 and 40, overlaying the true pose (dark gray), the pose re-simulated without the change (dashed) and the pose predicted with all updates. Each row disables one update, that of the compared model (top) or one of \ewm{}'s. Darker joints mark where this update removes more error, and the number below each image is the error it removes at that step as a share of the error without it (negative: added error).

At step 5, \ewm{}'s fast correction removes 55\% on Panda and 36\% on UR5e, mostly at the payload-carrying shoulder and elbow. At step 40, the fast correction removes little, whereas the routing memory and the slow correction still remove 34\% and 74\% on Panda and 58\% and 10\% on UR5e. Here, the compared models' updates remove little at step 40 (8\% for MOLe and 5\% for ReBAL), and their predicted arms drift toward the dashed pose.

\Cref{fig:mechviz}d plots the average multi-step mean error on a logarithmic scale against the steps from the change to the prediction start, relative to that of the strongest compared models 17--30 steps after the change (gray line at 1). In the gray area, the change has not yet reached the context. After the change, \ewm{}'s error falls faster: by 59.2\% within four steps, versus 49.2\% for the strongest compared models (supp.\ Sec.~\hyperref[sec:supp-E5-single]{E.5.4}). At 20--30 steps after a change, \ewm{} is more accurate than the strongest compared models, but \ewm{} without memory updates is not, so this lead stems from these updates (gap ``online memories'').

\Cref{fig:mechviz}e uses Shapley values to split the gap in (d), the error removed by the online memories, among the fast correction, routing memory and slow correction. The three reach half of their later level (17--30 steps after the change) in 1,~3 and~7 steps (rings), so the fast correction acts first.

\Cref{fig:mechviz}f shows the share of each model's per-step error removed by its own updates 20--30 steps after a change. At step 40, \ewm{}'s updates still remove 33\%, versus 14\% for the strongest compared models. With only the fast correction at inference (``fast only''), this share falls to 11\% at step 40, whereas with only the slow one (``slow only''), it starts lower but approaches \ewm{}'s. The 33\% at step 40 requires the routing memory and slow state to stay fixed: if both decayed like the fast state, it would drop to 4\% (supp.\ Sec.~\hyperref[sec:supp-E5-held]{E.5.11}).

In contrast, the strongest compared models each satisfy only responsiveness or only persistence (supp.\ Sec.~\hyperref[sec:supp-E5-kept]{E.5.9}). ReBAL is responsive, but its recurrent state is fed its own predictions, and 20--30 steps after a change its updates remove only 1\%--9\% of its step-40 error. MOLe is persistent (27\%--33\% at step 40) but responds less promptly: \mbox{1--4} steps after a change, it removes 14\% of its step-1 error on Panda and none on Hopper, versus 37\% and 24\% for \ewm{}'s updates.

The slow state exhibits stability: on Panda, UR5e and Hopper, it moves 1.4--1.9 times as far in the four steps after a persistent change as in the four steps before, but no farther after a transient disturbance (supp.\ Sec.~\hyperref[sec:supp-E5-transient]{E.5.6}). \ewm{} thus leads after a change for three reasons. First, its fast correction acts immediately (responsiveness). Second, its slow state does not follow transient disturbances (stability). Third, its routing memory and slow state stay fixed (persistence).

\subsection{Role of each design}
\label{sec:mechanism}

\noindent\textbf{Multiple internal experts.} Preventing the prediction errors from updating the routing memory at inference raises the multi-step mean error by 0.9\% on \mbox{ESCP-HC} and by 5.1\%--30.3\% on the other datasets (\cref{fig:mechanism}a), so error-driven reweighting improves the prediction accuracy.

\noindent\textbf{Fast--slow timescale adaptation.} Removing the fast correction at inference raises the step-1 error by 5.2\%--60.9\%, and removing the slow correction raises the step-40 error by 9.9\%--37.6\% (\cref{fig:mechanism}b), whereas each removal changes the error at the other step by at most 4.9\%. Thus, the fast correction serves early steps and the slow correction late steps. Retraining with one correction confirms this (\cref{fig:mechanism}c): the slow-only model's per-step error minus the fast-only model's, in percent of \ewm{}'s, is positive (``fast-only better'') at early steps and mostly at or below zero at late steps.

\noindent\textbf{Predictive correction.} At inference, we remove both corrections (``no corrections''), zero the prediction errors (``zeroed''), delay them by 4 steps (``delayed 4 steps'') or substitute those of another sequence (``other sequence's''; supp.\ Sec.~\hyperref[sec:supp-E3]{E.3}). Every alteration raises the multi-step mean error on every dataset (\cref{fig:mechanism}d), most for another sequence's prediction errors (by 23.3\%--149.0\%); the content of the prediction errors thus matters, not merely their statistics.

\subsection{Generalization and control}
\label{sec:generalization}

With the \mbox{ESCP-HC} scale factors drawn outside the training range, \ewm{}'s multi-step mean error is at least 5.9\% below that of every compared model, and on the real ALFA flights at least 13.1\% below and lower in all 4 seeds (supp.\ Tab.~\hyperref[tab:supp-further]{S3}, Sec.~\hyperref[sec:supp-alfa]{C.4}), so its lead extends beyond the training range and to real data. In the joint-failure task of GrBAL on JSHC~\cite{nagabandi2019grbal}, with the same planner for all models, \ewm{} attains the highest mean return (93.1, 13.8 above MOLe, the second best), and more accurate models tend to obtain higher returns (Spearman rank correlation 0.75; supp.\ Secs.~\hyperref[sec:supp-control]{C.5} and \hyperref[sec:supp-E4]{E.4}).

\subsection{Ablations and hyperparameter sensitivity}
\label{sec:ablation}

Retraining \ewm{} at about 0.70M parameters without the fast correction, the slow correction, the prediction-error inputs or the more uniform expert weights ($\gamma$ fixed to~1), one at a time, raises the multi-step mean error on every dataset by 2.8\%--51.3\%, so each component contributes at equal model size. On \mbox{ESCP-HC}, for $C=32$ or 128 and for $\gamma=0.05$, 0.5 or~1, \ewm{} outperforms the four compared models with the lowest error on this dataset and TD-MPC2-40 in all 5 seeds, by 8.8\%--40.0\% across all settings (supp.\ Sec.~\hyperref[sec:supp-ablation]{C.8}).

\section{Conclusion}
\label{sec:conclusion}

\ewm{} addresses a core challenge for world models: adapting to sudden, persistent, yet initially unknown changes in robot dynamics, which requires responsiveness, stability, and persistence that single-timescale adaptation cannot jointly provide. Inspired by multi-timescale predictive coding, \ewm{} emulates the brain's neural pathway reweighting, multiple timescales and error-driven adaptation with three designs: multiple internal experts (MiE), fast--slow timescale adaptation and predictive correction, which updates the online memories without gradients. On six datasets, including real UAV flights, \ewm{} predicts more accurately than ten baselines, and each design contributes to this accuracy. Our analyses show how it meets the three requirements: the fast correction acts immediately (responsiveness), the slow state follows persistent changes but not transient disturbances (stability), and the routing memory and slow state stay fixed over the open-loop rollout (persistence). We will extend \ewm{} to visual inputs and real-robot control.

{
    \small
    \bibliographystyle{ieeenat_fullname}
    \bibliography{main}
}

\clearpage
\appendix
\renewcommand{\thesection}{\Alph{section}}
\setcounter{table}{0}\renewcommand{\thetable}{S\arabic{table}}\renewcommand{\theHtable}{S\arabic{table}}
\setcounter{figure}{0}\renewcommand{\thefigure}{S\arabic{figure}}\renewcommand{\theHfigure}{S\arabic{figure}}
\setcounter{equation}{0}\renewcommand{\theequation}{S\arabic{equation}}\renewcommand{\theHequation}{S\arabic{equation}}
\raggedbottom
\setcounter{dbltopnumber}{3}
\renewcommand{\dbltopfraction}{0.9}
\makeatletter\setlength{\@dblfptop}{0pt}\makeatother
\makeatletter\setlength{\@dblfpsep}{8pt}\setlength{\@dblfpbot}{0pt plus 1fil}\makeatother
\maketitlesupplementary

This supplementary material gives the details needed to reproduce the main paper, together with examples and more analyses. \Cref{sec:supp-a} describes the datasets, \cref{sec:supp-b} the implementation of \ewm{} and of the compared models, and \cref{sec:supp-c} the evaluation protocols of all experiments. \Cref{sec:supp-d} shows examples of the predictions on all six datasets, and \cref{sec:supp-e} gives more analyses. Tables, figures and equations numbered with the prefix S belong to this supplement. Section, equation, figure and table numbers without the prefix refer to the main paper. Throughout, the analyses relate to the three requirements for dynamics adaptation in the main paper: responsiveness, stability and persistence.

\section{Datasets}
\label{sec:supp-a}

\begin{table*}[t]
\centering
\scriptsize
\setlength{\tabcolsep}{3pt}
\begin{tabular}{>{\raggedright\arraybackslash}p{1.45cm}>{\raggedright\arraybackslash}p{2.38cm}>{\raggedright\arraybackslash}p{2.38cm}>{\raggedright\arraybackslash}p{2.38cm}>{\raggedright\arraybackslash}p{2.38cm}>{\raggedright\arraybackslash}p{2.38cm}>{\raggedright\arraybackslash}p{2.38cm}}
\toprule
 & JSHC & ESCP-HC & Panda & UR5e & Hopper & ALFA \\
\midrule
Source & joint-disable protocol following GrBAL~\cite{nagabandi2019grbal} and MBCD~\cite{alegre2021mbcd} & sudden-change protocol following ESCP~\cite{luo2022escp} & self-built & self-built & self-built & real recorded flights~\cite{keipour2021alfa} \\
Robot model & HalfCheetah (Gymnasium HalfCheetah-v5, MuJoCo 3.11) & HalfCheetah (Gymnasium HalfCheetah-v5, MuJoCo 3.11) & Franka Panda (MuJoCo Menagerie), payload body at the tool point & UR5e (MuJoCo Menagerie), payload body at the flange & Hopper (Gymnasium Hopper-v5 model), joint position servos & CMU fixed-wing UAV \\
State & 17: torso height, pitch, 6 joint angles, 9 velocities & 17: as JSHC & 14: 7 joint angles, 7 joint velocities & 12: 6 joint angles, 6 joint velocities & 11: torso height, pitch, 3 joint angles, 6 velocities & 12: attitude (2), heading sin/cos, body rates (3), NED velocity (3), altitude, airspeed \\
Action & 6: joint torques & 6: joint torques & 7: compliant joint-servo targets & 6: compliant joint-servo targets & 3: joint-servo targets & 3: autopilot roll, pitch, airspeed commands \\
Parameters that change & one of 5 of the 6 joints has zero torque & torso mass, joint damping, gravity, each $\times1.5^{U(-3,3)}$ & payload 0 or 0.5--2.5\,kg; joint friction $\times$0.1--3; motor strength 1 or $\times$0.8--1.2 & payload 0 or 1--4\,kg; joint friction $\times$0.1--3; motor strength 1 or $\times$0.8--1.2 & ground friction $\times$0.1--3; servo (motor) strength 1 or $\times$0.6--1.4; torso mass 1 or $\times$1.3--2.5 & engine or control-surface failure at a labeled time; some flights without failure \\
Change schedule & failed joint redrawn every 100 steps, random phase & new factors every 100 steps (ESCP schedule) & 6 sequence types on segments [0,40,120,190,260,300], incl.\ drift & as Panda & as Panda & failure persists after onset \\
Transient disturbances & none & none & 4--12-step pulses (tool force 5--25\,N, strength, friction); 1-step impacts & as Panda (flange) & 4--12-step pulses (strength, friction); 1-step impacts & none added \\
Splits (train/val./ test/TE-B) & 288/144/144/144 seq. & 288/144/144/144 seq. & 288/144/144/144 seq. & 288/144/144/144 seq. & 288/144/144/144 seq. & 240/86/100/122 seq.\ from 12/5/7/7 flights \\
Action generation & open-loop plans & open-loop plans & open-loop plans & open-loop plans & open-loop plans & closed-loop autopilot \\
\bottomrule
\end{tabular}
\caption{Datasets and change protocols. All sequences have 300 transitions at $\Delta t = 0.05$\,s (on ALFA, cut from the recorded flights, \cref{sec:supp-alfa}); queries use a 64-step context and the 40 following actions and states. JSHC and ESCP-HC follow protocols defined in prior work, with data generated by us; Panda, UR5e and Hopper are self-built; ALFA consists of recorded real flights. On ESCP-HC, each change switches to one of 40 fixed combinations of the three factors, drawn at random; the validation and test sequences use 40 other combinations than the training sequences. On Panda, UR5e and Hopper, the listed ranges are those from which new values are drawn; in the drift sequences, the later small steps can leave them, up to a payload of 3\,kg (Panda) and 5\,kg (UR5e), a motor strength of $\times$0.7--1.3 (arms) and $\times$0.5--1.5 (Hopper) and a torso mass of $\times$3. Splits: training / validation / test (TE) / second test split (TE-B; generated like TE with a different random seed, on ALFA seven other flights); the prediction results use TE and TE-B together as the test set. On ALFA, all comparisons use the 4 seeds shared by all models (\cref{sec:supp-stage1}).}
\label{tab:datasets}
\end{table*}

\Cref{tab:datasets} lists, for each of the six datasets, the robot model, the state and action spaces, the parameters that change and how these parameters change, the transient disturbances and the data splits. Section~4.1 of the main paper gives a summary. The three datasets newly built by us (Panda, UR5e and Hopper) are based on standard robot models from Gymnasium and MuJoCo Menagerie~\cite{todorov2012mujoco,towers2024gymnasium,zakka2022menagerie}. On Panda and UR5e, the payload, the joint friction and the motor strength change, and on Hopper, the torso mass, the ground friction and the motor strength change. Simulation provides the ground-truth change times and physical parameters. Each simulated dataset is split into training, validation and test sets by whole episodes (an episode, also called a sequence, is one complete simulation run from start to end), and the episodes generated together as one group (\cref{sec:supp-metrics}) always fall into the same split.

\section{Implementation details}
\label{sec:supp-b}

\subsection{Hyperparameters of \ewm{}}
\label{sec:supp-hparams}

\begin{table*}[t]
\centering
\scriptsize
\setlength{\tabcolsep}{3pt}
\begin{tabular}{>{\raggedright\arraybackslash}p{1.75cm}>{\raggedright\arraybackslash}p{5.2cm}>{\raggedright\arraybackslash}p{2.0cm}rrr>{\raggedright\arraybackslash}p{3.4cm}}
\toprule
Model & Main sizes (range over datasets) & Optimizer (native LR) & LR mult. & Peak LR & Clip & Chosen on validation \\
\midrule
\ewm{} (ours) & Stage 1 (then fixed): router $2\times128$, 3 base experts $3\times256$ (434--447k). Stage 2: $U_q$ and $U_F$, each $2\times128$; readout network $G^F_k$ $2\times128$ per internal expert; fast state 32 and slow state 3 per internal expert; hidden widths of the measurement network 86--108 and 116--146, of the readout network $G^S_k$ 73--92 and 75--94, and of the filter network 18--23 & Adam ($10^{-4}$) & $\times10$ & $10^{-3}$ & 1.0 & inverse routing temperature $\gamma$ per seed: JSHC 0.127--0.15, ESCP-HC 0.15--0.2, Panda 0.1, UR5e 0.05--0.15, Hopper 0.2--0.3, ALFA 0.127--0.2 \\
TD-MPC2 & latent 128 (SimNorm 8), encoder $1\times256$, dynamics/decoder width 480--482, consistency horizon 3 & Adam ($3\cdot10^{-4}$; encoder $\times0.3$) & $\times10$ & $3\cdot10^{-3}$ & 20 & -- \\
TD-MPC2-40 & as TD-MPC2; consistency/decoder horizon 40 & Adam ($3\cdot10^{-4}$; encoder $\times0.3$) & $\times1$ & $3\cdot10^{-4}$ & 20 & -- \\
ReBAL & residual MLP recurrent cell, width 580--584 & Adam ($10^{-2}$) & $\times10$ & 0.1 & 1.0$^*$ & -- \\
GrBAL & MLP $3\times$581--585; one inner step on the last 16 transitions & Adam, outer ($10^{-3}$) & $\times3$ & $3\cdot10^{-3}$ & 1.0$^*$ & inner step: JSHC 0.1, ESCP-HC $10^{-4}$, Panda 0.1, UR5e 0.01, Hopper 0.1, ALFA 0.01 \\
MOLe & MLP $3\times$581--585; inner step 0.01, M-step size 0.01, CRP concentration 1 & Adam, outer ($10^{-3}$) & $\times3$ & $3\cdot10^{-3}$ & 1.0$^*$ & -- \\
MBCD & ensemble of 5 (2 elites), MLP $4\times$98--102; library of $\le4$ models & Adam ($10^{-3}$) & $\times10$ & $10^{-2}$ & 1.0$^*$ & -- \\
MTS3 & latent observation 30, window length 15 (MTS3's $H$); hidden 734--777 / 1468--1554 & Adam ($3\cdot10^{-3}$) & $\times0.1$ & $3\cdot10^{-4}$ & 5 & -- \\
DreamerV3-style RSSM & deter 512, stoch $32\times4$, units 64, core hidden 416--417 & LaProp + AGC ($4\cdot10^{-5}$) & $\times30$ & $1.2\cdot10^{-3}$ & AGC 0.3 & evaluation: mean of 16 samples (all datasets) \\
GPMM & $\le5$ GP components, $\le$3{,}488--5{,}592 stored points each, CRP concentration 1.5 & Adam on GP hyperparameters (5 steps, 0.1) & -- & -- & -- & -- \\
MLP & MLP $4\times$476--478 (SiLU), history length 1 & Adam ($3\cdot10^{-4}$) & $\times3$ & $9\cdot10^{-4}$ & 1.0 & -- \\
\bottomrule
\end{tabular}
\caption{Model sizes and training settings. Common to all gradient-trained models: 16{,}000 updates with batches of 32 windows (each a 64-step context and the 40 steps after it, used with each model's own training objective, \cref{sec:supp-baselines}), the same sampling schedule per seed, linear warm-up over 320 updates (DreamerV3-style RSSM: 1{,}000) followed by cosine decay, validation every 1{,}000 updates with best-checkpoint selection, 5 seeds (4 used on ALFA); for MBCD, the cap applies to each of its up to four library models, which stop earlier by MBCD's native early-stopping rule, and GPMM builds its library in one streaming pass over the training data. MBCD and GPMM set aside 10\% of the training sequences as their own holdout (MBCD for its early stopping and its detection threshold). Peak learning rate $=$ native rate $\times$ a multiplier chosen once from $\{0.1, 0.3, 1, 3, 10\}$ (one extra step, $\times 30$ or $\times 0.03$, when the choice fell on a boundary) on the validation split of a separate HalfCheetah development dataset and transferred to every dataset. CRP: Chinese restaurant process; AGC: adaptive gradient clipping. $^*$No native clipping; global-norm clipping at 1.0 under our protocol.}
\label{tab:hparams}
\end{table*}

Number of internal experts $K=3$; context length $C=64$, prediction horizon $H=40$; $\alpha=0.5$ in Eq.~(1); fast-state dimension $d_F=32$, decay factor $\rho=0.8$ (the fast state decays by 20\% per step, half-life of about 3 steps); slow-state dimension $d_c=3$, initial variance $\sigma_0^2=1$, process noise $\sigma_P^2=0.01$; in Eq.~(8), $\sigma=0.5$, $\lambda=0.5$ and $\mathcal{H}=\{1,5,10,20,40\}$. Stage 1: segment length $n_b=5$, $\lambda_b=0.5$, $\lambda_u=0.01$, a usage floor of 0.1 in $\mathcal L_{\text{use}}$, and the same mixture-Gaussian $\sigma$ as above; AdamW (learning rate $3\times10^{-4}$, weight decay $10^{-4}$), batch size 128, at most 30 epochs, stopped early once the best candidate epoch (\cref{sec:supp-stage1}) lies 10 epochs back; the expert-alive check (\cref{sec:supp-stage1}) requires the usage of every base expert to be at least 0.05. $\gamma$ is selected from 10 candidate values by the multi-step mean error of the base experts and the router on the validation set (before stage~2, \ie, with $\vq_{j+1}=\vp_j$ and no corrections). Stage 2 uses 16,000 Adam updates in total, with about \mbox{0.25--0.27M} trainable parameters (about 0.70M including the base experts and the router). The remaining hyperparameters are given in \cref{tab:hparams} and in \cref{sec:supp-method,sec:supp-stage1}.

\paragraph{Networks.} All networks are MLPs. The base experts, the router, $U_q$, $U_F$ and the readout networks $G^F_k$ use SiLU activations (the router, $U_q$ and $U_F$ with layer normalization). The measurement network $U^{\mathrm m}_S$, the filter network $U^{\mathrm f}_S$ and the readout networks $G^S_k$ use tanh. Each internal expert's $U^{\mathrm m}_S$ and $U^{\mathrm f}_S$ together form its slow-state update $U_S$ (Sec.~3.3). One network $U_F$ is shared by all internal experts. The network sizes are listed in \cref{tab:hparams}. The measurement network, the filter network and the readout networks $G^S_k$ are sized per dataset so that the whole model has about 0.70M parameters. No recurrent cell such as a GRU is used: the fast state follows the fixed decay of Eq.~(2).

\paragraph{Further details.} The 10 candidate values of $\gamma$ are $\{0.05, 0.08, 0.1, 0.127, 0.15, 0.2, 0.3, 0.5, 0.7, 1\}$; $\gamma$ is selected separately for each seed, for its base experts and router, and ties go to the larger value. If the best value lies on the boundary of this set, one further value (0.03 below, 1.4 above) is also scored; this happened for two seeds on UR5e, where 0.03 was scored but not selected. The selected values are listed in \cref{tab:hparams}. In stage 1, the training segments are taken with stride 1 from the training episodes and reshuffled every epoch, and the global gradient norm is clipped at 10. \Cref{sec:supp-baselines} describes how the settings of all models were chosen without the test data.

\subsection{Method details}
\label{sec:supp-method}

\begin{figure*}[t]
  \centering
  \includegraphics[width=\textwidth]{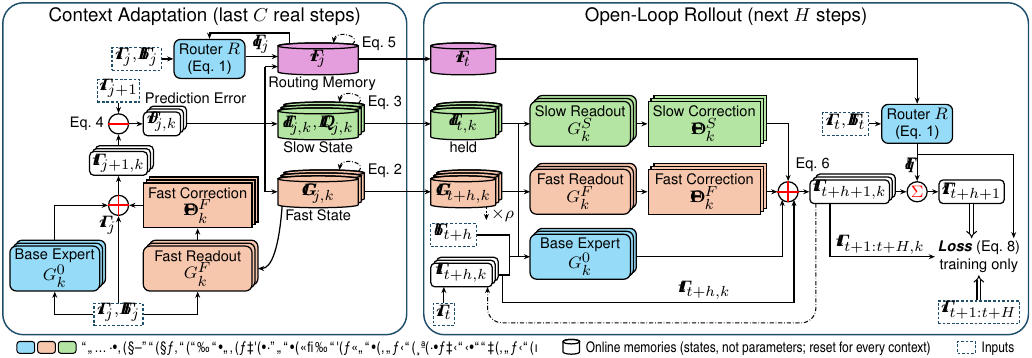}
  \caption{Computation of \ewm{} in detail, complementing the architecture in Fig.~2 of the main paper. Left, context adaptation (Secs.~3.2--3.4): each one-step prediction error updates first the pink routing memory, then the orange fast states and the green slow states. Right, open-loop rollout (Sec.~3.5): the expert weights $\hvp$ are computed once, the fast states decay, the slow states stay fixed, and the internal experts' corrected predictions are combined with these expert weights. The routing memory adjusts the expert weights. Only the online memories (cylinders) change at inference. Equation numbers refer to the main paper. The curved arrows marked Eq.~2, Eq.~3 and Eq.~5 are the updates $U_F$, $U_S$ and $U_q$. For clarity, some inputs are not drawn, \eg, the fast state, which the slow-state update $U_S$ and the readout network $G^S_k$ also read (Sec.~3.3).}
  \label{fig:compute}
\end{figure*}

\Cref{fig:compute} shows the computation of \ewm{} in detail and complements the architecture in Fig.~2 of the main paper. Its left part shows how one context transition is read (Secs.~\mbox{3.2--3.4} of the main paper), and its right part the open-loop rollout~of~Sec.~3.5.

\paragraph{Lower bound on each expert weight.} In the implementation, \mbox{Eq.~(1)} contains a small probability floor $\beta=10^{-3}$ on both the routing memory and the output:
\begin{equation}
  \begin{aligned}
    \pi_\gamma(\vz,\va;\vq)={}&(1-\beta)\,\mathrm{softmax}\Big(\gamma\,R(\vz,\va)\\
    &+\alpha\log\big((1-\beta)\,\vq+\tfrac{\beta}{K}\big)\Big)+\tfrac{\beta}{K}.
  \end{aligned}
  \label{eq:floor}
\end{equation}
Every expert weight is therefore at least $\beta/K$ ($1/3000$ for $K=3$). The same function gives the expert weights $\vp_j$ in Eq.~(5) and the fixed expert weights $\hvp$ of the open-loop rollout (Sec.~3.5).

\paragraph{Prediction-error clipping.} Before entering the routing-memory update $U_q$ (Eq.~(5)) and the fast-state update $U_F$ (Eq.~(2)), the prediction error of each internal expert is clipped by its Euclidean norm: a vector $\boldsymbol{v}$ is replaced by
\begin{equation}
  \boldsymbol{v}\cdot\min\!\Big(1,\ \frac{c_e}{\lVert \boldsymbol{v}\rVert_2+10^{-8}}\Big).
  \label{eq:clip}
\end{equation}
The threshold $c_e$ is computed once per seed after stage 1: it is the 99th percentile, over all transitions of the training set, of the smallest one-step prediction-error norm among the base experts, $\min_k\lVert \vz_{j+1}-\vz_j-G^0_k(\vz_j,\va_j)\rVert_2$ (in standardized coordinates). The slow-state update (Eq.~(3)) uses the unclipped prediction error $\veps_{j,k}$.

\paragraph{Slow-state update.} The implementation of Eq.~(3) adds small constants and truncates the variance update. The uncertainty is floored before use, $\bar{\vP}_{j,k}=\max(\vP_{j,k},10^{-4})$ (element-wise), and $\bar{\vP}_{j,k}$ replaces $\vP_{j,k}$ everywhere in Eq.~(3). $\vS_k$ contains an additional constant, $\vS_k=\bar{\vP}_{j,k}\odot\operatorname{diag}(\vJ_k^{\top}\vJ_k)+\vr_k+10^{-3}$. The factor that shrinks the variance is truncated to the interval $[0.02,1]$ element-wise:
\begin{equation}
  \begin{aligned}
    \vP_{j+1,k}={}&\operatorname{clip}\Big(1-\tfrac1D\,\vg_k\odot\operatorname{diag}(\vJ_k^{\top}\vJ_k),\ 0.02,\ 1\Big)\\
    &\odot\bar{\vP}_{j,k}+\sigma_P^2 .
  \end{aligned}
  \label{eq:trunc}
\end{equation}
The sensitivity is bounded, $\vJ_k=\tanh(\tilde{\vJ}_k)$, where $\tilde{\vJ}_k$ is the corresponding output of the measurement network, and the measurement noise is $\vr_k=0.05+\operatorname{softplus}(\tilde{\vr}_k)$, where $\tilde{\vr}_k$ is the output of the filter network. Hence $\vS_k\ge0.051$ and the gain $\vg_k=\bar{\vP}_{j,k}\oslash \vS_k$ is always finite.

\paragraph{Definition of $\mathrm{MSE}_n$.} In Eq.~(8), $\mathrm{MSE}_n$ is the average of two normalized squared errors of the final prediction $\hvz$ over steps 1 to $n$ of the prediction horizon:
\begin{equation}
  \begin{aligned}
    \mathrm{MSE}_n=\frac12\Big[&\frac{1}{nD}\sum_{h=1}^{n}\big\lVert\hvz_{t+h}-\vz_{t+h}\big\rVert^2\\
    &+\frac{1}{nD\,\bar\sigma^2}\sum_{h=1}^{n}\big\lVert \boldsymbol{\sigma}_z\odot(\hvz_{t+h}-\vz_{t+h})\big\rVert^2\Big],
  \end{aligned}
  \label{eq:msen}
\end{equation}
where the states are standardized per dimension on the training set (as are the actions), $\boldsymbol{\sigma}_z\in\mathbb{R}^D$ holds the training-set standard deviations of the state dimensions (so $\boldsymbol{\sigma}_z\odot(\hvz-\vz)$ is the error in the original physical units), and $\bar\sigma^2=\frac1D\sum_{d}\sigma_{z,d}^2$ is their mean variance. The term without $\boldsymbol{\sigma}_z$ is the nMSE used for evaluation, and the term with $\boldsymbol{\sigma}_z$ weights the state dimensions by their physical scale.

\subsection{Stage 1: loss and the expert-alive check}
\label{sec:supp-stage1}

\paragraph{Stage-1 loss.} The base experts and the router are trained jointly on training segments of length $n_b$ with the loss $\mathcal L_{\text{base}}$ of Eq.~(7) of the main paper.
$\mathcal L_{\text{step}}$ and $\mathcal L_{\text{traj}}$ are mixture-Gaussian negative log-likelihoods of the one-step prediction and of each base expert's $n_b$-step open-loop trajectory from the segment start; the mixture likelihood assigns the training signal by how well each base expert predicts, so no change labels are used. $\mathcal L_{\text{bal}}$ pushes the average expert weights toward uniform, and $\mathcal L_{\text{use}}$ keeps each base expert's usage above a floor; both prevent a collapse onto a few base experts. The epoch whose base experts and router are kept is chosen as follows.

\paragraph{Expert-alive check.} After every stage-1 epoch, the base experts and the router are evaluated on all validation segments of length $n_b$, and the epoch counts as \emph{alive} if all of the following hold:
\begin{enumerate}[(i)]
  \item the average expert weight that the router gives each base expert (computed before the next state is seen) is at least 0.05;
  \item every base expert's average one-step responsibility, \ie, its posterior share under the one-step mixture likelihood, is at least 0.05;
  \item for every pair of base experts, the root-mean-square difference between their predicted state increments, averaged over the validation transitions, is at~least~$10^{-4}$;
  \item for every base expert, moving its weight to any other base expert raises the mean one-step validation negative log-likelihood by at least $10^{-5}$.
\end{enumerate}
Criteria (i) and (ii) implement the usage threshold of 0.05 in \cref{sec:supp-hparams}; criteria (iii) and (iv) rule out duplicate and redundant base experts. An epoch is a candidate if it is at least the 4th epoch (the 5th on ALFA, whose smaller training set gives fewer updates per epoch) and both this and the preceding epoch are alive. Among the candidates, the epoch with the lowest validation loss $\mathcal L_{\text{step}}+\lambda_b\mathcal L_{\text{traj}}$ is selected. Training runs for at most 30 epochs and stops once 10 epochs have passed since the best candidate epoch. If no epoch becomes a candidate, that seed is left out and not replaced; this happened for one seed on ALFA (\cref{tab:datasets}), so all ALFA comparisons use the other 4 seeds, which all models share.

\subsection{Compared models and training protocol}
\label{sec:supp-baselines}

Following the three classes of Sec.~2 of the main paper, we select ten compared models: four general-purpose world models, namely RSSM (a DreamerV3-style recurrent state-space model~\cite{hafner2025dreamerv3}), TD-MPC2~\cite{hansen2024tdmpc2}, TD-MPC2-40 trained with a 40-step loss, and a plain MLP; five online-adaptive world models, namely GrBAL, ReBAL~\cite{nagabandi2019grbal}, MOLe~\cite{nagabandi2019mole}, MBCD~\cite{alegre2021mbcd}, and the Gaussian-process mixture model GPMM~\cite{xu2020gpmm}; and the multi-timescale world model MTS3~\cite{shaj2023mts3}. To rule out effects of model capacity, all neural models have about 0.70M parameters (for \ewm{} including the base experts and the router); GPMM is non-parametric, and the number of points it stores is capped so that its stored values (hyperparameters and stored points) also number about 0.70M (\cref{tab:hparams}). All models run under the same training and evaluation pipeline, with the same data splits, 64-step context and 40-step prediction horizon. Models trained by gradients also share the same cap on the number of updates (for MBCD, for each model of its library) and the same batch size (\ewm{} additionally trains its base experts and router in stage~1). The base experts and the router alone, \ie, \ewm{} without memory updates, do not account for \ewm{}'s lead: 20--30 steps after a persistent change, their multi-step mean error is higher than that of the strongest compared model on Panda and UR5e, and on the other three simulated datasets the 95\% interval of the difference includes zero (\cref{sec:supp-E5-datasets}). To avoid tuning on evaluation data, the learning-rate multiplier of each model is selected once, on a HalfCheetah development dataset that is not used for evaluation, and then kept for all datasets. Apart from the checkpoints (\cref{tab:hparams}) and the stage-1 epoch (\cref{sec:supp-stage1}), the only quantities selected on the validation set are the $\gamma$ of \ewm{} (for each seed, \cref{sec:supp-hparams}) and, for each dataset, the inner step size of GrBAL (the step size of its inference-time gradient adaptation) and the evaluation mode of the DreamerV3-style RSSM (the mean of 16 samples on every dataset; \cref{tab:hparams}). MOLe's noise variance is estimated on the validation sequences, without labels. \Cref{tab:hparams} lists the sizes, optimizers and learning rates of all models.

\paragraph{Training objectives.} \ewm{}'s stage 2 is trained on its open-loop rollouts over the whole 40-step horizon (Eq.~(8) of the main paper). ReBAL, GrBAL and MOLe are trained, as in their original works, on one-step predictions of the state increment. ReBAL predicts each next state from the true states before it (teacher forcing) along the whole training window. GrBAL and MOLe meta-train their networks with model-agnostic meta-learning (MAML), so that one gradient step on 16 consecutive transitions improves the one-step predictions of the next 16. TD-MPC2 and TD-MPC2-40 train their latent dynamics on open-loop rollouts of 3 and 40 steps, so TD-MPC2-40 shows how a general-purpose model performs when it is trained over the whole horizon. The MLP is trained with \ewm{}'s stage-2 loss (Eq.~(8) of the main paper, with a single expert) on its 40-step open-loop rollouts, and MTS3 with its original Gaussian negative log-likelihood on the 40 predicted steps. The DreamerV3-style RSSM is trained with its reconstruction and KL losses on the whole training window, MBCD with the Gaussian negative log-likelihood of one-step state increments, and GPMM by maximizing the marginal likelihood of its Gaussian processes. The comparisons with ReBAL and MOLe in Sec.~4.3 of the main paper and \cref{sec:supp-E5} therefore concern these two models as trained with their original objectives.

\section{Evaluation protocols}
\label{sec:supp-c}

This section describes how each experiment of Sec.~4 of the main paper is run and how the examples are chosen (\cref{sec:supp-examples,sec:supp-mech}). The main results are given in the main paper, the ablations and the sensitivity test in \cref{sec:supp-ablation}, and further analyses in \cref{sec:supp-e}.

\subsection{Metrics and intervals}
\label{sec:supp-metrics}

A query, also called a window or, in the main paper, a test sample, consists of a 64-step context and the 40 following actions and states. Each test sequence gives 5 queries, whose predictions start at steps 64, 96, 128, 160 and 192 (1{,}440 queries per simulated dataset and 1{,}110 on ALFA). The per-step error at predicted step $h$ is the squared error of the prediction, averaged over the state dimensions, with each dimension normalized by its training-set standard deviation (nMSE). The \emph{multi-step mean error} averages the mean errors over horizons $n\in\mathcal H=\{1,5,10,20,40\}$, where the mean error over a horizon $n$ averages the per-step errors at steps $1,\dots,n$. For each seed, these errors are averaged over all queries of both test splits, and the reported value is the mean over seeds. A relative difference $(\text{\ewm{}}-B)/B$ compares \ewm{} with a compared model $B$ on the same queries and seeds; negative values mean that \ewm{} has the lower error. A reduction is the same quantity with the sign reversed, so a positive reduction means that \ewm{} has the lower error.

Unless stated otherwise (as in \cref{sec:supp-E5}), the 95\% intervals are percentile intervals from a bootstrap with 10{,}000 replicates. In each replicate, the groups of 12 related test episodes that were generated together are resampled with replacement within each test split (on ESCP-HC, whose episodes are generated independently, a group is a block of 12 consecutive episodes). On ALFA, one group is one flight. The same draws are used for all models and seeds, and the seeds are not resampled. Each percentage divides the paired mean difference by $B$'s mean error on the same seeds, and this denominator stays fixed in the bootstrap.

\subsection{Change phases and windows without a change}
\label{sec:supp-phase}

For JSHC, Panda, UR5e and Hopper, each test query is assigned to one of five phases by the number of steps between the latest change and the prediction start: 1--4, 5--19, 20--59, at least 60, or no change before the start. On ESCP-HC, all parameters are redrawn every 100 steps, so the phases are defined by where the latest change lies relative to the 40-step horizon: a change inside the horizon at step 20 or earlier, a change inside the horizon after step 20, and otherwise 0--15, 16--63, or at least 64 steps since the latest change or no change. Transient pulses and impacts do not count as changes. A window without a change contains no change in its context or horizon, but on Panda, UR5e and Hopper it may still contain transient disturbances. It counts as a window with transient disturbances if at least one transition in the last 24 context steps or in the horizon lies inside a friction or motor-strength pulse or carries an impact (not tool-force pulses). Figure~1b of the main paper shows, for each phase of JSHC, Panda, UR5e and Hopper, the reduction of \ewm{}'s multi-step mean error relative to the closest compared model, \ie, the smallest of the ten reductions. \Cref{fig:E-phases} adds ESCP-HC.

\subsection{Changes beyond the training range}
\label{sec:supp-ood}

All models are trained on ESCP-HC data whose scale factors are $1.5^u$ with $u\sim U(-3,3)$. For this test, new test data are generated with $u\sim U(-4,4)$, and only the 916 queries whose context or horizon contains at least one factor outside the training range are evaluated. If a model predicts a non-finite value on a query, this query is left out of the comparison between \ewm{} and that model for that seed, so the comparison uses only queries on which both predictions are finite. This rule removed one query, for MOLe (in three of its five seeds).

\subsection{Real flight data}
\label{sec:supp-alfa}

ALFA consists of real fixed-wing flights in which the engine or one or more control surfaces fail at a labeled time; some flights have no failure (\cref{tab:datasets}). The actions are the autopilot's commanded roll, pitch and airspeed recorded in flight, not the recorded control-surface and throttle outputs, which freeze at the failure and would reveal it. A flight cannot be repeated with a different action sequence, so only prediction is evaluated. The recorded signals are resampled to a common 20\,Hz grid ($\Delta t=0.05$\,s), and each flight is cut into sequences of 300 transitions, one starting every 160 steps regardless of the failure time. A sequence is kept only if all signals are present without long gaps and the aircraft is armed and in an autonomous flight mode. The sequences are split by whole flights. All models are trained and evaluated with the same protocol as on the simulated datasets, and all comparisons use the 4 seeds shared by all models (\cref{sec:supp-stage1}).

\subsection{Control task}
\label{sec:supp-control}

The task is the joint-failure HalfCheetah task of GrBAL~\cite{nagabandi2019grbal}: the robot runs forward while one of five of its six leg joints (back shin, back foot, front thigh, front shin, front foot) has zero torque. The failed joint switches to a different joint at model-predictive-control (MPC) steps 40 and 140, as in the switching protocol of MBCD~\cite{alegre2021mbcd}, and the models are not told which joint has failed. The reward at each step is the forward velocity minus $0.05$ times the squared norm of the applied, masked action, and the return is the sum of rewards over 240 MPC steps (12\,s). Each episode starts with 64 steps of open-loop actions generated as in the JSHC data, which fill the context, and the 48 test episodes are disjoint from~the~datasets.

All models use the same planner, the cross-entropy method (CEM) of PETS~\cite{chua2018pets}, without any per-model tuning: at every control step it samples 500 candidate action sequences over a 30-step horizon, keeps the 50 best, refits the sampling distribution with smoothing 0.1, repeats this for 5 iterations, executes the first action of the final mean of the sampling distribution and replans at the next step. The planning cost of a candidate is the sum over the horizon of the negative model-predicted forward velocity plus $0.05$ times the squared action norm, and candidates with non-finite predictions get a very large cost. Each world model uses its trained JSHC checkpoint of each of the 5 seeds. The 95\% intervals of the differences in mean return are percentile intervals from a bootstrap with 10{,}000 replicates that resamples the 48 test episodes and the 5 seeds. The rank correlation in Sec.~4.5 of the main paper is the Spearman correlation, over the eleven models, between the negated multi-step mean error on the JSHC test set and the mean return, so a positive value means that more accurate models have higher returns.

\subsection{Examples and their selection}
\label{sec:supp-examples}

The examples of Fig.~4 of the main paper convert the predicted joint angles of Panda and UR5e back into arm poses. On each robot, the compared model is the strongest one in Tab.~1 of the main paper: MOLe on Panda and ReBAL on UR5e. In the Panda example, a payload of about 1.4\,kg is attached to the end effector 8 steps before the prediction start and the joint friction changes with it. In the UR5e example, the payload is removed 8 steps before the prediction start, and the friction and motor strength return to normal. In both examples, transient disturbances (pulses of friction or motor strength, and on UR5e also an impact) occur during the prediction. The per-joint errors compared in Sec.~4.2 of the main paper average, over all test queries and seeds, the normalized squared error of each joint (its angle and angular velocity, \cref{sec:supp-E2}) at step 40.

\paragraph{Samples and rendering.} On each robot, the 336 test samples with a change near the prediction start are those in which the payload, the joint friction or the motor strength changes between 20 steps before and 9 steps after the prediction start (transient pulses and impacts do not count as changes, and a window that occurs identically in several test episodes counts once). For each of these samples, we subtract \ewm{}'s multi-step mean error from the compared model's, with both models from the training seed with the smallest number. The example shown is the sample at rank 168 of 336 when sorted from \ewm{}'s largest lead down, \ie, \ewm{}'s median case rather than its best one. The predicted pose is overlaid on the true pose at 55\% opacity. The link deviation is the mean displacement of the visible surface points of a link between the predicted and the true pose, and the color scale runs from 0 to the largest link deviation of both models over the rendered steps, rounded up to a whole centimeter. In Fig.~4 of the main paper, each robot has its own color scale (Panda 0--21\,cm, UR5e 0--9\,cm). The number below each frame is the mean over the links. The frames show predicted steps 5, 10, 20 and 40, and the poses use only the predicted joint angles, while the multi-step mean error also counts the joint velocities.

\paragraph{Other examples.} Unlike those of Fig.~4, the examples of Fig.~5 of the main paper and of \cref{sec:supp-d,sec:supp-E5} are chosen only among windows in which \ewm{} is more accurate than the compared model at each shown step. These examples illustrate what \ewm{}'s lead looks like where it is visible, not how large it is on average, which the averages show (Tab.~1 and Figs.~3 and 5d--f of the main paper, \cref{sec:supp-e}). The rule for \cref{sec:supp-d} follows, and those for Fig.~5 and \cref{sec:supp-E5} are given in \cref{sec:supp-mech}.

\paragraph{Examples on all six datasets.} The examples of \cref{sec:supp-d} come from the stored test predictions of the training seed with the smallest number. Each is a test query in which \ewm{}'s multi-step mean error and its mean errors over steps 1 to $n$, for every $n$ up to 40, are lower than the compared model's, its mean link deviation is lower at steps 10, 20, 30 and 40, and the state quantities drawn in the state curves stay closer to the truth at all 40 steps. Among these queries, we show examples of different kinds of change: the kind of change sequence on Panda, UR5e and Hopper, plus windows with only transient disturbances, the failed joint on JSHC, the parameter that changes most and the direction of its change on ESCP-HC, and on ALFA the type of failure and whether the query lies before or after it. Within a kind, we mostly take the query whose reduction of the multi-step mean error is closest to the middle of the reductions in that kind.

\subsection{\texorpdfstring{\protect\raggedright}{}Tests of the designs and analyses around~a~change}
\label{sec:supp-mech}

Unless stated otherwise, the tests of the designs in Sec.~4.4 of the main paper and \cref{sec:supp-E3} intervene at inference on the trained \ewm{} of each seed, without retraining, and report the relative increase of the error over the unaltered model, $100\times(M_{\text{altered}}-M)/M$, where $M$ is the error of the unaltered model; this denominator stays fixed in the bootstrap.

\paragraph{Routing test.} In the routing test of Fig.~6a of the main paper, the routing memory is no longer updated by the prediction errors, \ie, $\vq_{j+1}=\vp_j$ in Eq.~(5), while the fast and slow states are updated as usual.

\paragraph{Removing one correction at inference.} The context is read as usual, so all online memories are unchanged. During the 40-step open-loop rollout, either the fast correction $\vDelta^F_k$ or the slow correction $\vDelta^S_k$ of every internal expert is set to zero; when the fast correction is removed, the slow correction still reads the unaltered fast state. Figure~6b of the main paper reports the increase of the step-1 and step-40 errors. For the comparison by time since a change on Panda and UR5e (\cref{sec:supp-E3}), the increase of the multi-step mean error is computed separately for queries whose latest change lies 1--4 steps before the prediction start and for queries whose latest change lies at least 60 steps before it or that have no earlier change; in both groups, no change occurs within~the~horizon.

\paragraph{Retrained single-correction models.} The models that keep only the slow or only the fast correction are the corresponding ablations of \cref{fig:supp-ablsens}a (\cref{sec:supp-ablation}). Figure~6c of the main paper shows, at each predicted step, the per-step error of the slow-only model minus that of the fast-only model, divided by the per-step error of \ewm{}.

\paragraph{Changing the role of the prediction errors.} Figure~6d of the main paper uses four alterations. (i) Both corrections are removed at the prediction start (``no corrections''), which keeps only the base experts, the router and the routing memory obtained from the context. (ii) All prediction errors of the context are set to zero (``zeroed'') before entering the memory updates. The routing memory then follows the expert weights ($\vq_{j+1}=\vp_j$), the fast states stay zero, and the slow states move only with the bias $\vb_k$ of the measurement model in Eq.~(3). (iii) The prediction errors are delayed by 4 steps (``delayed 4 steps''), so that each update receives the prediction error of 4 transitions earlier. (iv) The prediction errors of the context are replaced with those of a query from another test sequence (``other sequence's'').

\paragraph{Linear probes.} After the 64-step context of each query is read, either the slow states of all internal experts ($3\times3$ values) or their fast states ($3\times32$ values) are taken as features, standardized with training-set statistics. The targets are the current change parameters at the last context transition: the failed joint on JSHC (5 classes), the logarithms of the three scale factors on ESCP-HC, and on Panda, UR5e and Hopper the payload (the torso mass on Hopper) and the logarithms of the friction and motor-strength factors. Transient pulses are left out of the targets. Continuous targets use ridge regression and the failed joint a multinomial L2 logistic regression; the regularization strength is chosen on the validation set from a fixed grid, and the probe is refit on the training and validation windows (5 windows per training sequence) and scored on the test queries. The score is the mean $R^2$ over the three targets, or the balanced accuracy for the failed joint, whose chance level is 0.2. As a control, the same slow-state probe is fit after the training and validation labels are shuffled with a fixed seed and scored with the true test labels.

The inference-time analyses above re-run each trained model, and we use a re-run only if its unaltered predictions reproduce the stored ones within $10^{-5}$. For one UR5e seed, the re-runs for removing one correction at inference, for the routing test and for all probes were computed on a different GPU type and differ by up to $2.2\times10^{-5}$; therefore, on UR5e, these analyses use the other 4 seeds. All other re-runs of the analyses above pass the check. The analyses around a change (Fig.~5d--f of the main paper and \cref{sec:supp-E5}) are separate re-runs that are not selected by this check; on UR5e, they use the same 4 seeds for every model they compare, and \cref{sec:supp-E5-measures} reports how closely they reproduce the stored predictions. All other analyses use 5 seeds on the simulated datasets and, on ALFA, the 4 seeds shared by all models.

\paragraph{Examples of Fig.~5.} On Panda and UR5e, the candidates are the windows whose context ends 8 steps after one of the persistent changes of \cref{sec:supp-E5-measures} in which the payload increases. These windows are predicted by the trained models of one seed per robot, the one with the smallest number among the seeds used in \cref{sec:supp-E5}. Unlike the samples of Fig.~4, a window that occurs identically in several test episodes counts once for each of these episodes. We keep the windows in which, at each of the shown steps 5, 10, 20 and 40, both \ewm{}'s per-step error and its mean link deviation are lower than those of the compared model and of \ewm{} without memory updates. The example shown is the kept window at the middle rank of \ewm{}'s reduction of the multi-step mean error relative to the compared model; with an even number of kept windows, it is the one of the two middle windows with the smaller reduction. Transient disturbances are not excluded, and both examples contain some within the 40 predicted steps. In the titles of Fig.~5b,\,c, a factor such as friction $\times$1.58 is the value after the change divided by the value before it. On UR5e, the joint friction goes from 1.85 to 2.91 times normal and the motor strength from 1.18 to 0.80 times normal, within the ranges of \cref{tab:datasets} (the factors in the titles are computed before rounding).

\paragraph{Examples of where the corrections act.} For each simulated dataset, \cref{sec:supp-E5-body} shows a window whose context ends 8 steps after one of the persistent changes of \cref{sec:supp-E5-measures}, predicted by the models of the seed with the smallest number among those used there. It is chosen among the windows in which \ewm{}'s per-step error is lower than the compared model's at each of the five shown steps (1, 5, 10, 20 and 40), as the one at the middle rank of \ewm{}'s reduction of the multi-step mean error relative to the compared model; with an even number of such windows, it is the one of the two middle windows with the smaller reduction. Windows are counted as in the previous paragraph.

\subsection{Ablations and hyperparameter sensitivity}
\label{sec:supp-ablation}

\begin{figure*}[t]
\centering
\includegraphics[width=0.86\textwidth,trim=0 6.4 0 0,clip]{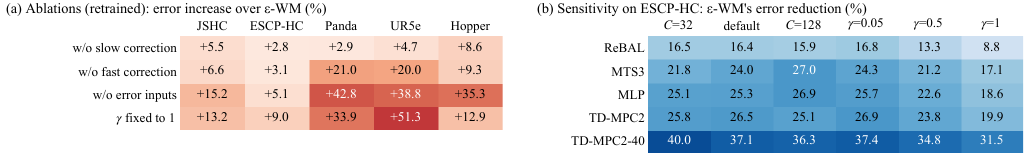}
\caption{Multi-step mean error. \textbf{(a)}~Ablations, retrained at about 0.70M parameters: ``w/o error inputs'' sets the prediction errors fed to the online memories to zero, and ``$\gamma$~fixed to~1'' removes the more uniform expert weights. \textbf{(b)}~Sensitivity to $C$ and $\gamma$ on ESCP-HC: \ewm{}'s error reduction relative to each compared model (row) for each setting (column; ``default'': $C=64$ and the validated $\gamma$).}
\label{fig:supp-ablsens}
\end{figure*}

In the ablations of \cref{fig:supp-ablsens}a, which Sec.~4.6 of the main paper summarizes, one component is removed at a time and the model is retrained. When a correction is removed, the remaining networks are widened so that the parameter count stays at about 0.70M. In the variant without the prediction-error inputs, the prediction errors fed to the routing memory, the fast state and the slow state are all set to zero. The variant with $\gamma$ fixed to 1 uses the unscaled router logits.

In the sensitivity test of \cref{fig:supp-ablsens}b, the five compared models are the four with the lowest multi-step mean error on ESCP-HC in Tab.~1 (ReBAL, MTS3, MLP and TD-MPC2) and TD-MPC2-40, the variant of TD-MPC2 trained over the whole 40-step horizon (\cref{sec:supp-baselines}). For context lengths 32 and 128, these five models and \ewm{}'s stage~2 are retrained with that context length, and \ewm{} keeps the base experts, the router and the $\gamma$ selected with the default context of 64 steps; with a context of 128 steps, only the queries whose prediction starts at steps 128, 160 and 192, for which 128 steps of history exist, are evaluated. For the fixed values of $\gamma$, only \ewm{}'s stage~2 is retrained. In all 25 comparisons (5 settings $\times$ 5 compared models), \ewm{}'s multi-step mean error is lower in all 5 seeds, by 8.8\%--40.0\%, and all 95\% intervals of the reductions lie above zero.

\section{Examples}
\label{sec:supp-d}

This section shows examples of the predictions on all six datasets (\crefrange{fig:ex-jshc-r1}{fig:ex-alfa-c}). Each example is drawn as 3D poses (the robot arm or the HalfCheetah, and on ALFA the attitude of the aircraft), as state curves or, in most cases, as both. Some examples are also drawn as paths in space. On the five simulated datasets, the compared model is the strongest one in Tab.~1 of the main paper: ReBAL on JSHC, ESCP-HC and UR5e, and MOLe on Panda and Hopper. On ALFA, it is MOLe, the strongest compared model there (\cref{tab:supp-further}).

\paragraph{How the examples were chosen.} The examples illustrate how the predictions differ, not how often \ewm{} is ahead; Tab.~1 of the main paper gives the results averaged over all test samples. Unlike the examples of Fig.~4 of the main paper (\cref{sec:supp-examples}), these examples were chosen only among the test samples in which \ewm{} is more accurate than the compared model (both from the same training seed) at all shown steps: its mean error up to each step $n$ (over steps 1 to $n$) is lower for every $n$ up to 40, its mean link deviation is lower at steps 10, 20, 30 and 40, and its prediction of at least one state quantity is closer to the truth at all 40 steps. The state curves show only such quantities. Among these samples, the examples of a dataset cover different kinds of change. Where on the robot \ewm{}'s online memories act is shown in \cref{sec:supp-E5-body} for the fast and slow corrections and in \cref{sec:supp-E5-parts} for the error that each online memory removes on each body part.

\paragraph{How to read the figures.} Every example is one test window: each model reads 64 context steps and then predicts the next 40 steps (2\,s) open loop. In the pose renders, the gray robot is the true pose and the colored robot is the predicted pose at steps 10, 20, 30 and 40, with \ewm{} in the top row and the compared model in the bottom row. Each predicted link is colored by its link deviation, \ie, its distance from the true link (one color scale per example), and the number under a frame is the mean over the links, so the closer the colored robot stays to the gray one, the more accurate the prediction. The path figures follow one point (the tool flange at the end of the Panda arm, or the aircraft) through space, with equal scales on all axes: black is the true path, blue is \ewm{} and dashed gray is the compared model, both predictions start at the true point (open circle), and the dots mark the four steps of the renders. The closer a path stays to the black line, the more accurate the prediction. Paths are drawn only for examples in which \ewm{}'s point stays closer to the true point than the compared model's at all 40 steps, in 3D and in every view drawn. The state curves show a few state quantities from 19 steps before the prediction start to step 40 (the HalfCheetah's torso height is measured from its initial height of 0.7\,m), with the mean error up to each step underneath. The numbers in that lower panel are the multi-step mean errors of both models. Here too, a more accurate prediction stays closer to the true curve (black) and has a lower error curve.

\subsection{JSHC}
\label{sec:supp-d-jshc}

\begin{figure*}[tp]
\centering
\sbox{\excapbox}{\parbox{\textwidth}{\small\figurename~S00.\ \input{fig/cap_ex_jshc_render1.tex}}}%
  \setlength{\exmaxheight}{\dimexpr\textheight-\ht\excapbox-\dp\excapbox-\abovecaptionskip-\belowcaptionskip-8pt\relax}%
\includegraphics[width=\textwidth,height=\exmaxheight,keepaspectratio]{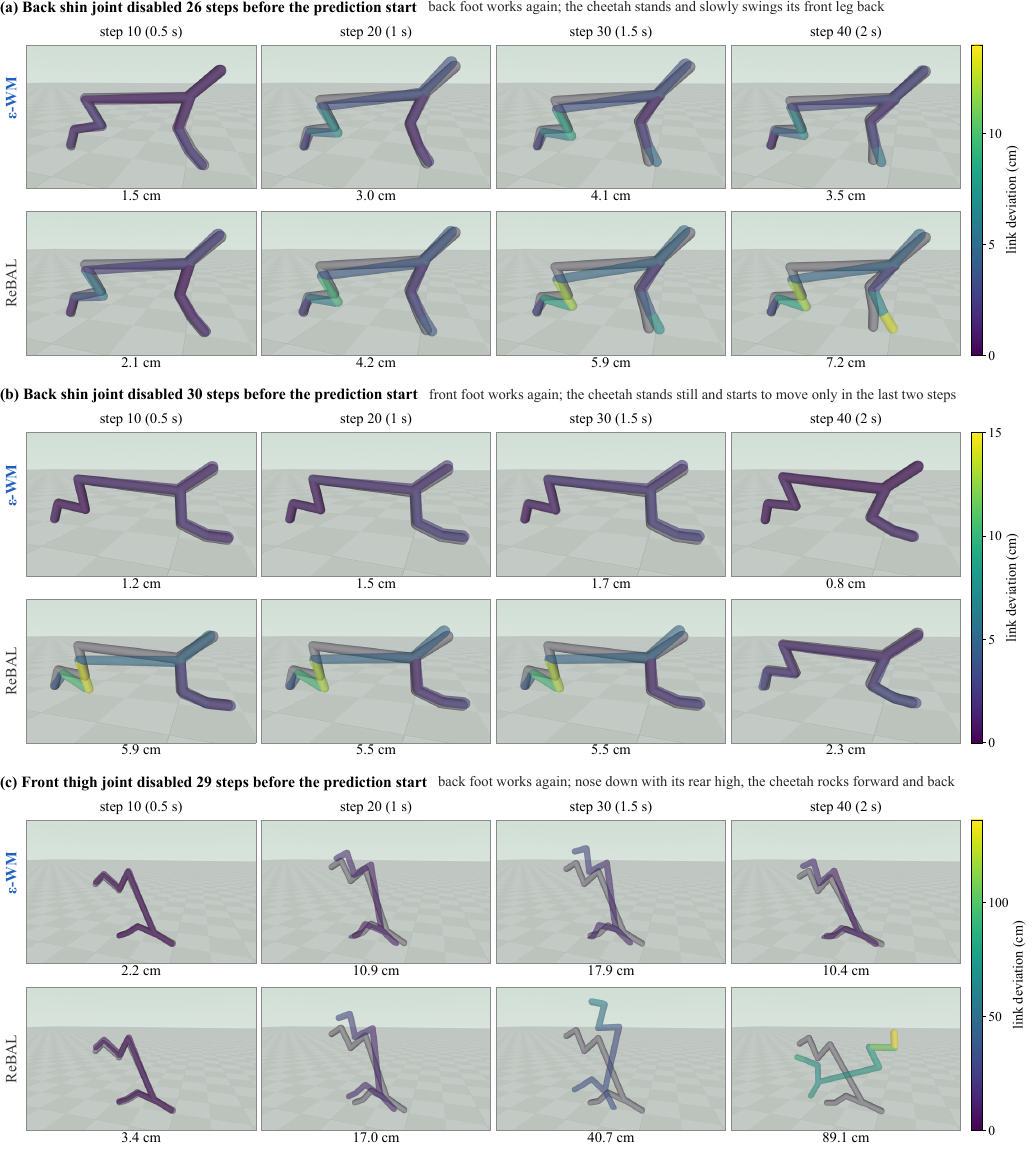}
\caption{\protect\input{fig/cap_ex_jshc_render1.tex}}
\label{fig:ex-jshc-r1}
\end{figure*}

\begin{figure*}[tp]
\centering
\sbox{\excapbox}{\parbox{\textwidth}{\small\figurename~S00.\ \input{fig/cap_ex_jshc_render2.tex}}}%
  \setlength{\exmaxheight}{\dimexpr\textheight-\ht\excapbox-\dp\excapbox-\abovecaptionskip-\belowcaptionskip-8pt\relax}%
\includegraphics[width=\textwidth,height=\exmaxheight,keepaspectratio]{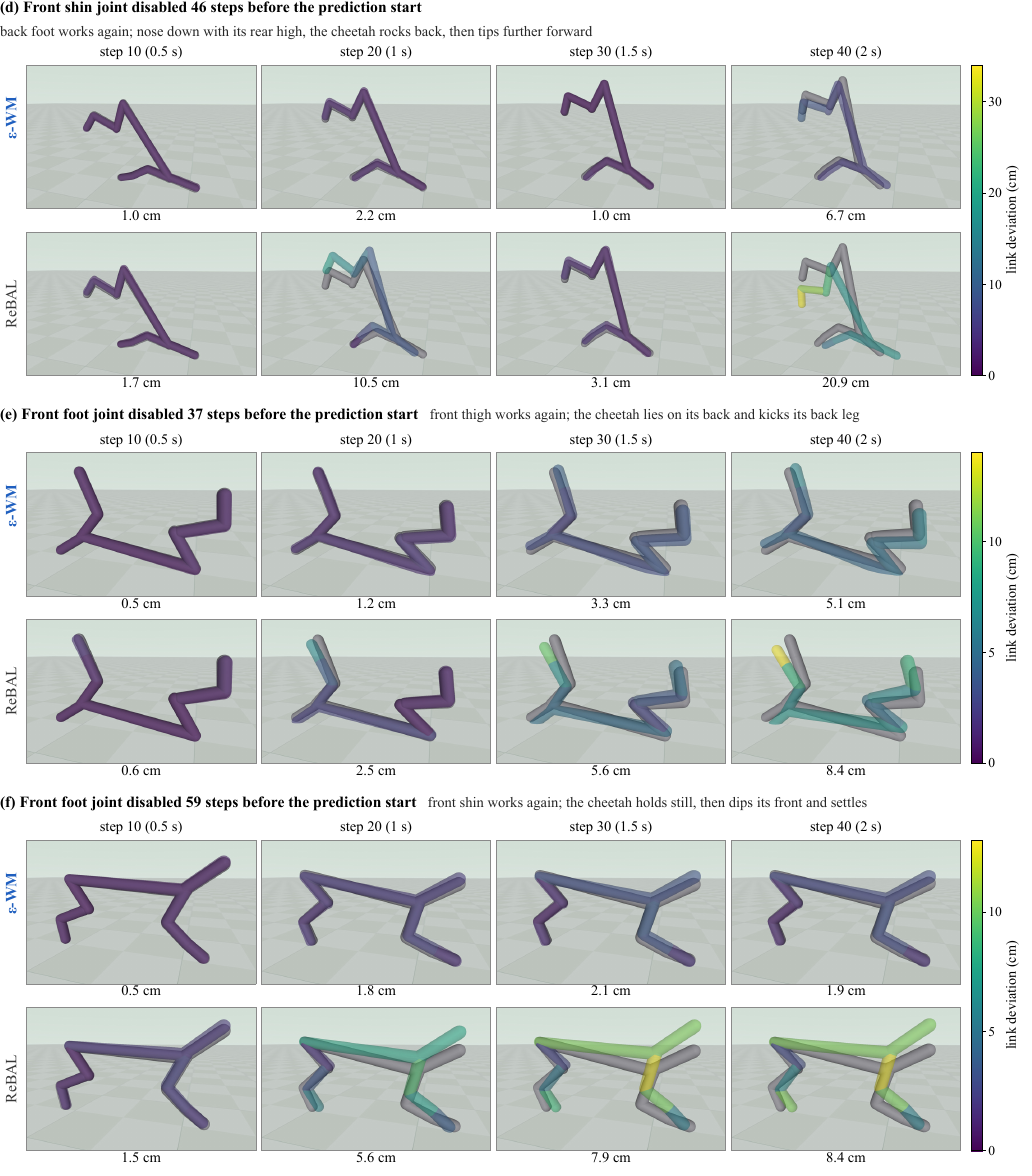}
\caption{\protect\input{fig/cap_ex_jshc_render2.tex}}
\label{fig:ex-jshc-r2}
\end{figure*}

\begin{figure*}[tp]
\centering
\sbox{\excapbox}{\parbox{\textwidth}{\small\figurename~S00.\ \input{fig/cap_ex_jshc_curves.tex}}}%
  \setlength{\exmaxheight}{\dimexpr\textheight-\ht\excapbox-\dp\excapbox-\abovecaptionskip-\belowcaptionskip-8pt\relax}%
\includegraphics[width=\textwidth,height=\exmaxheight,keepaspectratio]{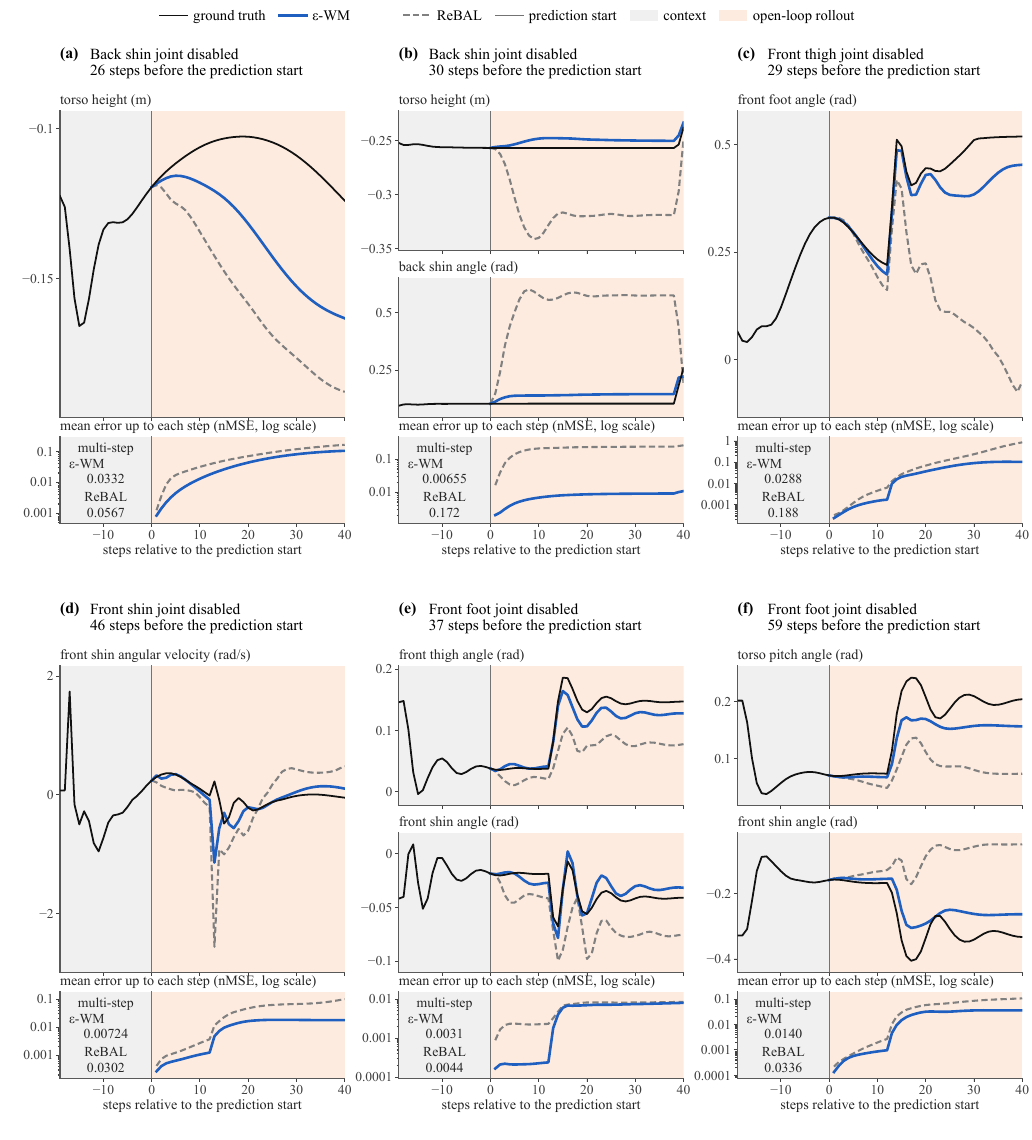}
\caption{\protect\input{fig/cap_ex_jshc_curves.tex}}
\label{fig:ex-jshc-c}
\end{figure*}

JSHC disables a random leg joint of the HalfCheetah every 100 steps, and the disabled (failed) joint produces no torque (\cref{fig:ex-jshc-r1,fig:ex-jshc-r2,fig:ex-jshc-c}). In all six examples, one leg joint is disabled 26 to 59 steps before the prediction start (the back shin in (a) and (b), the front thigh in (c), the front shin in (d) and the front foot in (e) and (f)), and the joint that was disabled until then works again. In (c), ReBAL's predicted pose drifts farther from the truth at every shown step, with a mean link deviation of 40.7\,cm at step 30 and 89.1\,cm at step 40, while \ewm{}'s is 17.9\,cm and 10.4\,cm. In (b), \ewm{}'s deviation is 1.2, 1.5, 1.7 and 0.8\,cm at the four shown steps, whereas ReBAL's is 5.9, 5.5, 5.5 and 2.3\,cm.

\subsection{ESCP-HC}
\label{sec:supp-d-escphc}

\begin{figure*}[tp]
\centering
\sbox{\excapbox}{\parbox{\textwidth}{\small\figurename~S00.\ \input{fig/cap_ex_escphc_render1.tex}}}%
  \setlength{\exmaxheight}{\dimexpr\textheight-\ht\excapbox-\dp\excapbox-\abovecaptionskip-\belowcaptionskip-8pt\relax}%
\includegraphics[width=\textwidth,height=\exmaxheight,keepaspectratio]{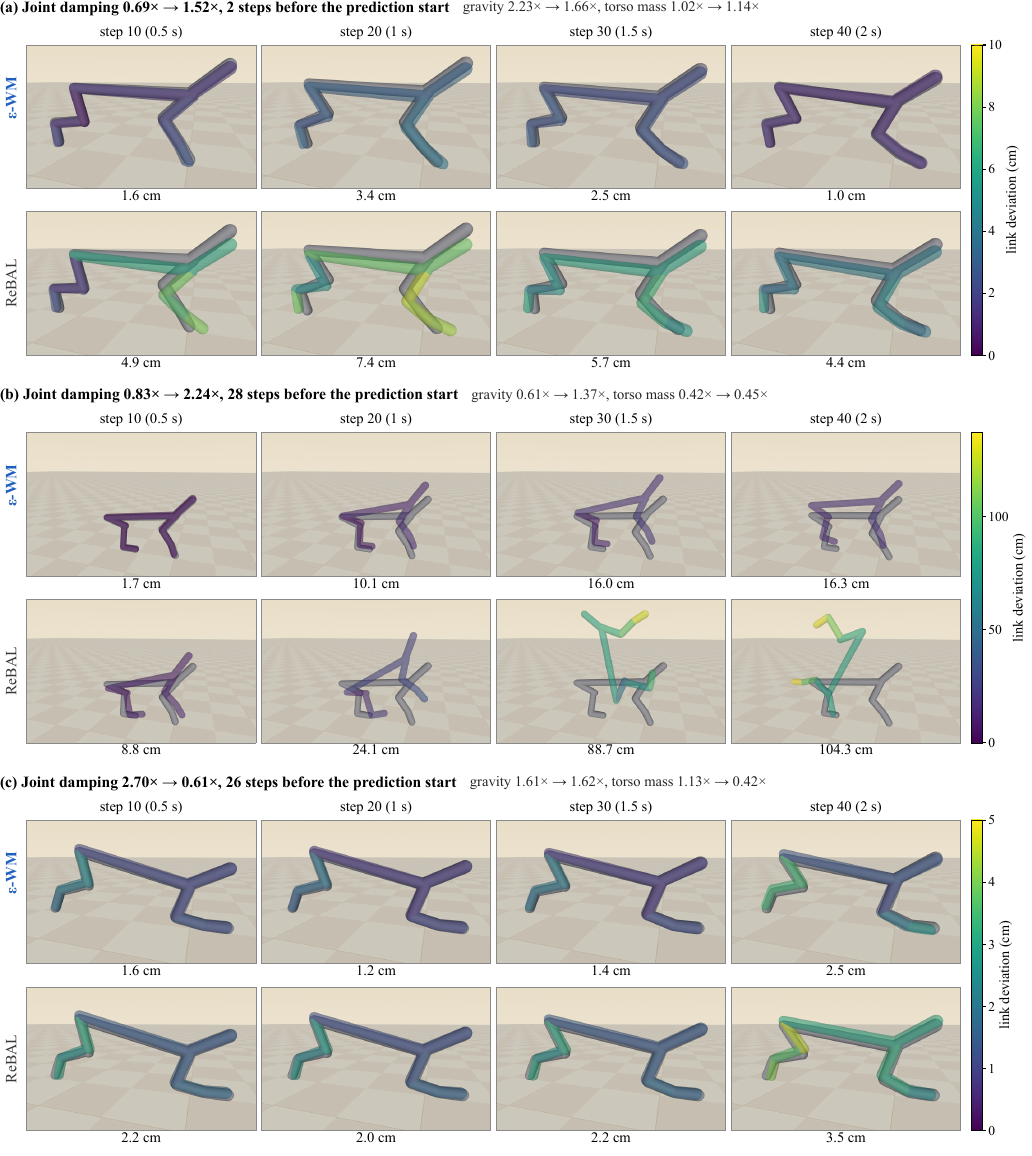}
\caption{\protect\input{fig/cap_ex_escphc_render1.tex}}
\label{fig:ex-escphc-r1}
\end{figure*}

ESCP-HC scales the gravity, the damping of all joints and the torso mass of the HalfCheetah by new random factors every 100 steps (\cref{fig:ex-escphc-r1}). In these examples, gravity, joint damping and torso mass all take new values at the same moment, 2 to 28 steps before the prediction start. The largest of these changes is a rise of joint damping in (a) and (b) and a drop of joint damping in (c). In (a), \ewm{}'s mean link deviation is 1.6, 3.4, 2.5 and 1.0\,cm at the four shown steps, while ReBAL's is 4.9, 7.4, 5.7 and 4.4\,cm.

\subsection{Panda}
\label{sec:supp-d-panda}

\begin{figure*}[tp]
\centering
\sbox{\excapbox}{\parbox{\textwidth}{\small\figurename~S00.\ \input{fig/cap_ex_panda_render1.tex}}}%
  \setlength{\exmaxheight}{\dimexpr\textheight-\ht\excapbox-\dp\excapbox-\abovecaptionskip-\belowcaptionskip-8pt\relax}%
\includegraphics[width=\textwidth,height=\exmaxheight,keepaspectratio]{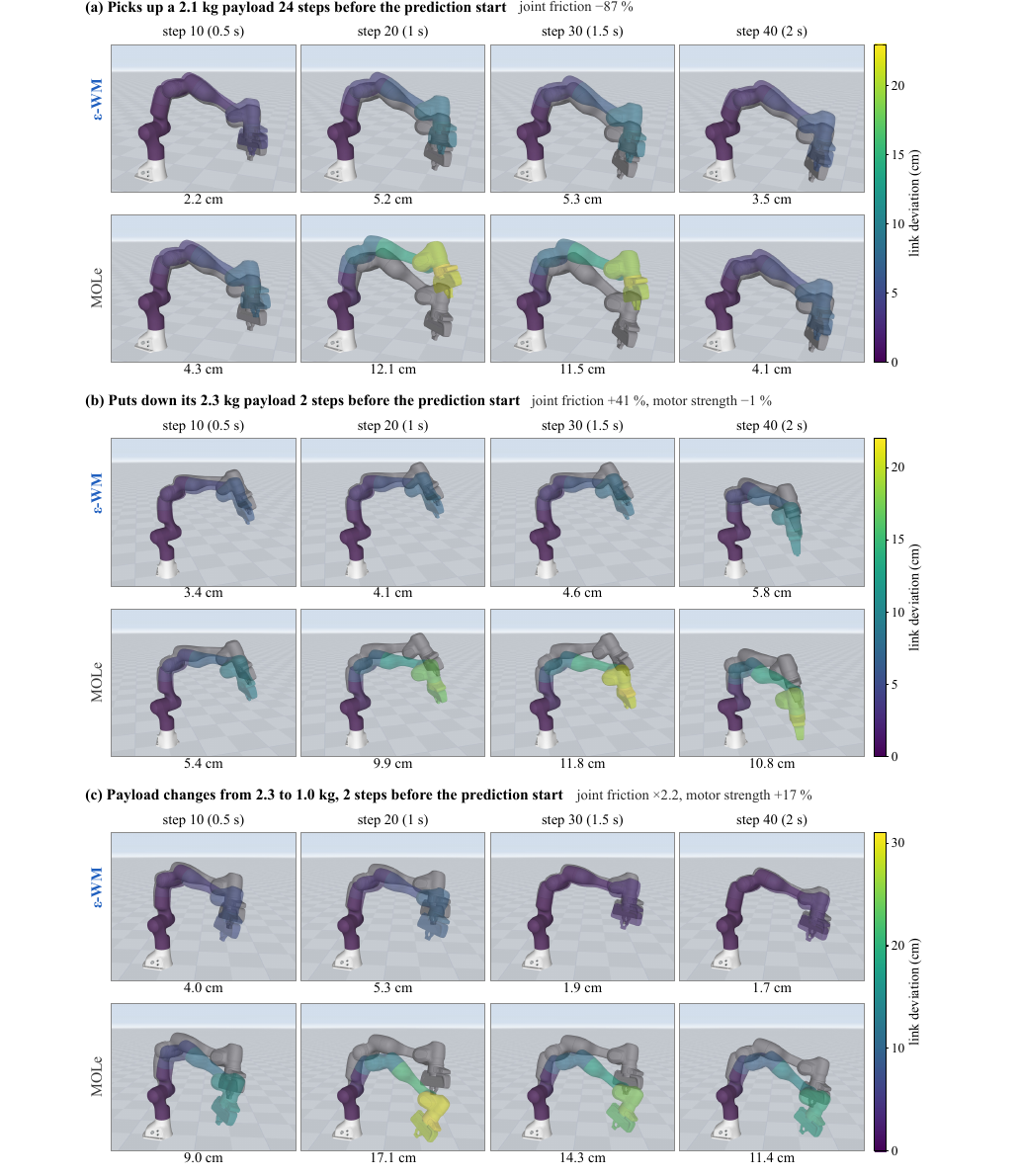}
\caption{\protect\input{fig/cap_ex_panda_render1.tex}}
\label{fig:ex-panda-r1}
\end{figure*}

\begin{figure*}[tp]
\centering
\sbox{\excapbox}{\parbox{\textwidth}{\small\figurename~S00.\ \input{fig/cap_ex_panda_render2.tex}}}%
  \setlength{\exmaxheight}{\dimexpr\textheight-\ht\excapbox-\dp\excapbox-\abovecaptionskip-\belowcaptionskip-8pt\relax}%
\includegraphics[width=\textwidth,height=\exmaxheight,keepaspectratio]{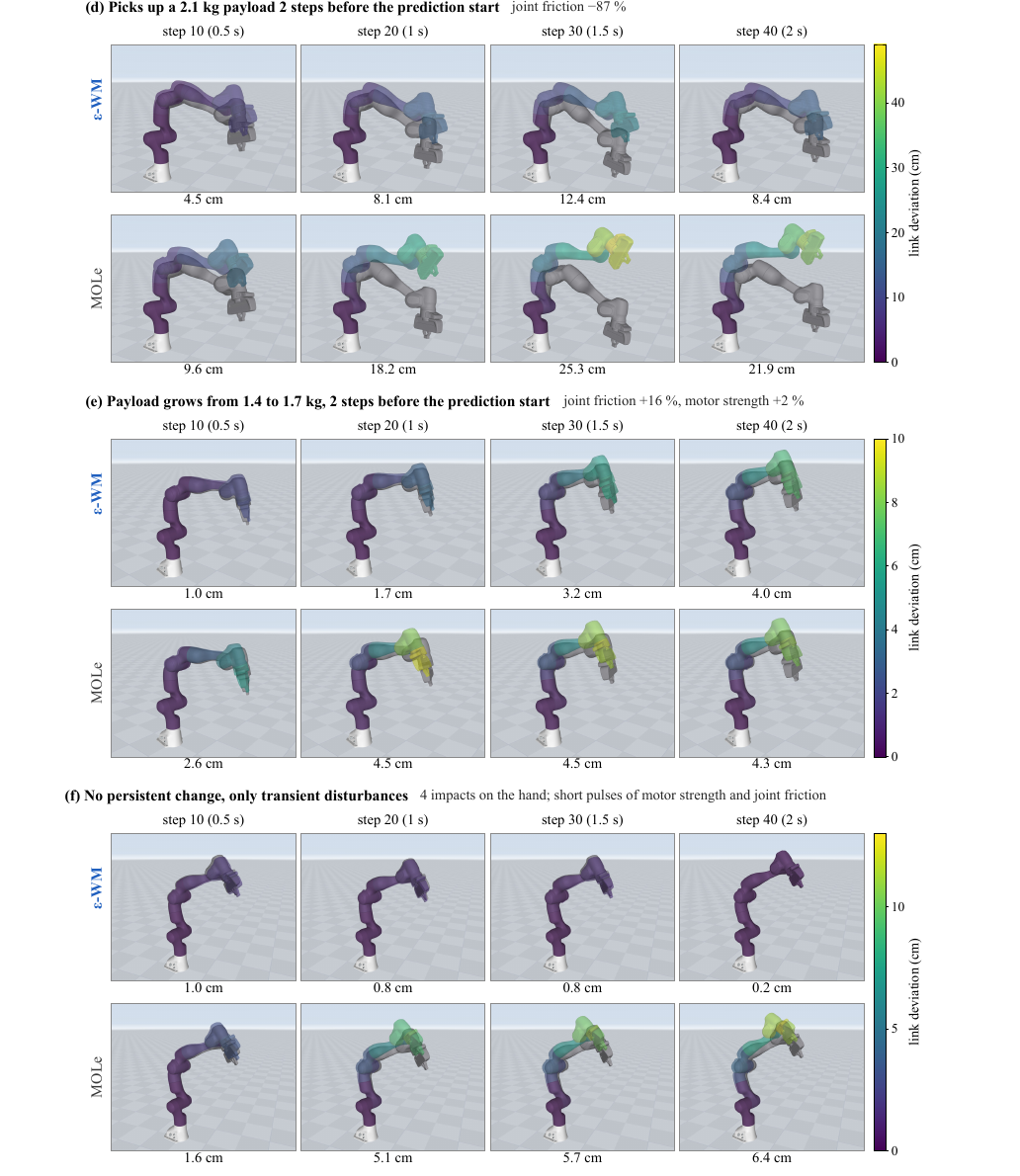}
\caption{\protect\input{fig/cap_ex_panda_render2.tex}}
\label{fig:ex-panda-r2}
\end{figure*}

\begin{figure*}[tp]
\centering
\sbox{\excapbox}{\parbox{\textwidth}{\small\figurename~S00.\ \input{fig/cap_ex_panda_paths.tex}}}%
  \setlength{\exmaxheight}{\dimexpr\textheight-\ht\excapbox-\dp\excapbox-\abovecaptionskip-\belowcaptionskip-8pt\relax}%
\includegraphics[width=\textwidth,height=\exmaxheight,keepaspectratio]{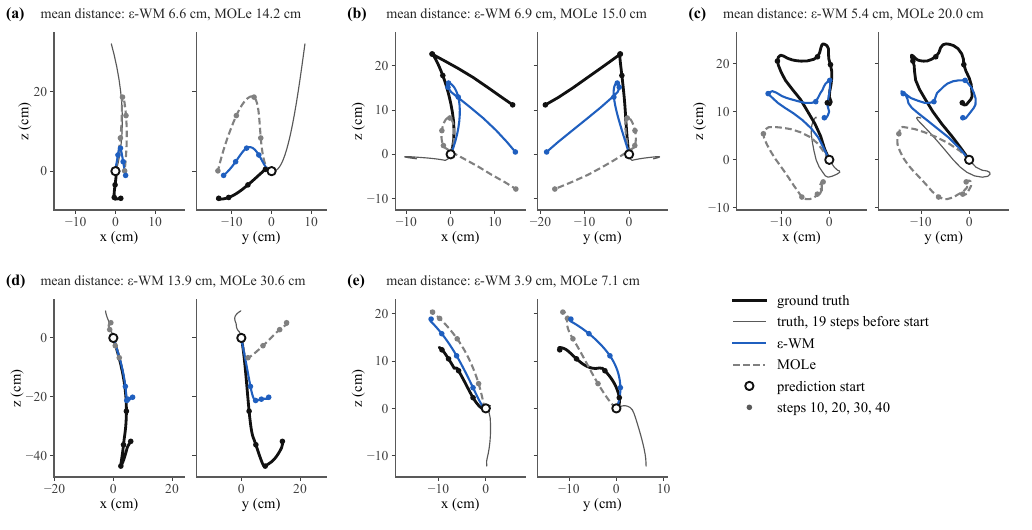}
\caption{\protect\input{fig/cap_ex_panda_paths.tex}}
\label{fig:ex-panda-p}
\end{figure*}

\begin{figure*}[tp]
\centering
\sbox{\excapbox}{\parbox{\textwidth}{\small\figurename~S00.\ \input{fig/cap_ex_panda_curves.tex}}}%
  \setlength{\exmaxheight}{\dimexpr\textheight-\ht\excapbox-\dp\excapbox-\abovecaptionskip-\belowcaptionskip-8pt\relax}%
\includegraphics[width=\textwidth,height=\exmaxheight,keepaspectratio]{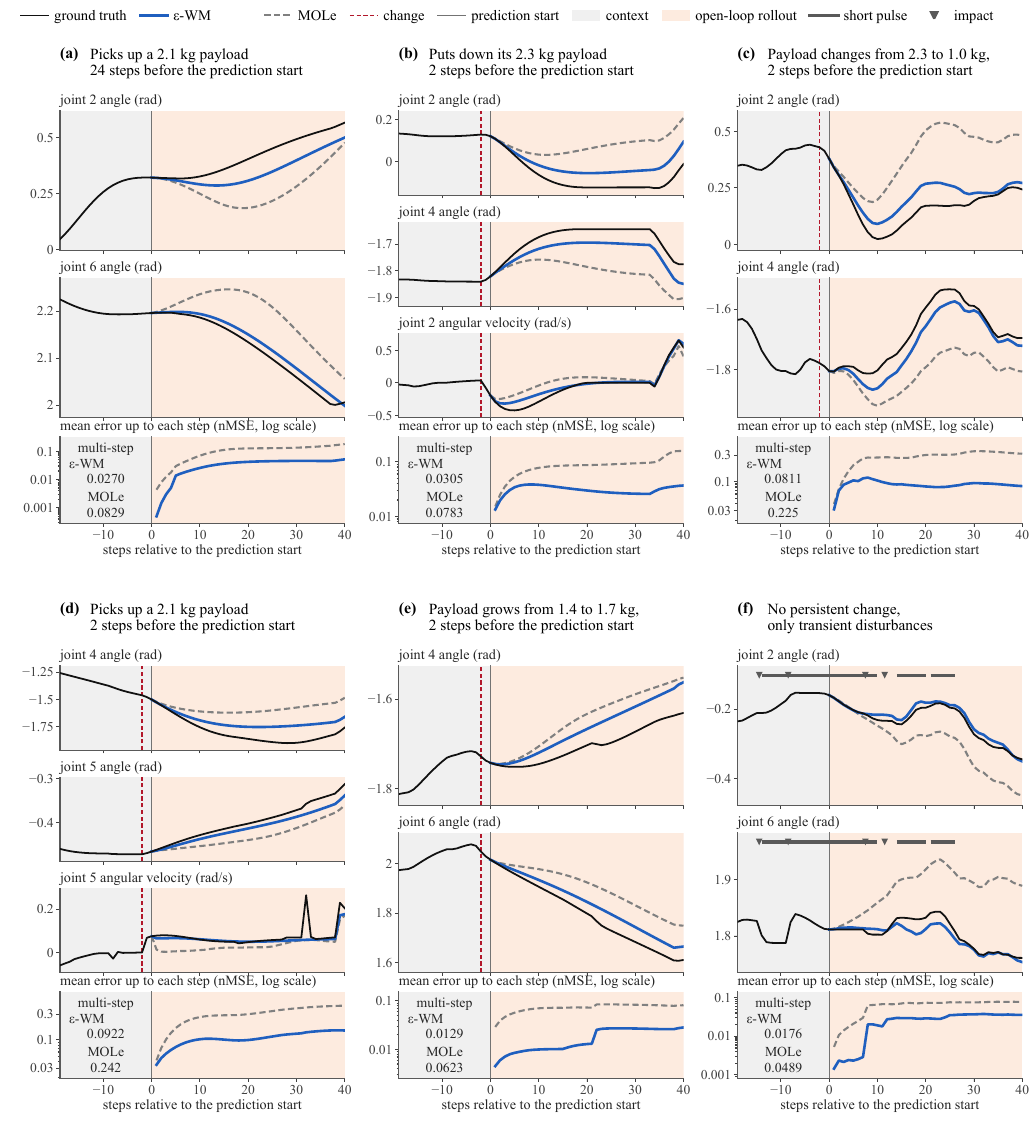}
\caption{\protect\input{fig/cap_ex_panda_curves.tex}}
\label{fig:ex-panda-c}
\end{figure*}

On the Panda arm, the payload, the joint friction and the motor strength change, with short pulses and impacts in between (\cref{fig:ex-panda-r1,fig:ex-panda-r2,fig:ex-panda-c}). In (a)--(d), the arm picks up, puts down or swaps its payload (1.0 to 2.3\,kg), and the joint friction changes with it; (a) and (d) come from the same group of episodes, which share the payload and friction values. In (e), the payload grows from 1.4 to 1.7\,kg as part of a slow drift. In (f), nothing persistent changes, but impacts hit the hand, and the motor strength and joint friction change briefly. In (c), after the 2.3\,kg payload is replaced by 1.0\,kg two steps before the prediction start, MOLe's predicted arm has a mean link deviation of 17.1\,cm at step 20, while \ewm{}'s has 5.3\,cm. In panel (c) of the path figure (\cref{fig:ex-panda-p}), MOLe's tool flange is on average 20.0\,cm from the true one over the 40 steps and \ewm{}'s 5.4\,cm.

\subsection{UR5e}
\label{sec:supp-d-ur5e}

\begin{figure*}[tp]
\centering
\sbox{\excapbox}{\parbox{\textwidth}{\small\figurename~S00.\ \input{fig/cap_ex_ur5e_render1.tex}}}%
  \setlength{\exmaxheight}{\dimexpr\textheight-\ht\excapbox-\dp\excapbox-\abovecaptionskip-\belowcaptionskip-8pt\relax}%
\includegraphics[width=\textwidth,height=\exmaxheight,keepaspectratio]{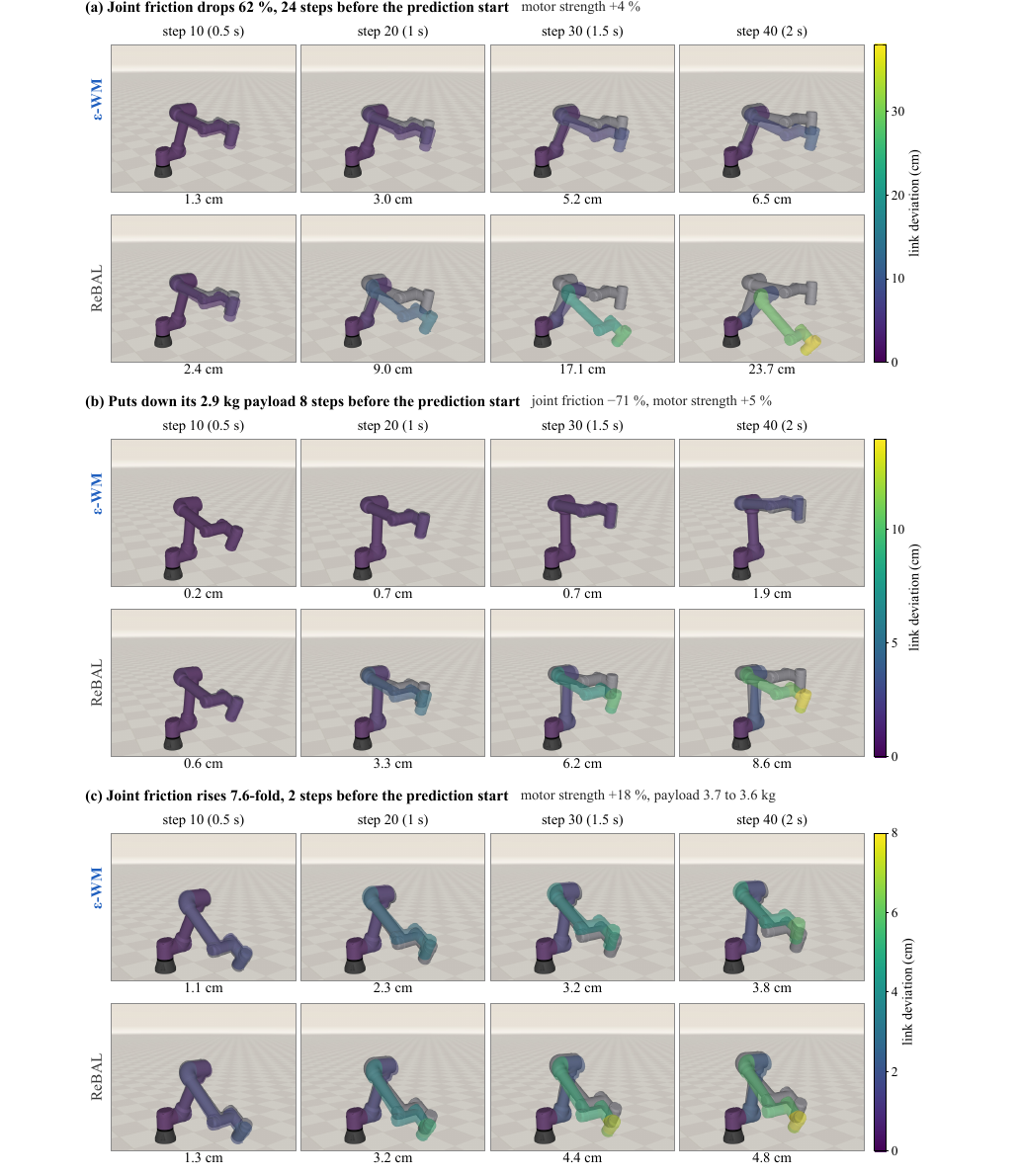}
\caption{\protect\input{fig/cap_ex_ur5e_render1.tex}}
\label{fig:ex-ur5e-r1}
\end{figure*}

\begin{figure*}[tp]
\centering
\sbox{\excapbox}{\parbox{\textwidth}{\small\figurename~S00.\ \input{fig/cap_ex_ur5e_render2.tex}}}%
  \setlength{\exmaxheight}{\dimexpr\textheight-\ht\excapbox-\dp\excapbox-\abovecaptionskip-\belowcaptionskip-8pt\relax}%
\includegraphics[width=\textwidth,height=\exmaxheight,keepaspectratio]{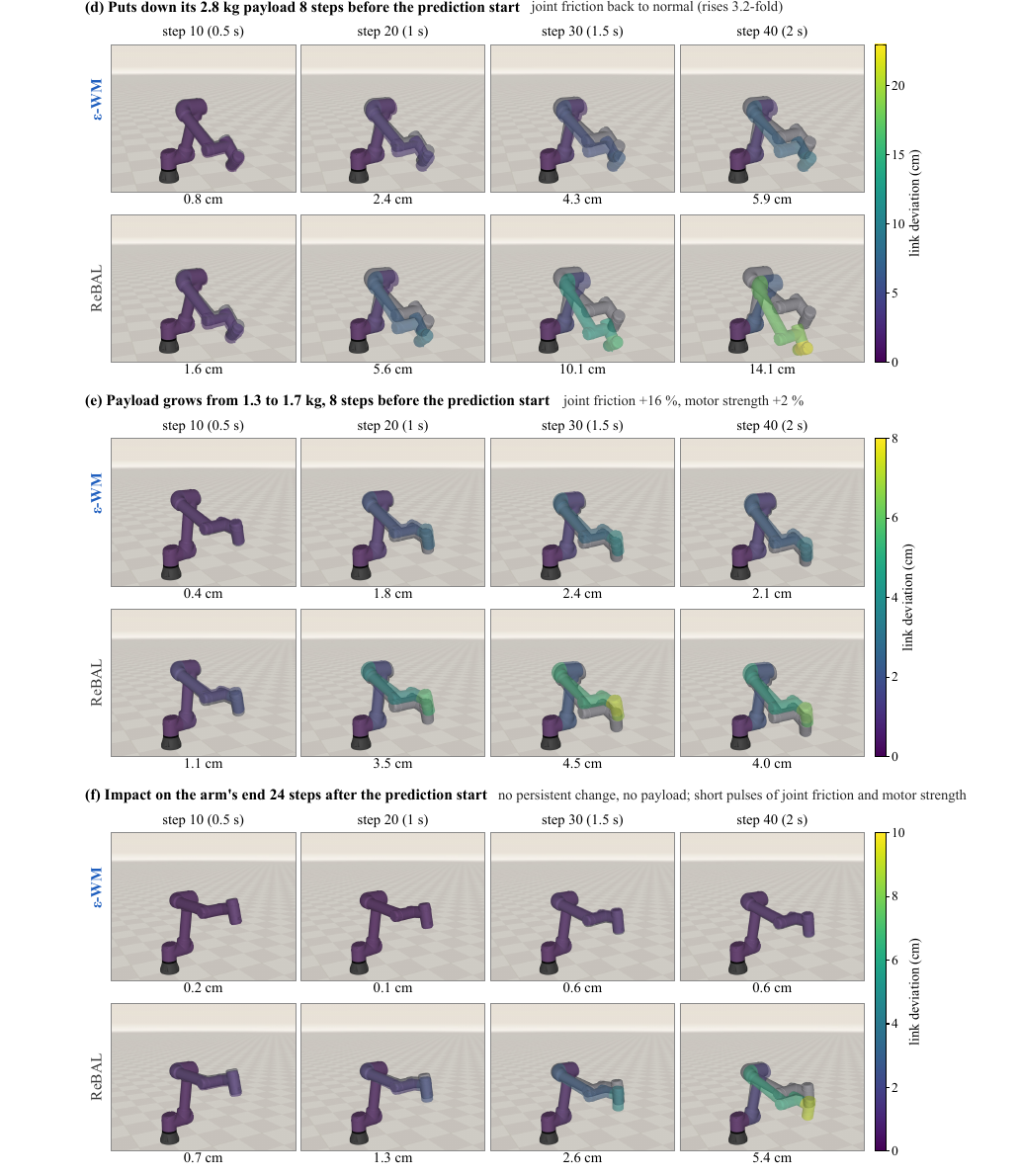}
\caption{\protect\input{fig/cap_ex_ur5e_render2.tex}}
\label{fig:ex-ur5e-r2}
\end{figure*}

\begin{figure*}[tp]
\centering
\sbox{\excapbox}{\parbox{\textwidth}{\small\figurename~S00.\ \input{fig/cap_ex_ur5e_curves.tex}}}%
  \setlength{\exmaxheight}{\dimexpr\textheight-\ht\excapbox-\dp\excapbox-\abovecaptionskip-\belowcaptionskip-8pt\relax}%
\includegraphics[width=\textwidth,height=\exmaxheight,keepaspectratio]{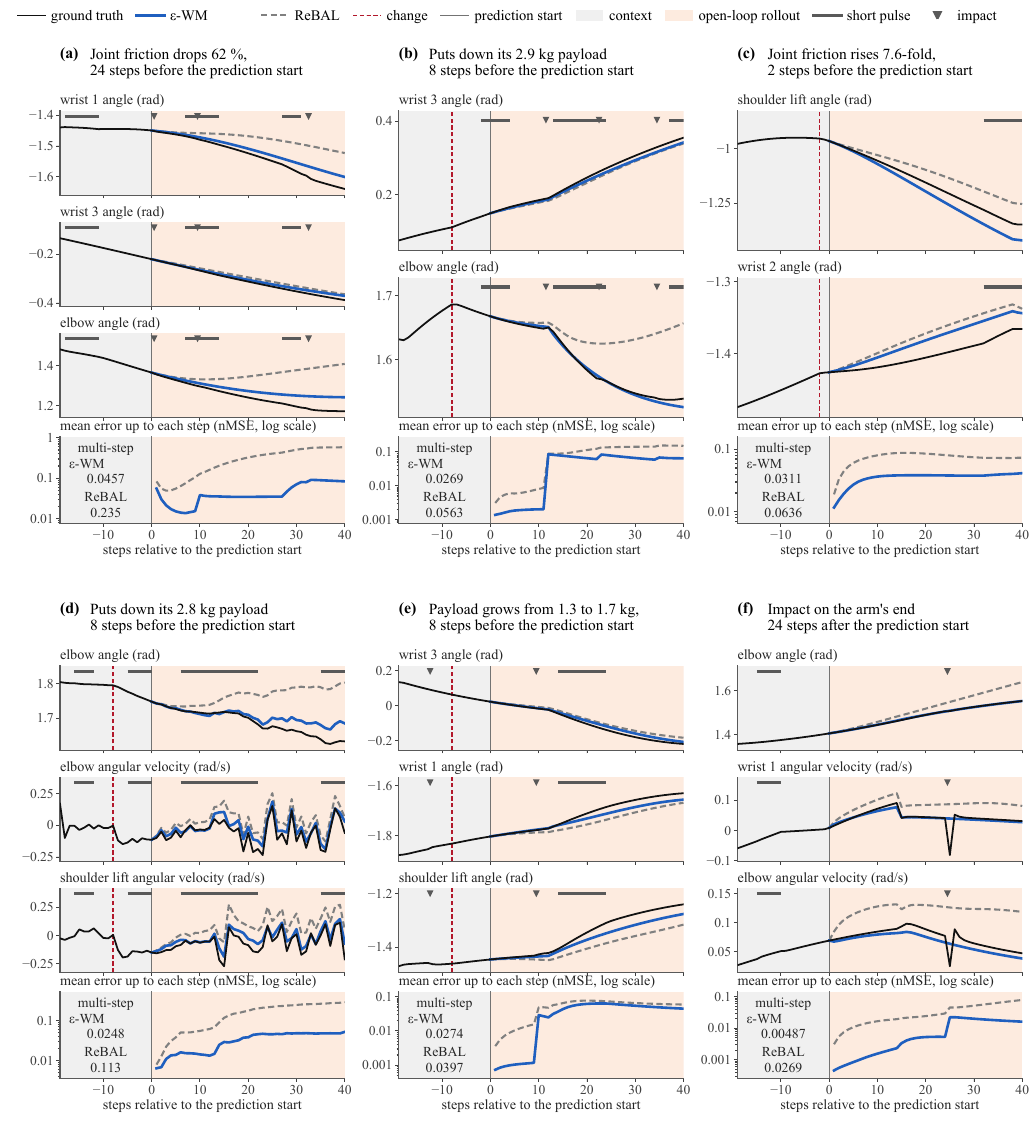}
\caption{\protect\input{fig/cap_ex_ur5e_curves.tex}}
\label{fig:ex-ur5e-c}
\end{figure*}

The UR5e arm has the same kinds of change as Panda, with the payload at the flange (\cref{fig:ex-ur5e-r1,fig:ex-ur5e-r2,fig:ex-ur5e-c}). Joint friction drops by 62\% in (a) and rises 7.6-fold in (c), the arm puts down a 2.9\,kg and a 2.8\,kg payload in (b) and (d), the payload grows from 1.3 to 1.7\,kg in (e), and (f) has only short pulses and impacts, one impact on the end of the arm 24 steps after the prediction start, which neither model can know in advance. In (a), ReBAL's predicted arm drifts from the truth, with a mean link deviation of 9.0, 17.1 and 23.7\,cm at steps 20, 30 and 40, while \ewm{}'s is 3.0, 5.2 and 6.5\,cm.

\subsection{Hopper}
\label{sec:supp-d-hopper}

\begin{figure*}[tp]
\centering
\sbox{\excapbox}{\parbox{\textwidth}{\small\figurename~S00.\ \input{fig/cap_ex_hopper_curves.tex}}}%
  \setlength{\exmaxheight}{\dimexpr\textheight-\ht\excapbox-\dp\excapbox-\abovecaptionskip-\belowcaptionskip-8pt\relax}%
\includegraphics[width=\textwidth,height=\exmaxheight,keepaspectratio]{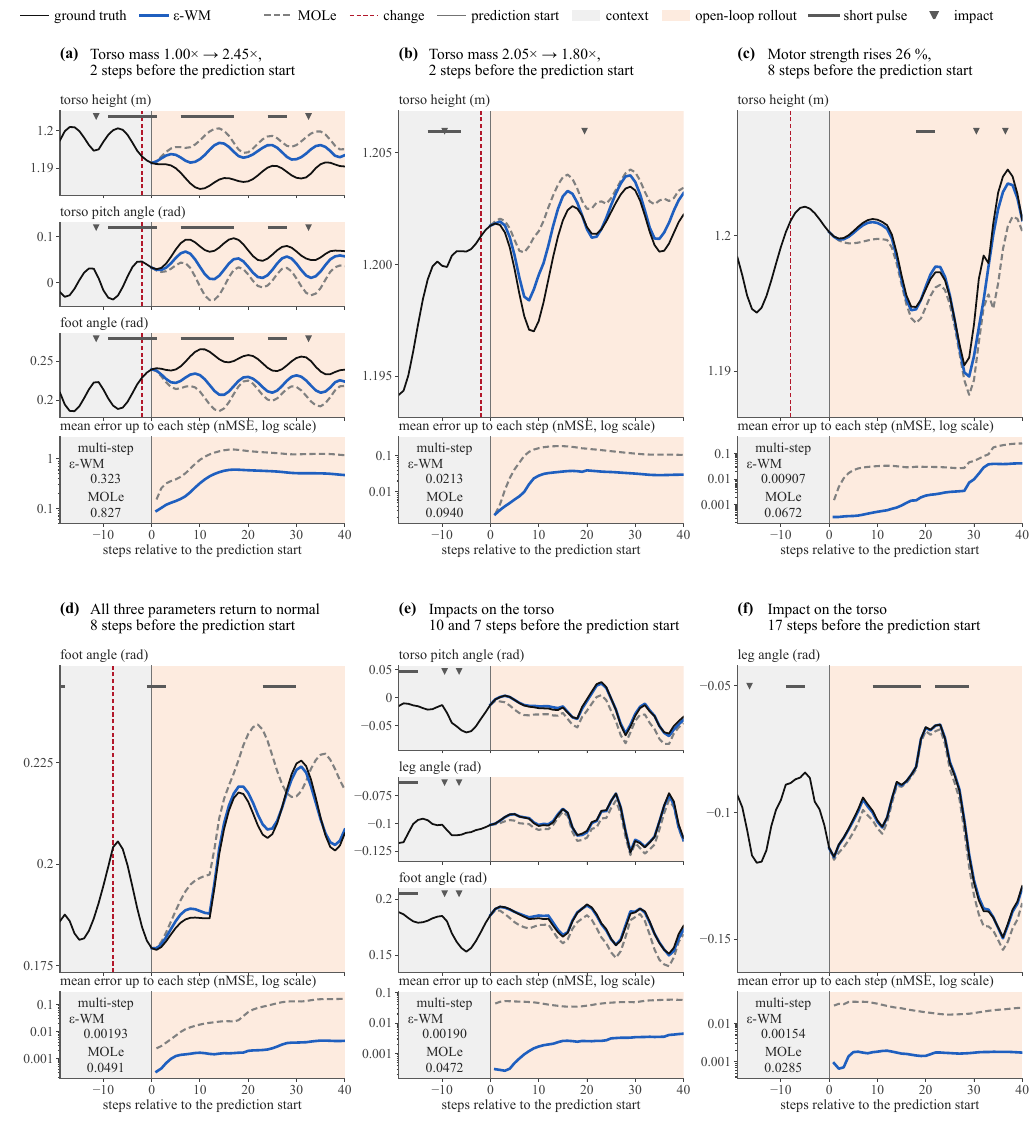}
\caption{\protect\input{fig/cap_ex_hopper_curves.tex}}
\label{fig:ex-hopper-c}
\end{figure*}

On Hopper, the torso mass, the ground friction and the motor strength change, with short pulses and impacts in between (\cref{fig:ex-hopper-c}). The torso mass rises from 1.00 to 2.45 times its normal value in (a) and drops from 2.05 to 1.80 times its normal value in (b), motor strength rises by 26\% in (c), and torso mass, ground friction and motor strength return to normal in (d), each 2 or 8 steps before the prediction start; in (a)--(c), the other two parameters change at the same moment. In (e) and (f) nothing persistent changes, but the torso is hit by impacts 7 to 17 steps before the prediction start. In all six examples, the Hopper stands and sways in place, so we show these examples as state curves only.

\subsection{ALFA}
\label{sec:supp-d-alfa}

\begin{figure*}[tp]
\centering
\sbox{\excapbox}{\parbox{\textwidth}{\small\figurename~S00.\ \input{fig/cap_ex_alfa_render1.tex}}}%
  \setlength{\exmaxheight}{\dimexpr\textheight-\ht\excapbox-\dp\excapbox-\abovecaptionskip-\belowcaptionskip-8pt\relax}%
\includegraphics[width=\textwidth,height=\exmaxheight,keepaspectratio]{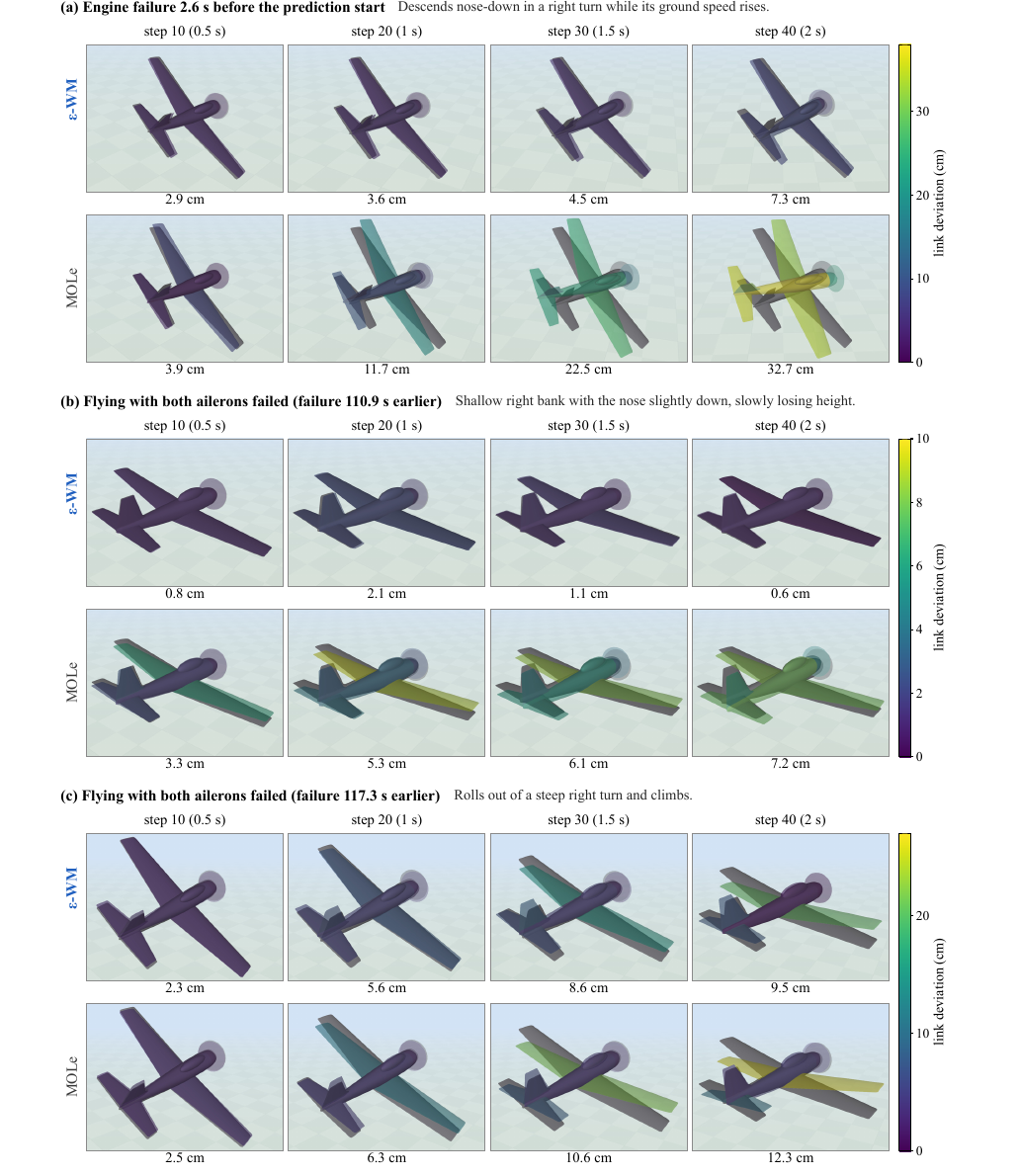}
\caption{\protect\input{fig/cap_ex_alfa_render1.tex}}
\label{fig:ex-alfa-r1}
\end{figure*}

\begin{figure*}[tp]
\centering
\sbox{\excapbox}{\parbox{\textwidth}{\small\figurename~S00.\ \input{fig/cap_ex_alfa_render2.tex}}}%
  \setlength{\exmaxheight}{\dimexpr\textheight-\ht\excapbox-\dp\excapbox-\abovecaptionskip-\belowcaptionskip-8pt\relax}%
\includegraphics[width=\textwidth,height=\exmaxheight,keepaspectratio]{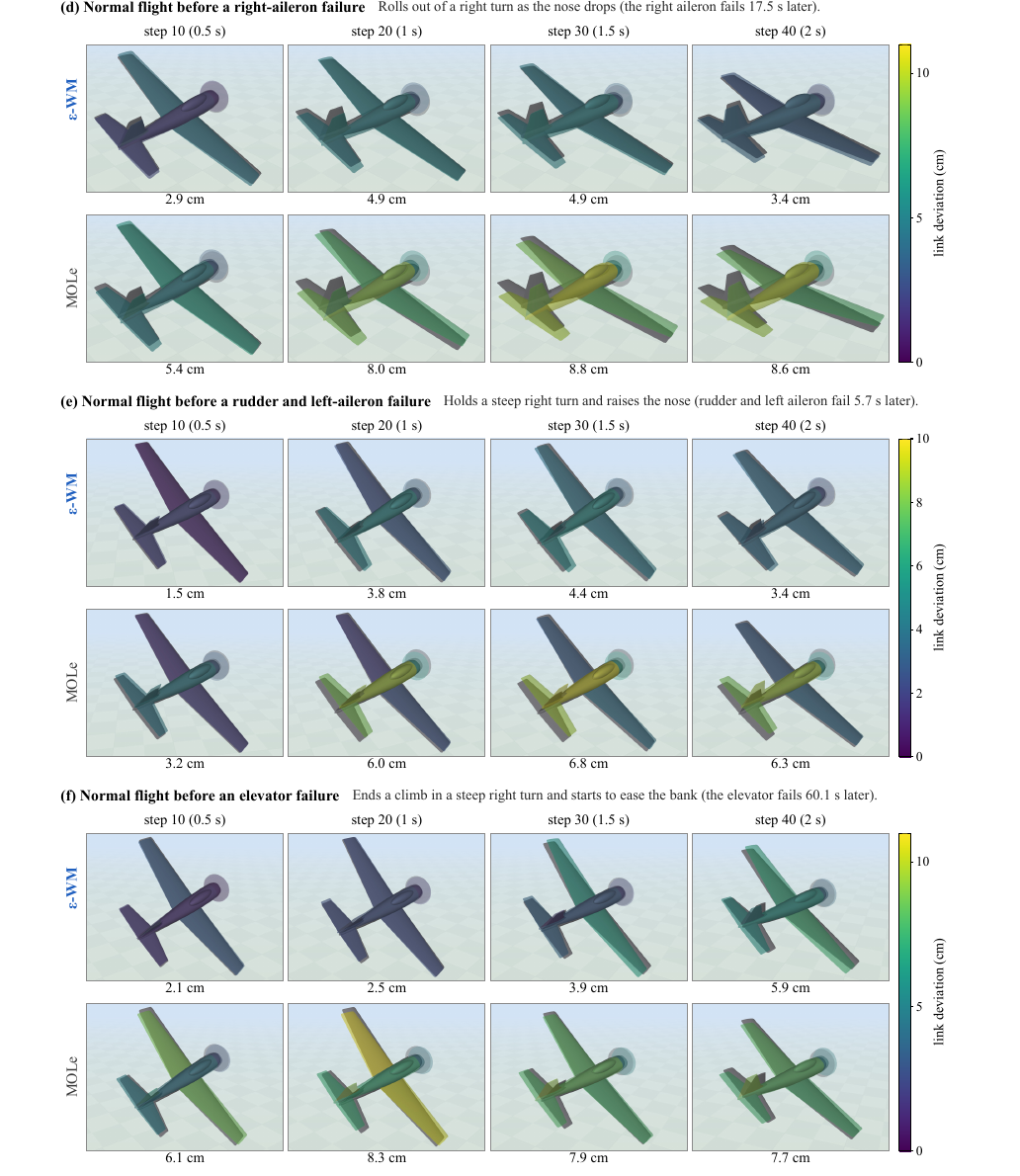}
\caption{\protect\input{fig/cap_ex_alfa_render2.tex}}
\label{fig:ex-alfa-r2}
\end{figure*}

\begin{figure*}[tp]
\centering
\sbox{\excapbox}{\parbox{\textwidth}{\small\figurename~S00.\ \input{fig/cap_ex_alfa_paths.tex}}}%
  \setlength{\exmaxheight}{\dimexpr\textheight-\ht\excapbox-\dp\excapbox-\abovecaptionskip-\belowcaptionskip-8pt\relax}%
\includegraphics[width=\textwidth,height=\exmaxheight,keepaspectratio]{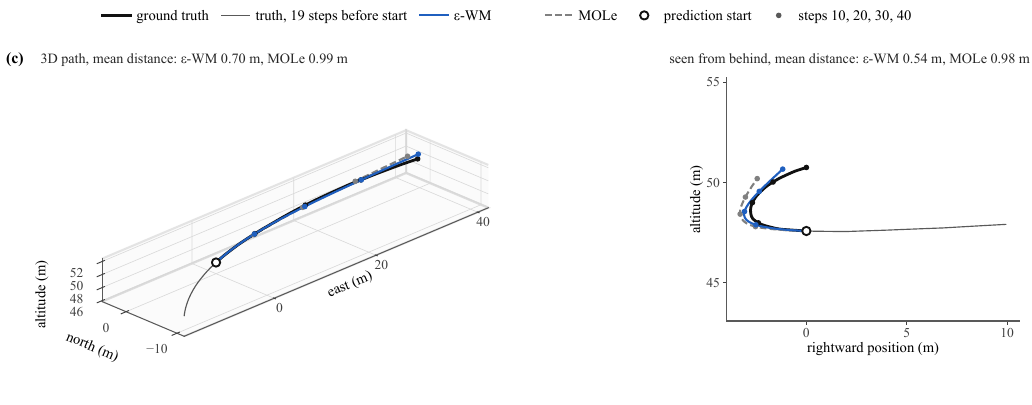}
\caption{\protect\input{fig/cap_ex_alfa_paths.tex}}
\label{fig:ex-alfa-p}
\end{figure*}

\begin{figure*}[tp]
\centering
\sbox{\excapbox}{\parbox{\textwidth}{\small\figurename~S00.\ \input{fig/cap_ex_alfa_curves.tex}}}%
  \setlength{\exmaxheight}{\dimexpr\textheight-\ht\excapbox-\dp\excapbox-\abovecaptionskip-\belowcaptionskip-8pt\relax}%
\includegraphics[width=\textwidth,height=\exmaxheight,keepaspectratio]{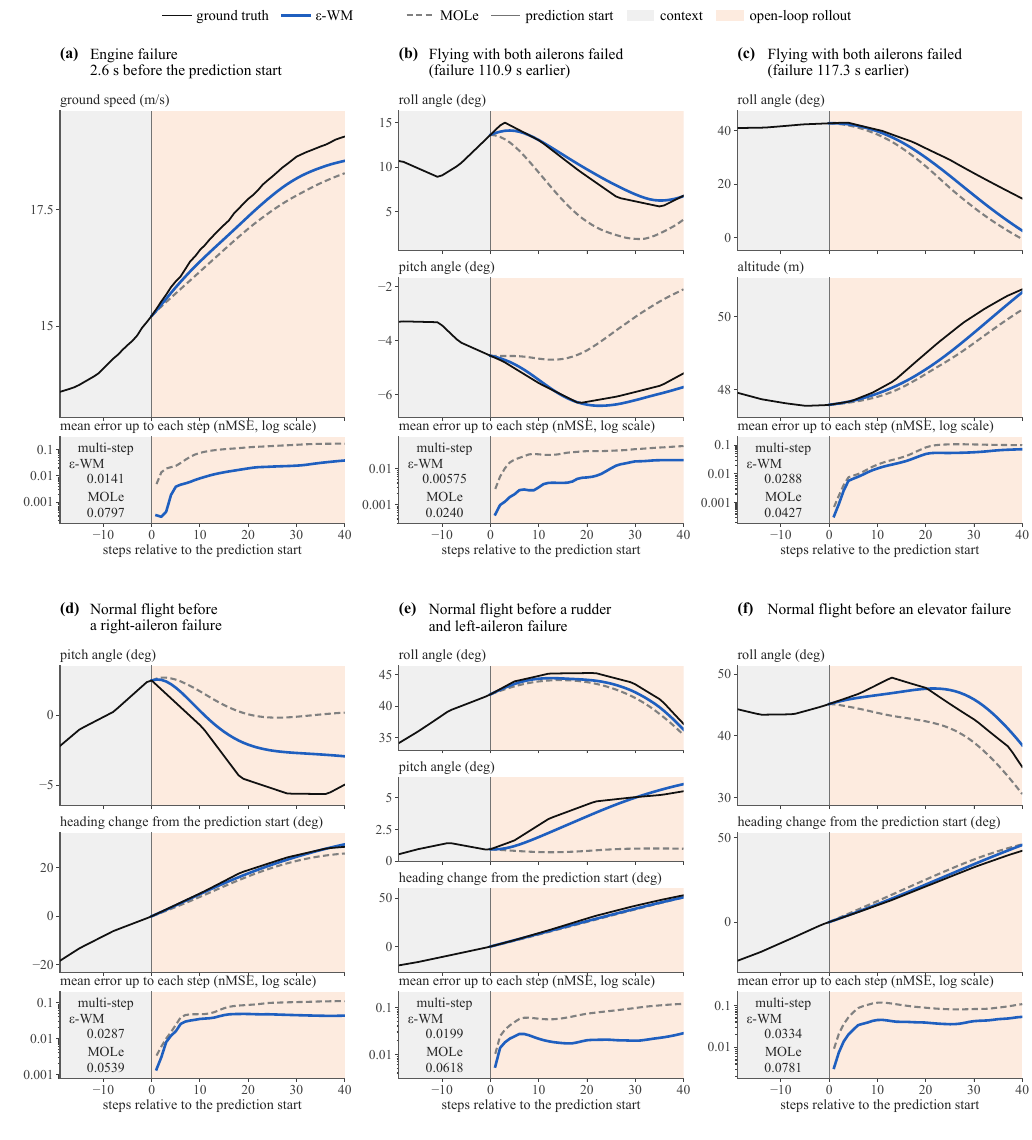}
\caption{\protect\input{fig/cap_ex_alfa_curves.tex}}
\label{fig:ex-alfa-c}
\end{figure*}

ALFA contains real flights in which the engine or one or more control surfaces fail at a labeled time. The renders draw the aircraft at a fixed position and thus show only its attitude (roll, pitch and heading). Each airframe part (nose, fuselage, wings and tail surfaces) plays the role of a link: it is colored by its deviation, and the number under a frame is the mean over the parts (\cref{fig:ex-alfa-r1,fig:ex-alfa-r2,fig:ex-alfa-c}). In (a), the prediction starts 2.6\,s after an engine failure, in (b) and (c) 110.9\,s and 117.3\,s after both ailerons failed, and (d)--(f) show normal flight, with no failure in the context or in the horizon; their titles name the failure that comes later in the same flight. In (a), MOLe's predicted attitude moves farther from the truth at each shown step (3.9, 11.7, 22.5 and 32.7\,cm), while \ewm{}'s stays between 2.9 and 7.3\,cm. In the path figure (\cref{fig:ex-alfa-p}) of (c), the aircraft covers 39.8\,m in the 40 steps, and the predicted positions are on average 0.70\,m (\ewm{}) and 0.99\,m (MOLe) from the true position.

\clearpage %

\section{More analyses}
\label{sec:supp-E}\label{sec:supp-e}%

This section adds further analyses to Sec.~4 of the main paper. These analyses use the test sets of the main paper and, unless stated otherwise, the protocols of \cref{sec:supp-c}. Unless a figure says otherwise, the error is the main metric, the multi-step mean error. A window is one test query: a 64-step context and the 40 steps that follow. When we compare \ewm{} with a model $B$, the \emph{reduction} $100\times(B-\text{\ewm{}})/B$ is the negative of the relative difference of \cref{sec:supp-metrics}, so a positive value means that \ewm{}'s error is lower. Unless stated otherwise, all intervals are 95\% bootstrap intervals, computed as in \cref{sec:supp-metrics}. Analyses over several horizons use the mean error over steps 1 to $H$ ($n$ in the main paper).

\paragraph{Main results.} On each of the five simulated datasets, \ewm{}'s multi-step mean error in Tab.~1 of the main paper is lower than that of each of the ten compared models in all 5 seeds, and the 95\% intervals of all 50 relative differences lie below zero.

\paragraph{Further tests.} \Cref{tab:supp-further} gives \ewm{}'s margin over each of the ten compared models in the further tests of Secs.~4.2 and 4.5 of the main paper. Beyond the training range (\cref{sec:supp-ood}), \ewm{}'s multi-step mean error on the 916 test windows is 5.9\% (ReBAL) to more than 99.9\% (MOLe) lower. In windows without a change (\cref{sec:supp-phase}), it is lower than that of every compared model on each of the five datasets, by at least 6.6\% (ReBAL on ESCP-HC), and in such windows with transient disturbances on Panda, UR5e and Hopper, by at least 28.1\% (MOLe on Hopper). On ALFA's real flights (\cref{sec:supp-alfa}), it is 13.1\% (MOLe) to 99.8\% (GrBAL) lower, and lower than that of each compared model in all 4 seeds. In the control task (\cref{sec:supp-control}), \ewm{}'s mean return of 93.1 exceeds those of the compared models by 13.8 (MOLe) to 105.0 (MTS3).

\begin{table}[t]
\centering
\footnotesize
\setlength{\tabcolsep}{3pt}
\begin{tabular}{lccccc}
\toprule
Compared model & \makecell{Beyond\\range} & \makecell{No\\change} & \makecell{With\\disturb.} & \makecell{Real\\flights} & \makecell{Control\\(return)} \\
\midrule
RSSM & 31.1 & 30.7 & 65.3 & 85.9 & +57.3 \\
TD-MPC2 & 14.5 & 13.0 & 57.3 & 79.8 & +33.2 \\
TD-MPC2-40 & 27.2 & 35.7 & 58.9 & 90.6 & +40.9 \\
MLP & 13.9 & 19.0 & 42.0 & 17.3 & +25.2 \\
\midrule
GrBAL & 60.8 & 40.4 & 40.1 & 99.8 & +34.9 \\
ReBAL & \underline{5.9} & \underline{6.6} & 35.0 & 27.5 & +21.9 \\
MOLe & $>$99.9 & 28.5 & \underline{28.1} & \underline{13.1} & \underline{+13.8} \\
MBCD & 26.8 & 38.9 & 57.8 & 52.2 & +95.9 \\
GPMM & 48.4 & 55.0 & 61.8 & 40.8 & +101.8 \\
\midrule
MTS3 & 9.6 & 7.4 & 33.1 & 59.4 & +105.0 \\
\midrule
Outperformed & 10 of 10 & 10 of 10 & 10 of 10 & 10 of 10 & 10 of 10 \\
\bottomrule
\end{tabular}
\caption{Further tests of Secs.~4.2 and 4.5 of the main paper: \ewm{}'s margin over each compared model. The first four columns give the reduction of the multi-step mean error in \%: beyond the training range on ESCP-HC (916 windows), in windows without a change (smallest over the five datasets), in such windows with transient disturbances (smallest over Panda, UR5e and Hopper) and on ALFA's real flights. The last column gives the gain in mean return in the control task. Positive values mean that \ewm{} is better. The smallest margin in each column is underlined. On ALFA, \ewm{}'s multi-step mean error is lower than that of each compared model in all 4 seeds. The margins compare means over seeds; their 95\% intervals lie above zero except beyond the training range for ReBAL and MTS3, in windows without a change for ReBAL and MTS3 (on ESCP-HC, 25 windows), on ALFA for MOLe, and in the control task for six of the ten compared models.}
\label{tab:supp-further}
\end{table}

\paragraph{Per-step errors.} Figure~3 of the main paper shows the per-step errors on the five simulated datasets. Comparing \ewm{} with each compared model at each predicted step of each dataset gives 2,000 comparisons (5 datasets $\times$ 10 models $\times$ 40 steps). \ewm{}'s per-step error is lower in 1,999 of these comparisons. The only exception is step 1 on JSHC, where ReBAL's error is lower. In 1,987 of the 2,000 comparisons, the 95\% interval of the reduction lies above zero. Thus, from step 2 on, \ewm{} is the most accurate model at every step on every simulated dataset.

\subsection{Situations of change}
\label{sec:supp-E1}

\begin{figure*}[t]
\centering
\includegraphics[width=\textwidth]{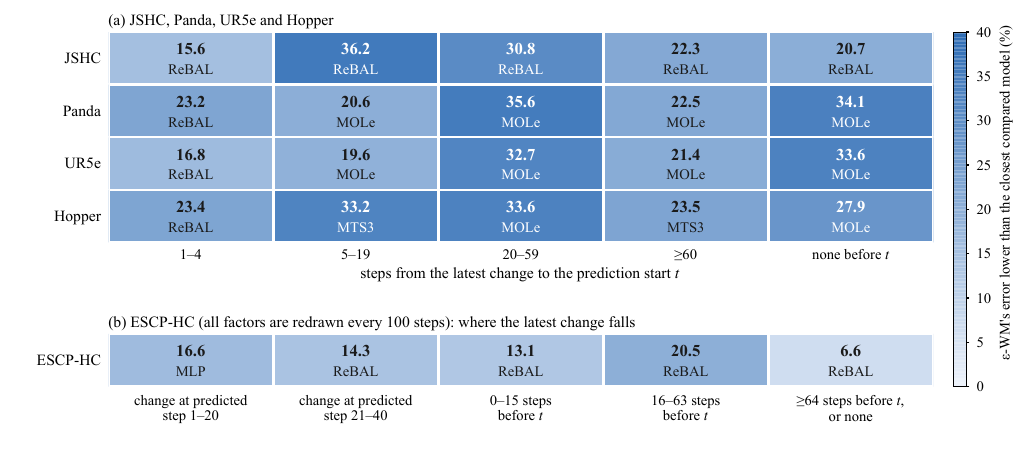}
\caption{\ewm{}'s lead in each change phase. Each cell shows the reduction of \ewm{}'s multi-step mean error relative to the closest of the ten compared models in one phase, and names that model. (a)~Phases by the number of steps from the latest change to the prediction start $t$, as in Fig.~1b of the main paper. (b)~ESCP-HC redraws all its factors every 100 steps and is split by its own rule (\cref{sec:supp-phase}). In its two left phases, the change falls inside the 40 predicted steps. All 25 cells are positive (6.6\% to 36.2\%), so \ewm{} has the lower error in all 250 dataset--model--phase comparisons. In 247 of these comparisons, the 95\% interval lies above zero. The other three fall in the two cells with the fewest windows: JSHC 1--4 steps against ReBAL (33 windows) and the last ESCP-HC cell against ReBAL and MTS3 (25 windows). Each cell holds 25 to 672 windows.}
\label{fig:E-phases}
\end{figure*}

\paragraph{When the latest change happened.} Figure~1b of the main paper shows the change phases of JSHC, Panda, UR5e and Hopper. It leaves out ESCP-HC, whose phases follow a rule of their own (\cref{sec:supp-phase}). \Cref{fig:E-phases} adds ESCP-HC. In each phase, we compare \ewm{} with all ten compared models and show the reduction relative to the closest one. All 25 cells are positive, from 6.6\% to 36.2\%, so \ewm{}'s error is lower than that of every compared model in every phase, and in 247 of the 250 comparisons (25 cells $\times$ 10 compared models) the 95\% interval lies above zero. On ESCP-HC, \ewm{}'s error is also lower when the change falls inside the 40 predicted steps (by 16.6\% and 14.3\%). Thus, in simulation, \ewm{} leads regardless of when the latest change happened.

\begin{figure*}[t]
\centering
\includegraphics[width=\textwidth]{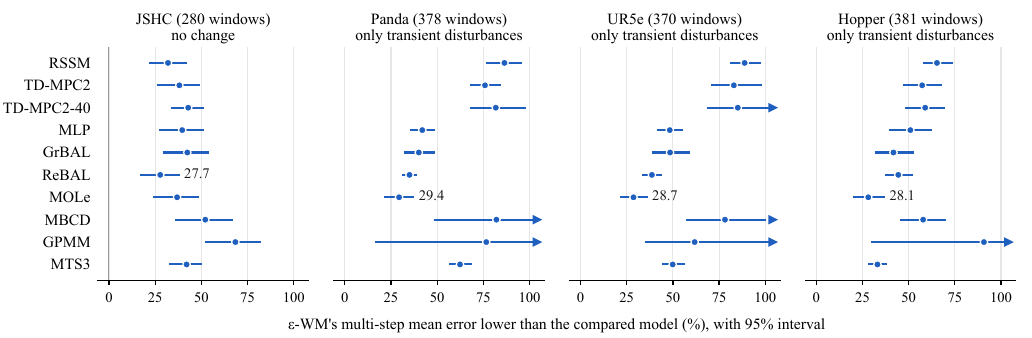}
\caption{Windows without a persistent change. Each dot shows the reduction of \ewm{}'s multi-step mean error relative to one compared model, with its 95\% interval. JSHC: windows with no change in the context or the horizon (280 windows). Panda, UR5e and Hopper: windows with no change, but with at least one impact or pulse of friction or motor strength in the last 24 context steps or in the horizon (378, 370 and 381 of the 384 windows without a change). \ewm{}'s error is lower than that of all ten compared models in every panel, by at least 27.7\% on JSHC and 28.1\% on the other three datasets, and every interval lies above zero. The printed number marks the smallest reduction in each panel. Intervals that extend past 100\% are cut at the edge (arrows). This can happen because each difference is divided by the compared model's mean error, which stays fixed in the bootstrap. The longest intervals in the figure belong to MBCD and GPMM, whose multi-step mean errors on Panda, UR5e and Hopper also vary most across seeds (Tab.~1 of the main paper).}
\label{fig:E-nochange}
\end{figure*}

\paragraph{Windows without a persistent change.} Adaptation also requires stability (Sec.~1 of the main paper): a transient disturbance should not degrade a good prediction. On JSHC, we take the 280 windows with no change in the context or the horizon. On Panda, UR5e and Hopper, almost all such windows contain transient disturbances, namely an impact or a pulse of friction or motor strength in the last 24 context steps or in the horizon (378, 370 and 381 of 384), and we take these. \ewm{}'s error is lower than that of all ten compared models on all four datasets (\cref{fig:E-nochange}). It is at least 27.7\% lower on JSHC and at least 28.1\% lower on the three new datasets, and every interval lies above zero. ESCP-HC redraws its factors every 100 steps and has only 25 windows without a change. These windows form the last cell of \cref{fig:E-phases}b, where \ewm{}'s error is 6.6\% below that of the closest compared model, ReBAL, the margin given for windows without a change in \cref{tab:supp-further}. Thus, on JSHC, Panda, UR5e and Hopper, \ewm{} stays the most accurate of the eleven models when nothing persistent changes, and on the 25 such windows of ESCP-HC its multi-step mean error is also the lowest.

\begin{figure*}[t]
\centering
\includegraphics[width=\textwidth]{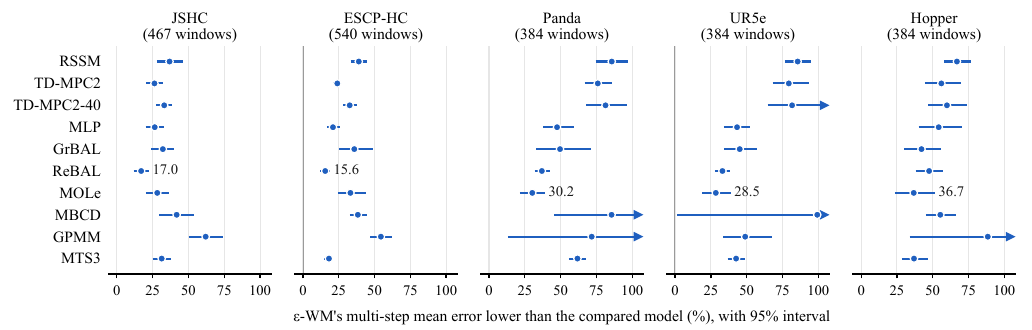}
\caption{A persistent change during the prediction. Each dot shows the reduction of \ewm{}'s multi-step mean error relative to one compared model, with its 95\% interval, in windows where a persistent change falls inside the 40 predicted steps: on any step on JSHC and ESCP-HC, and on step 25 or 31 on Panda, UR5e and Hopper. \ewm{}'s error is lower than that of all ten compared models on all five datasets, by at least 15.6\% (ESCP-HC, against ReBAL), and every interval lies above zero. The printed number marks the smallest reduction in each panel. Arrows mark intervals cut at 100\%, and the longest intervals again belong to MBCD and GPMM, as in \cref{fig:E-nochange}.}
\label{fig:E-during}
\end{figure*}

\paragraph{A change during the prediction.} A persistent change can also take effect after the context has ended, so the context does not show it at all. \Cref{fig:E-during} takes the windows in which such a change falls inside the 40 predicted steps. On JSHC and ESCP-HC, it can fall on any step. On Panda, UR5e and Hopper, whose changes follow a fixed schedule, it falls on step 25 or 31. \ewm{}'s error is lower than that of all ten compared models on all five datasets, by at least 15.6\% (ESCP-HC, against ReBAL), and every interval lies above zero. Thus, in simulation, \ewm{} keeps its lead even when the prediction runs across a change that the context could~not~show.

\begin{figure*}[t]
\centering
\includegraphics[width=\textwidth]{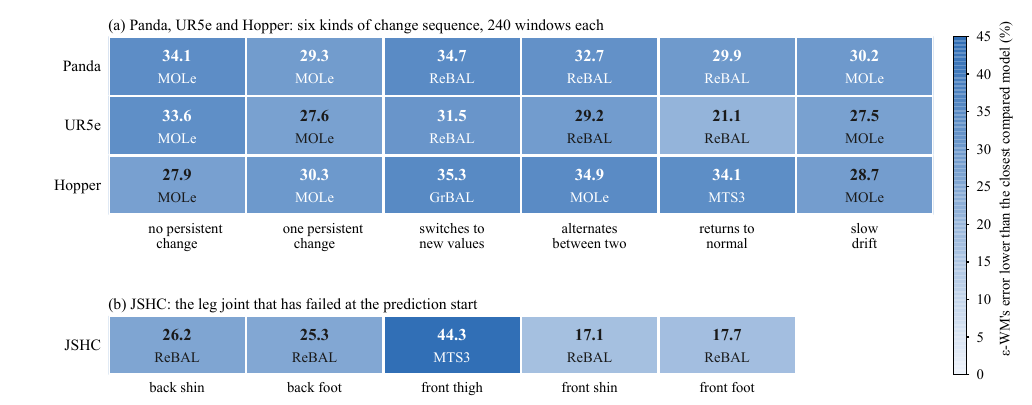}
\caption{\ewm{}'s lead for each kind of change. Each cell shows the reduction of \ewm{}'s multi-step mean error relative to the closest of the ten compared models, and names that model. (a)~Panda, UR5e and Hopper each contain six kinds of change sequence in equal numbers (240 windows each): no persistent change (only transient disturbances), one persistent change, switches to new values, alternation between two settings, changes that return to normal, and slow drift. (b)~JSHC windows grouped by the leg joint that has failed at the prediction start. \ewm{}'s error is lower in every cell, by 21.1\% to 35.3\% in (a) and 17.1\% to 44.3\% in (b). In all 230 comparisons behind the figure, \ewm{}'s error is lower in all 5 seeds, and the 95\% interval lies above zero.}
\label{fig:E-kinds}
\end{figure*}

\paragraph{Kinds of change.} The new datasets mix several kinds of change (Sec.~4.1 of the main paper and \cref{tab:datasets}), so here we compare the models within each kind. Each episode of Panda, UR5e and Hopper follows one of six kinds of change sequence, in equal numbers: no persistent change, one persistent change, switches to new values, alternation between two settings, changes that return to normal, and slow drift. On JSHC, we group the windows by the leg joint that has failed at the prediction start. In every cell of \cref{fig:E-kinds}, \ewm{}'s error is lower than that of the closest compared model, by \mbox{21.1\%--35.3\%} across the kinds of change and by 17.1\%--44.3\% across the failed joints. Thus, on these datasets, the lead does not rest on one kind of change.

\subsection{Errors by body part}
\label{sec:supp-E2}

The per-step errors so far are averages over all state dimensions. Here we split these errors by body part: a joint of an arm, or, for HalfCheetah (JSHC and ESCP-HC) and Hopper, the torso or one leg joint. The error of a part is the nMSE over that part's own state dimensions, that is, the angle and angular velocity of a joint, or the height, pitch and body velocities of the torso. Each state dimension belongs to one part. As in Tab.~1 of the main paper, all values are averaged over the 1,440 test windows of each dataset and 5 seeds.

\begin{figure*}[t]
\centering
\includegraphics[width=\textwidth]{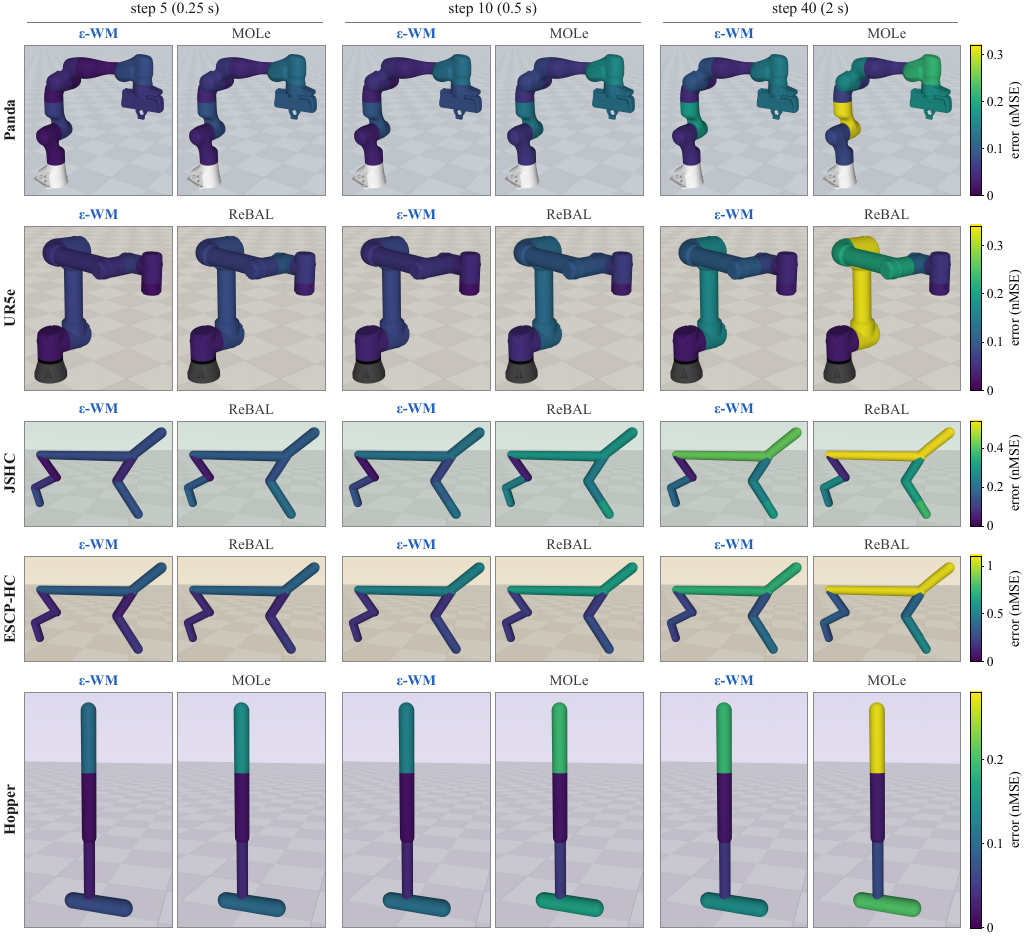}
\caption{Error of each body part on the five simulated datasets, painted on the robot in one fixed pose. Each row shows \ewm{} next to the strongest compared model of that dataset (MOLe on Panda and Hopper, ReBAL on UR5e, JSHC and ESCP-HC) at predicted steps 5, 10 and 40 (0.25, 0.5 and 2\,s ahead). A part's color is its error: the nMSE over the part's own state dimensions (a joint's angle and angular velocity, or the torso's height, pitch and body velocities), averaged over all 1,440 test windows and 5 seeds. Each row has one color scale, shared by both models and all three steps; a darker part has a lower error. The Panda hand has no joint of its own and takes the color of joint 7. On every part and at every step shown, \ewm{}'s error is lower than that of the strongest compared model. The margin is below 2.1\% only on the ESCP-HC front shin at step 5 and on the UR5e shoulder pan and wrist 3 at step 40.}
\label{fig:E-parts-render}
\end{figure*}

\paragraph{Errors on the robot.} To show where on the robot the per-step errors arise, \cref{fig:E-parts-render} colors each part by its error, for \ewm{} next to the strongest compared model of each dataset (Tab.~1 of the main paper), at predicted steps 5, 10 and 40. On all five datasets, \ewm{}'s error is lower on every part at all three steps. At step 40, the largest part errors of both models lie on the joint that lifts each arm (Panda joint 2 and the UR5e shoulder lift) and on the torso of HalfCheetah and Hopper. Thus, at these steps, \ewm{}'s lower error covers the whole robot, not just a few parts. \Cref{sec:supp-E5-body,sec:supp-E5-parts} show where on the robot \ewm{}'s corrections act and how much error each of its online memories removes~on~each~part.

\begin{figure*}[t]
\centering
\includegraphics[width=\textwidth]{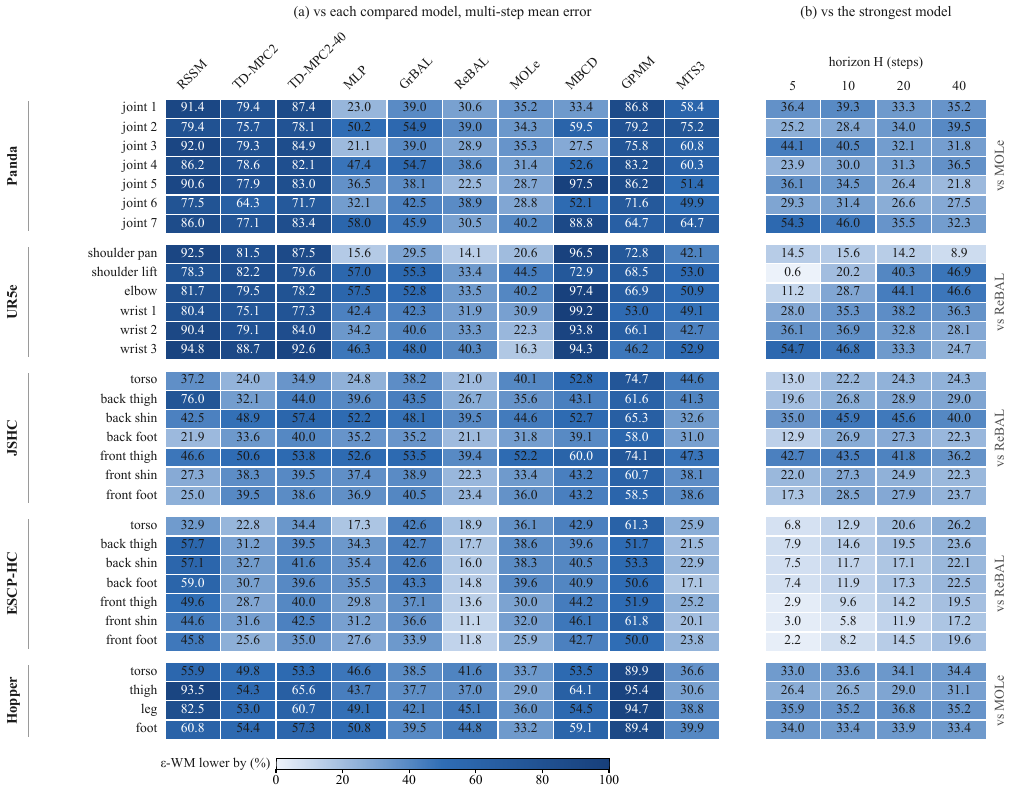}
\caption{Errors by body part against the compared models. Each row is one part: a joint of an arm, or for HalfCheetah (JSHC, ESCP-HC) and Hopper the torso or one leg joint. Each cell shows how much lower \ewm{}'s error is (\%), averaged over all 1,440 test windows and 5 seeds. (a)~Multi-step mean error against each of the ten compared models: \ewm{}'s error is lower in all 310 cells, by 11.1\% to 99.2\%. (b)~Mean error over steps 1 to $H$ against the strongest compared model of each dataset, with $H$ from 5, the earliest step drawn in \cref{fig:E-parts-render}: \ewm{}'s error is lower on every part at $H=5$, 10, 20 and 40, and also at every horizon in between.}
\label{fig:E-parts-heat}
\end{figure*}

\paragraph{Against every compared model.} A lower average error could still come from a few parts. \Cref{fig:E-parts-heat}a therefore compares \ewm{} with each of the ten compared models on every part, using the multi-step mean error. \ewm{}'s error is lower in all 310 cells, by 11.1\% to 99.2\%. \Cref{fig:E-parts-heat}b compares \ewm{} with the strongest compared model on every part, using the mean error over steps 1 to $H$ with $H=5$, 10, 20 and 40. \ewm{}'s error is lower on every part at these four horizons and at every horizon in between. Thus, the lead of Tab.~1 of the main paper holds part by part.

\begin{figure*}[t]
\centering
\includegraphics[width=\textwidth]{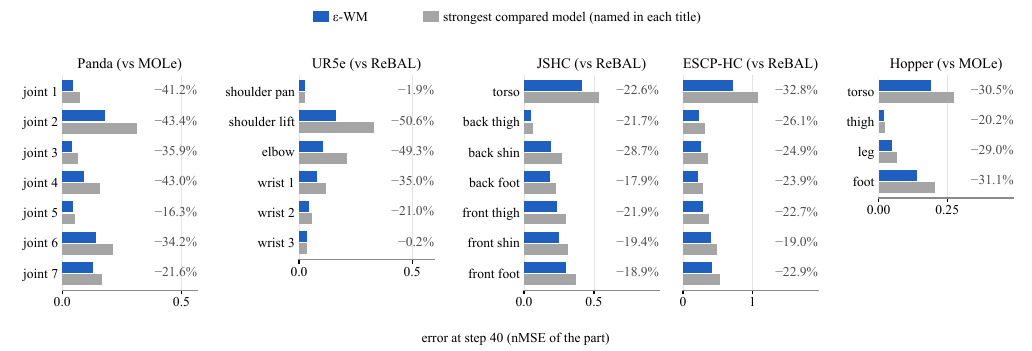}
\caption{Error of each body part at the last predicted step (step 40), for \ewm{} (blue) and the strongest compared model of each dataset (gray), averaged over all 1,440 test windows and 5 seeds; these are the step-40 values of \cref{fig:E-parts-render}. The numbers on the right give \ewm{}'s relative difference (negative: \ewm{} lower). \ewm{}'s error is lower on all 7 Panda joints, all 6 UR5e joints and every part of JSHC, ESCP-HC and Hopper, by 16.3\% to 50.6\% on 29 of the 31 parts, and by 1.9\% and 0.2\% on the UR5e shoulder pan and wrist 3.}
\label{fig:E-parts-step40}
\end{figure*}

\paragraph{At the last predicted step.} The last step is where persistence matters most. \Cref{fig:E-parts-step40} shows the step-40 errors of \cref{fig:E-parts-render} as bars, with \ewm{}'s relative difference on each part. \ewm{}'s error is lower on all 31 parts of the five datasets, by 16.3\%--50.6\% on 29 of these parts. On the UR5e shoulder pan and wrist 3, the two joints with the smallest step-40 errors of that arm (about 0.03), the two models are within 2\% of each other. On both arms, \ewm{}'s reduction is largest at the joint that lifts the arm: 43.4\% at Panda joint 2 and 50.6\% at the UR5e shoulder lift. On HalfCheetah and \mbox{Hopper}, \ewm{} lowers the torso error by 22.6\% (JSHC), 32.8\% (ESCP-HC) and 30.5\% (Hopper). Thus, even at the last predicted step, \ewm{}'s error is below that of the strongest compared model on every body part, and by at least 16\% on 29 of the 31 parts.

\subsection{The three designs}
\label{sec:supp-E3}

Section~4.4 of the main paper tests the three designs on the five simulated datasets, mostly with the multi-step mean error, and gives their key numbers. Here we give more results of these tests in the same order, multiple internal experts, fast--slow timescale adaptation and predictive correction, and also look at each horizon and each predicted step. In the predictive-correction tests and in the ablations of Sec.~4.6 of the main paper, we add ALFA's real flights. The interventions are those of \cref{sec:supp-mech}.

\paragraph{Multiple internal experts.} The routing test of Fig.~6a of the main paper stops the prediction errors from updating the routing memory at inference (\cref{sec:supp-mech}). This raises the multi-step mean error by 5.1\%--30.3\% on JSHC, Panda, UR5e and Hopper, with all 95\% intervals above zero, and by 0.9\% on ESCP-HC (95\% interval $-0.5\%$ to $2.1\%$), so reweighting the internal experts by their prediction errors improves the prediction accuracy. The results of the routing test around transient disturbances are in \cref{sec:supp-E5-transient}.

\begin{figure*}[t]
\centering
\includegraphics[width=0.686\textwidth]{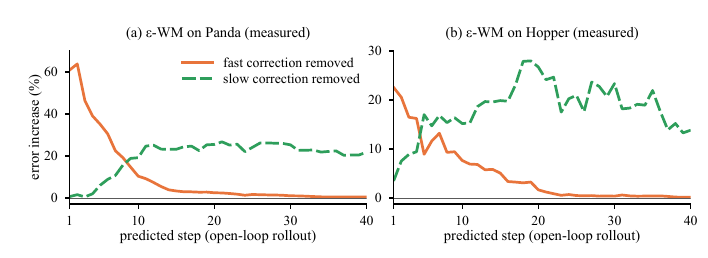}
\caption{Fast--slow timescale adaptation in \ewm{}, measured on (a)~Panda and (b)~Hopper: error increase at each predicted step when \ewm{}'s fast or slow correction is removed at inference (means over seeds). Removing the fast correction mainly raises the error at the early steps (step 1: +60.9\% on Panda and +22.6\% on Hopper), and removing the slow correction mainly raises it at the later ones (step 40: +21.9\% and +13.8\%). With no new prediction errors in the open-loop rollout, \ewm{}'s fast correction decays and its slow correction persists.}
\label{fig:E-dualrate}
\end{figure*}

\paragraph{Fast--slow timescale adaptation at steps 1 and 40.} If the fast correction dominates short-term prediction and the slow one long-term prediction, removing the fast one should mainly raise the step-1 error and removing the slow one mainly the step-40 error. As in Fig.~6b of the main paper, we remove one correction at inference. Removing the fast correction raises the step-1 error by 5.2\%--60.9\% but changes the step-40 error by at most 2.1\%. Removing the slow correction does the opposite: it changes the step-1 error by at most 4.9\% and raises the step-40 error by 9.9\%--37.6\%. The open-loop rollout of Eq.~(6) of the main paper does not force this division: the slow correction is added from step 1 on, yet removing it barely changes the step-1 error. Thus, on the five simulated datasets, the fast correction dominates short-term prediction and the slow one long-term prediction.

\paragraph{Early and late after a change.} If the fast correction acts early and the slow one persists, removing the fast correction should raise the error more right after a change, and removing the slow one more when the latest change is old or absent. On the two robot arms, we therefore compare these removals at different times since the latest change (\cref{sec:supp-mech}). Removing the fast correction raises the multi-step mean error more at 1--4 steps after a change (Panda 25.0\%, UR5e 24.9\%) than when the latest change is at least 60 steps old or absent (19.2\% and 15.0\%). Removing the slow one does the opposite: at 1--4 steps after a change, the increase is at most 6.7\%, less than for old or absent changes (17.2\% and 14.3\%). Thus, on both arms, the fast correction acts early and the slow one persists.

\paragraph{The division of labor after retraining.} Removing a correction at inference leaves the rest of \ewm{} as it was trained with both corrections. To check that the division of labor does not depend on this, we compare the models retrained with only the slow or only the fast correction (Fig.~6c of the main paper), which are the corresponding ablations of \cref{fig:supp-ablsens}a. At step 1, the slow-only model is worse than the fast-only one by 3.8\%--47.3\% of \ewm{}'s error. However, the two per-step error curves cross, and at step 40 it is better by 3.5\%--10.1\% on four datasets and about equal on UR5e (1.2\%). The full \ewm{} combines both strengths. At step 1, it is within 2.4\% of the fast-only model. At step 40, the fast-only model is 6.3\%--22.5\% worse than \ewm{} and the slow-only model 1.8\%--21.3\% worse. Thus, the retrained models show the same division of labor, and \ewm{}, which combines both corrections, is close to the fast-only model at step 1 and, at step 40, better than the fast-only model and as good as or better than the slow-only one.

\paragraph{Fast--slow timescale adaptation, step by step.} Figure~6b of the main paper checks the division of labor between the two corrections only at steps 1 and 40. \Cref{fig:E-dualrate} follows it over all 40 steps on one arm (Panda) and one legged robot (Hopper). Removing the fast correction at inference raises the step-1 error by 60.9\% on Panda and 22.6\% on Hopper, and its effect decays over the horizon, to under 1\% at step 40. On these two datasets, removing the slow correction changes the step-1 error by at most 3.5\% but raises the error at the later steps, by 21.9\% and 13.8\% at step 40. From step 9 on Panda and from step 5 on Hopper, removing the slow correction raises the error more than removing the fast one. Thus, once no new prediction errors arrive, the fast correction decays and the slow one persists.

\begin{figure*}[t]
\centering
\begin{minipage}[c]{0.47\textwidth}\centering
\includegraphics[width=\linewidth]{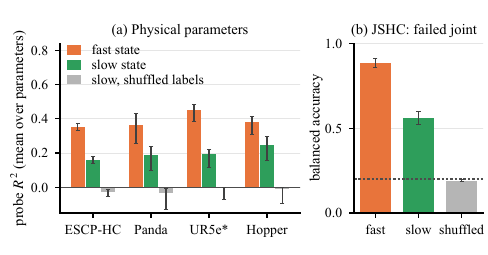}
\end{minipage}\hfill
\begin{minipage}[c]{0.49\textwidth}
\caption{Linear probes partly recover the hidden physical parameters from \ewm{}'s fast and slow states, although \ewm{} never receives these parameters as input. (a)~Mean $R^2$ over the changing parameters (ESCP-HC: gravity, joint damping and torso mass factors; Panda and UR5e: payload, joint friction and motor strength; Hopper: torso mass, ground friction and motor strength): 0.35--0.45 for the fast state and 0.16--0.25 for the slow state, while probes trained on shuffled labels score below zero. (b)~JSHC: balanced accuracy for the failed joint, 0.89 (fast) and 0.56 (slow), against 0.19 with shuffled labels (dashed line: chance, 0.2). Bars: scores on the test windows, with 95\% intervals. UR5e (*): 4 seeds.}
\label{fig:E-probes}
\end{minipage}
\end{figure*}

\paragraph{What the states hold.} If the fast and slow states carry the change, a linear probe should partly recover the hidden physical parameters. We fit linear probes on the states after the context (\cref{sec:supp-mech}). On ESCP-HC, Panda, UR5e and Hopper, their mean $R^2$ is 0.35--0.45 for the fast state ($3\times32$ values) and 0.16--0.25 for the slow state ($3\times3$ values), while probes trained on shuffled labels score below zero (\cref{fig:E-probes}a). On JSHC, the probe on the fast state identifies the failed joint with balanced accuracy 0.89 and that on the slow state with 0.56, against 0.19 with shuffled labels and a chance level of 0.2 (\cref{fig:E-probes}b). Thus, both states hold information about the change, although \ewm{} never receives these parameters as input. The fast state, with about ten times as many values, holds more of it but only decays in the open-loop rollout (Sec.~3.5 of the main paper), and \cref{sec:supp-E5-size} shows how the slow states take on a new value after a change.

\paragraph{Predictive correction on the multi-step mean error.} To compare the four ways of changing the role of \ewm{}'s prediction errors at inference (\cref{sec:supp-mech}), we measure the increase of the multi-step mean error on the five simulated datasets (Fig.~6d of the main paper). Every alteration raises it on every dataset. Removing both corrections raises it by 11.7\%--45.0\%, and setting the prediction errors to zero by 7.0\%--69.3\%. Zeroing differs from removing both corrections: with zero prediction errors, the routing memory follows the expert weights ($\vq_{j+1}=\vp_j$ in Eq.~(5) of the main paper), the fast states stay zero and the slow states move only with the learned bias $\vb_k$ of Eq.~(3), whereas removing both corrections keeps the routing memory obtained from the prediction errors. This fits the results: zeroing raises the error more than removing both corrections on Panda, UR5e and Hopper, where stopping the error-driven routing updates raises the error most (Fig.~6a of the main paper), about as much on JSHC, and less on ESCP-HC, where it raises the error least. Delaying the prediction errors by only 4 steps raises the multi-step mean error by 8.1\%--58.8\%, and using another sequence's prediction errors by 23.3\%--149.0\%, the most on every dataset. Thus, the content of the current sequence's prediction errors matters, not just their statistics.

\begin{figure*}[t]
\centering
\includegraphics[width=\textwidth]{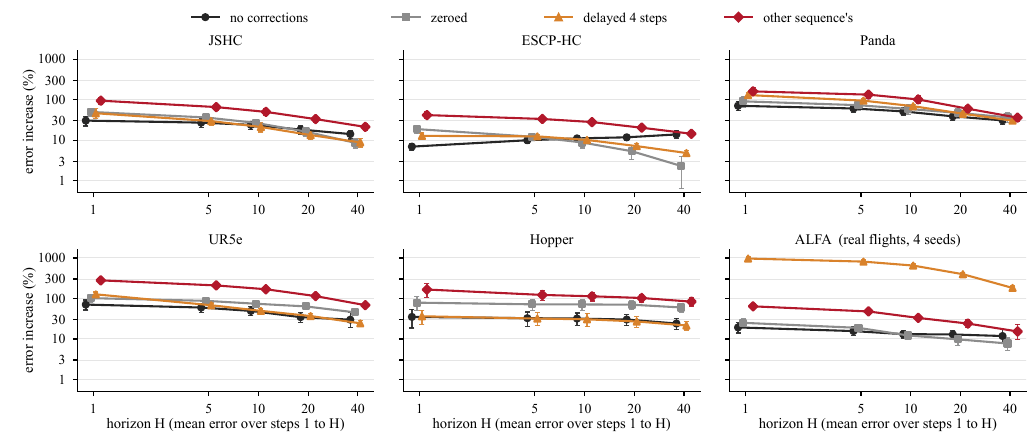}
\caption{Predictive correction on all six datasets. At inference, the role of \ewm{}'s prediction errors is changed in four ways (\cref{sec:supp-mech}): both corrections are removed (``no corrections''), or the prediction errors are set to zero (``zeroed''), delayed by 4 steps (``delayed 4 steps'') or taken from another sequence (``other sequence's''). Each point is the resulting increase of the mean error over steps 1 to $H$, for the horizons of the multi-step mean error, with 95\% intervals. Both axes are logarithmic. All 120 points and their intervals lie above zero. On the five simulated datasets, another sequence's prediction errors raise the multi-step mean error most (23.3\% to 149.0\%). On ALFA's real flights (4 seeds), delayed prediction errors do (414.8\%).}
\label{fig:E-predcoding}
\end{figure*}

\paragraph{Predictive correction at every horizon.} If \ewm{} relies on the current sequence's own prediction errors, changing these prediction errors should raise the error at every horizon, not only on average. As in Fig.~6d of the main paper, we change the role of the prediction errors at inference in four ways and measure the increase of the mean error over steps 1 to $H$ (\cref{fig:E-predcoding}). All 120 points and their intervals lie above zero, on all six datasets and at all five horizons. On ALFA's real flights, prediction errors delayed by 4 steps raise the error most at every horizon, by 964\% at $H=1$ and by 181\% at $H=40$ (the mean over steps 1 to 40). Only the prediction errors are delayed, not the observed states, so this tests whether the prediction errors must be timely, not robustness to a delayed sensor. Thus, \ewm{} depends on the current sequence's own, timely prediction errors, on ALFA's real flights and in simulation.

\begin{figure*}[t]
\centering
\includegraphics[width=\textwidth]{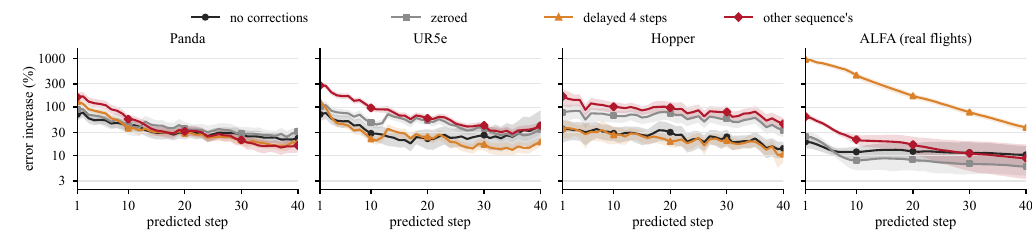}
\caption{Predictive correction at each predicted step on Panda, UR5e, Hopper and ALFA: the error increase when the role of \ewm{}'s prediction errors is changed as in \cref{fig:E-predcoding}, with 95\% intervals (log scale). All four alterations raise the error at every step, and every interval stays above zero up to step 40. On ALFA, delayed prediction errors raise the step-1 error by 964\%, and this increase shrinks at each later step.}
\label{fig:E-predcoding-step}
\end{figure*}

\paragraph{Predictive correction at every step.} Changing the role of the prediction errors might matter only for the early predicted steps, so here we also look at each step. \Cref{fig:E-predcoding-step} shows the same four alterations at each of the 40 predicted steps on the three new datasets (Panda, UR5e and Hopper) and on ALFA's real flights. All four raise the error at every step, and every interval stays above zero up to step 40. On ALFA, delayed prediction errors raise the step-1 error by 964\%, and the increase then shrinks at every step, to 38.3\% at step 40. Thus, on these datasets, the prediction errors of the context matter for every predicted step.

\begin{figure*}[t]
\centering
\includegraphics[width=\textwidth]{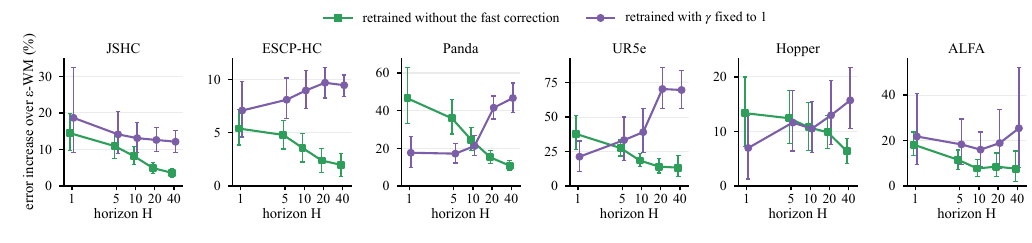}
\caption{Two retrained ablations across the prediction horizon, on all six datasets. Each point is the increase of the mean error over steps 1 to $H$ when \ewm{} is retrained without the fast correction (green) or with $\gamma$ fixed to 1 instead of its validated value (purple), with 95\% intervals. Both raise the error at every horizon on every dataset, including ALFA's real flights (4 seeds). Without the fast correction, the increase is largest at $H=1$ on every dataset.}
\label{fig:E-ablation}
\end{figure*}

\begin{figure*}[t]
\centering
\begin{minipage}[c]{0.47\textwidth}\centering
\includegraphics[width=\linewidth]{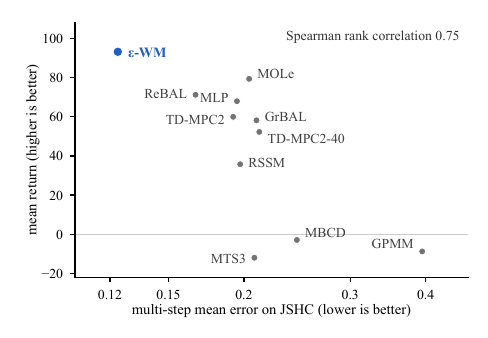}
\end{minipage}\hfill
\begin{minipage}[c]{0.49\textwidth}
\caption{Prediction accuracy and return in GrBAL's joint-failure control task on JSHC. All eleven models plan with the same cross-entropy-method planner (\cref{sec:supp-control}). Each point is one model, with its multi-step mean error on JSHC (Tab.~1 of the main paper; log scale) and its mean return over 5 seeds and 48 test episodes. \ewm{} has both the lowest multi-step mean error and the highest mean return (93.1, 13.8 above MOLe). More accurate models also tend to earn higher returns (Spearman rank correlation 0.75).}
\label{fig:E-control}
\end{minipage}
\end{figure*}

\paragraph{Ablations across the horizon.} \Cref{fig:supp-ablsens}a reports the retrained ablations on the multi-step mean error. \Cref{fig:E-ablation} follows two of these ablations over the horizon, \ewm{} retrained without the fast correction and \ewm{} retrained with the inverse routing temperature $\gamma$ of the multiple internal experts (Eq.~(1) of the main paper) fixed to 1, and adds ALFA (the third ablation, without the slow correction, is the fast-only model that Fig.~6c of the main paper already follows over the steps). Both raise the error at every horizon on all six datasets, and all 60 intervals lie above zero. Without the fast correction, the increase is largest at $H=1$ on every dataset (5.4\% to 46.5\%), as expected for a correction that decays over the horizon. With $\gamma$ fixed to 1, the error rises by 7.0\% to 70.4\% across datasets and horizons. On ALFA's real flights, the two models are 7.6\% to 25.4\% worse than \ewm{} at every horizon. Thus, the fast correction and the validated $\gamma$ also lower the error on real flights, and the fast correction lowers it most at short horizons.

\subsection{Control}
\label{sec:supp-E4}

\paragraph{Accuracy and return.} Section~4.5 of the main paper reports that \ewm{} has the highest mean return in GrBAL's joint-failure task on JSHC (protocol in \cref{sec:supp-control}). To show how the return relates to accuracy, \cref{fig:E-control} plots each model's mean return against its multi-step mean error on JSHC. All eleven models use the same planner. \ewm{} has both the lowest error and the highest mean return, 93.1, which is 13.8 above MOLe. Across the eleven models, more accurate models tend to earn higher returns (Spearman rank correlation 0.75). Thus, on this task, where all models plan in the same way, the most accurate model, \ewm{}, also attains the highest mean return.

\clearpage %
\makeatletter\global\@colroom\@colht\global\vsize\@colht\makeatother %

\subsection{Mechanism around a change}
\label{sec:supp-E5}

Section~4.3 and Fig.~5d--f of the main paper show, averaged over the five simulated datasets, how \ewm{}'s online memories keep it ahead of the strongest compared model after a persistent change. This subsection shows the same analysis on each dataset. It also adds the step-40 error and shows how the fast correction hands over to the slow one along the predicted steps (at each step since the change), what \ewm{} does with only one of its two corrections, and how the online memories turn around, \ie, change from fitting the old dynamics to fitting the new ones, in the context steps right after a change. \Cref{sec:supp-E5-transient,sec:supp-E5-body,sec:supp-E5-size} then compare persistent changes with transient disturbances, show where on the robot the corrections act, and relate \ewm{}'s response to the size of the change. \Cref{sec:supp-E5-kept} measures how long each model's correction works, \cref{sec:supp-E5-parts} shows what each online memory removes on each body part, and \cref{sec:supp-E5-held} tests whether the corrections of the routing memory and the slow state persist because these online memories are held fixed during the open-loop rollout.

\subsubsection{Changes and measures}
\label{sec:supp-E5-measures}

We use every persistent change in the test sets of the five simulated datasets with no other persistent change from 64 steps before to 40 steps after it and for which all windows whose prediction starts from 8 steps before to 48 steps after it exist. These are 329 changes on JSHC, 393 on ESCP-HC and 288 each on Panda, UR5e and Hopper, without the small drift steps (the small steps by which the parameters change during slow drift) on the last three (1,586 changes in all). Here a window is a 64-step context and the 40 steps that follow, taken at every step around a change, not only at the test queries. At $d$ steps since the change, the window starts its prediction $d$ steps after the change. For $d\geq 1$, the last $d$ transitions of its context follow the new dynamics. For $d\leq 0$, the context does not show the change yet, and the change reaches the prediction at step $1-d$. We evaluate the windows with $-5\leq d\leq 30$, where no window has another persistent change in its context or prediction; \cref{sec:supp-E5-turn,sec:supp-E5-body} also read windows up to $d=48$. The trained models of Tab.~1 of the main paper predict every window, without retraining. The same runs recompute the test queries of Tab.~1 and reproduce their stored predictions within $10^{-5}$, except \ewm{}'s on Hopper (up to $3.1\times10^{-5}$) and ReBAL's on UR5e (up to $2.1\times10^{-5}$), JSHC and ESCP-HC. On the last two, ReBAL's predictions of individual windows differ from the stored ones by more, although its multi-step mean error agrees to within 0.01\%; its errors for the individual windows in \cref{fig:E5b-ex1} are therefore taken from these runs. The compared model is the strongest one of Tab.~1 by the multi-step mean error: ReBAL on JSHC, ESCP-HC and UR5e, and MOLe on Panda and Hopper. At inference, we switch \ewm{}'s three online memories (the routing memory, the fast state and the slow state) on and off on the same contexts. The routing memory is either updated over the context or kept at its start value $\tfrac1K\mathbf 1$, and each correction is either kept or removed. With all three switched off, \emph{\ewm{} without memory updates} combines its base experts with expert weights from the router's scores of the current state and action alone, which is the same as keeping all online memories at their start values. From the eight combinations, exact Shapley values split the error that the three online memories remove together, \ie, the error of \ewm{} without memory updates minus that of \ewm{}. The value of an online memory is the error it removes when it is switched on, averaged over the six orders in which the three can be switched on, so the three values add up to the total. Because it also counts what an online memory removes while the others are switched off, a Shapley value can differ from the effect of removing one correction from the full \ewm{} (Fig.~6b of the main paper). Errors are averaged over the changes of each seed and then over the seeds (5 seeds; 4 on UR5e, as in \cref{sec:supp-mech}). Unless stated otherwise, bands and intervals in this subsection, including the numbers in square brackets, are 95\% bootstrap intervals from 2,000 replicates that resample the seeds and the test episodes, or the changes where stated. Differences between percentages are computed before rounding.

\subsubsection{Every dataset}
\label{sec:supp-E5-datasets}

\begin{figure*}[t]
\centering
\includegraphics[width=\textwidth]{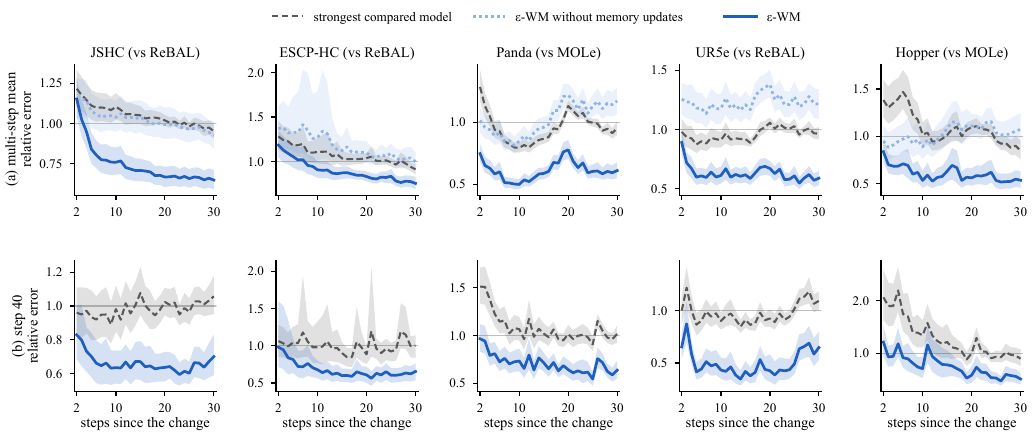}
\caption{Error after a persistent change on each simulated dataset, from 2 steps after the change (one step after it, the online memories are still turning around, \cref{sec:supp-E5-turn}), for \ewm{} (blue), the strongest compared model of Tab.~1 of the main paper (gray dashed) and \ewm{} without memory updates (light blue dotted; routing memory kept at its start value, no corrections). (a)~Multi-step mean error. (b)~Step-40 error of \ewm{} and the compared model. Each error is divided by the compared model's mean of the same error 17--30 steps after the change, so 1 marks that model's later level. \ewm{}'s error is lower than that of the compared model at every plotted step on all five datasets. Bands: 95\% intervals over test episodes and seeds.}
\label{fig:E5a-courses}
\end{figure*}

Figure~5d of the main paper averages the five datasets, which could hide differences between the datasets. \Cref{fig:E5a-courses} therefore shows each dataset from 2 steps after the change on, for the multi-step mean error and the step-40 error. On all five datasets, the multi-step mean error and the step-40 error of \ewm{} are below those of the compared model at every plotted step. Twenty to thirty steps after the change, \ewm{}'s multi-step mean error is 19.6\%--43.5\% below the compared model's and its step-40 error is 36.6\%--48.4\% below, and all ten 95\% intervals lie above zero. In the multi-step mean error, \ewm{} without memory updates is then less accurate than the compared model on Panda and UR5e, and on JSHC, ESCP-HC and Hopper the 95\% interval of their difference includes zero, so the base experts and the router alone do not account for \ewm{}'s lead. On each dataset, the error removed by the three online memories then equals 94\%--161\% of \ewm{}'s lead in the multi-step mean error (routing memory 42\%--72\%, fast correction 22\%--62\%, slow correction 27\%--40\%). Values above 100\% mean that \ewm{} without memory updates has a higher error than the compared model. Right after a change, by contrast, \ewm{} without memory updates is more accurate than MOLe on Panda and Hopper (\cref{fig:E5a-courses}a), as MOLe's correction arrives slowly (Sec.~4.3 of the main paper). Thus, on every simulated dataset, \mbox{20--30} steps after a persistent change, \ewm{}'s multi-step lead comes mostly from its online memories, not from its base experts and router alone.

\subsubsection{When and where each online memory acts}
\label{sec:supp-E5-memories}

\begin{figure*}[t]
\centering
\includegraphics[width=\textwidth]{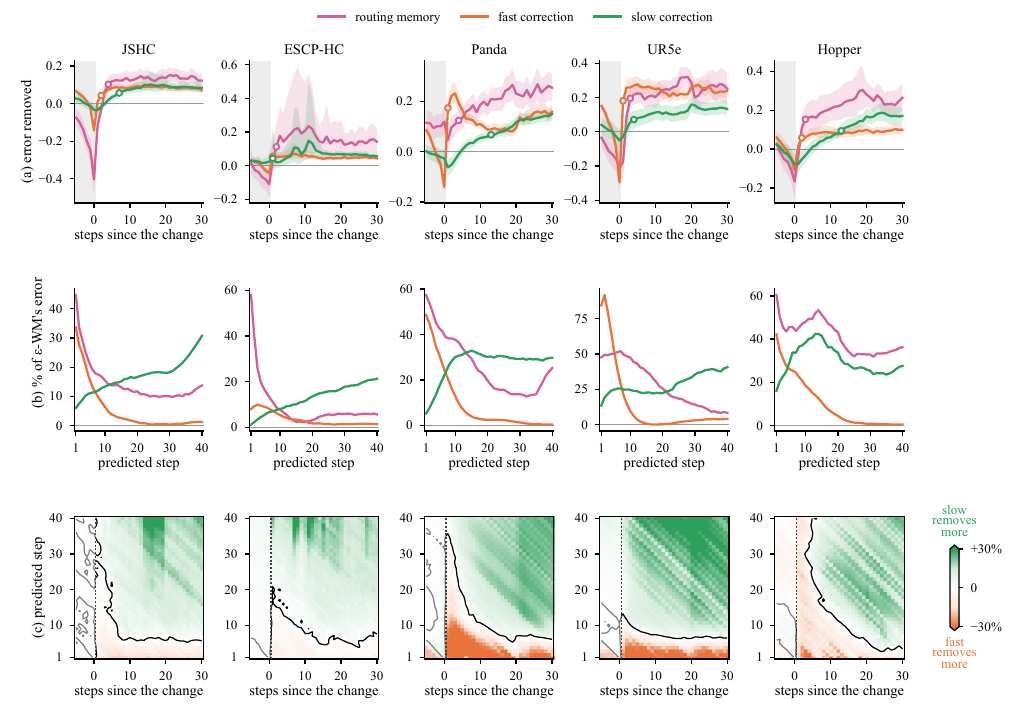}
\caption{What \ewm{}'s three online memories do on each simulated dataset. (a)~Error removed by the routing memory (pink), the fast correction (orange) and the slow correction (green), by steps since the change: Shapley values of the inference-time switches, in units of the compared model's multi-step mean error 17--30 steps after the change. Below 0, an online memory adds error. Gray: the change is not yet in the context. Rings: where each curve reaches half of its mean 17--30 steps after the change. Bands: 95\% intervals over test episodes and seeds. (b)~The same split at each predicted step, 20--30 steps after the change, in \% of \ewm{}'s error at that step. (c)~Which correction removes more error at each $d$ and predicted step: the error without the slow correction minus the error without the fast one, in \% of the error without both, with the routing memory as in \ewm{}. Black line (gray in the lighter area): equal. Lighter area left of the dotted line: the change is not yet in the context. Diagonal stripes join windows whose predicted steps reach the same future state.}
\label{fig:E5a-memories}
\end{figure*}

Figure~5e of the main paper shows on average when each online memory starts to remove error. \Cref{fig:E5a-memories}a shows the same Shapley values on each dataset, from 5 steps before the change. At $d=0$, the change has reached predicted step 1 but not yet the context, whose transitions all still follow the old dynamics, and the fast correction computed from these transitions still fits these dynamics (\cref{sec:supp-E5-turn}). From $d=1$ on, once the context shows the change, the fast correction removes error on every dataset. On average, the routing memory and the slow correction turn around later (\cref{sec:supp-E5-turn}). One step after the change, the routing memory still adds error on JSHC, UR5e and Hopper, and the slow correction on JSHC, Panda, UR5e and Hopper (on Panda and Hopper until 4 steps after the change). On JSHC, Panda, UR5e and Hopper, the fast correction reaches half of its later effect (its mean 17--30 steps after the change; counted at the first $d\geq 1$ at which its Shapley value is at least half of this mean, as for the rings of Fig.~5e of the main paper) 1--2 steps after the change, the routing memory 3--4 steps after it and the slow correction 4--13 steps after it. On ESCP-HC, all three reach it within 2 steps, the slow correction only through a brief rise one step after the change, after which it lies below this level again for two steps. Thus, on every dataset, the fast correction reaches half of its later effect no later than the routing memory and the slow correction, and on four datasets the slow correction reaches it last. Averaged over the five datasets (Fig.~5e of the main paper), the fast correction reaches it 1 step after the change, the routing memory 3 steps and the slow correction 7 steps after it.

During the open-loop rollout, the fast state only decays, while the slow state and the expert weights stay fixed (Sec.~3.5 of the main paper). The online memories should therefore serve different predicted steps. \Cref{fig:E5a-memories}b splits the error removed 20--30 steps after the change over the 40 predicted steps on each dataset. On every dataset, the fast correction removes more error than the slow one at step 1 and less at step 40. It removes error equal to 8.0\%--84.5\% of \ewm{}'s error at step 1, but at most 4.1\% at step 40, where the slow correction removes 21.3\%--40.8\%. The routing memory removes error at every predicted step. \Cref{fig:E5a-memories}c compares the two corrections at every $d$ and predicted step. One step after the change, the fast correction removes more error than the slow one at predicted steps 1--13 on every dataset, and at all 40 steps on Hopper. Averaged over 5--19 steps after the change, the slow correction removes more error from a predicted step between 7 and 12 onward (JSHC 7, ESCP-HC 9, Panda 10, UR5e 8, Hopper 12), and the fast correction removes more at every step before it. Thus, once the change is a few steps old, the fast correction serves the early predicted steps and hands over to the slow correction, which serves the later ones up to step 40, while the routing memory removes error at every predicted step.

\subsubsection{One correction alone}
\label{sec:supp-E5-single}

\begin{figure*}[t]
\centering
\includegraphics[width=\textwidth]{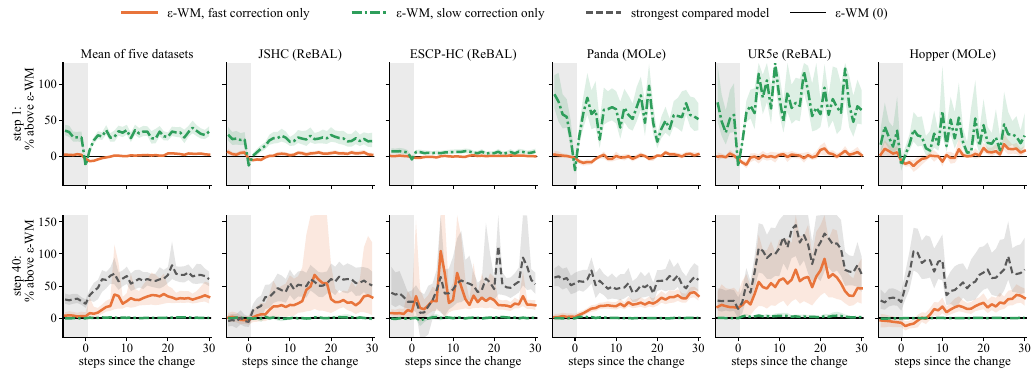}
\caption{One correction alone after a persistent change on each simulated dataset (the changes and seeds of \cref{sec:supp-E5-measures}). Error of \ewm{} with only the fast correction (orange, slow correction removed at inference) or only the slow correction (green dash-dot, fast correction removed), and of the strongest compared model of Tab.~1 of the main paper (gray dashed, bottom row), in \% above \ewm{}'s error at the same predicted step and the same number of steps since the change. Black line at 0: \ewm{}. Top: predicted step 1. Bottom: predicted step 40. Left: mean of the five datasets. Gray area: the change is not yet in the context. Bands: 95\% intervals over test episodes and seeds. Values outside the plotted range are cut off.}
\label{fig:E5a-single}
\end{figure*}

With only the fast correction, \ewm{}'s updates remove much less of its error at step 40, and with only the slow one at step 1 (Fig.~5f of the main paper). To check this on the per-step errors directly, we remove one of the two corrections at inference on the windows of \cref{sec:supp-E5-measures}, with the routing memory updated as in \ewm{}, and compare the step-1 and step-40 errors with \ewm{}'s at each step since the change (\cref{fig:E5a-single}). With only the fast correction, the step-1 error differs little from \ewm{}'s. One to four steps after the change, it is lower than \ewm{}'s on every dataset, by 5.1\% [4.1, 6.2] on average, as the slow state is then still turning around from the old dynamics (\cref{sec:supp-E5-turn}). Twenty to thirty steps after the change, it is 3.6\% [2.9, 4.3] above \ewm{}'s on average and at most 7.4\% above on each dataset. With only the fast correction, the step-40 error, however, rises above \ewm{}'s as more steps pass since the change. On average, it is more than 10\% above \ewm{}'s at every $d$ from 5 to 30, and \mbox{20--30} steps after the change it is 31.8\% [24.8, 44.0] above, and \mbox{22.4\%--53.1\%} above on the five datasets, with all intervals above 10\%. One to four steps after the change, it is lower than \ewm{}'s only on Hopper, by 9.0\% [2.9, 15.3], for the same reason as at step 1. With only the slow correction, the pattern is reversed. The step-40 error stays within 3\% of \ewm{}'s on every dataset, 1--4 and \mbox{20--30} steps after the change alike, but the step-1 error is higher on every dataset, by 5.6\%--52.1\% at 1--4 steps after the change (on average 19.9\% [17.0, 23.1]) and by \mbox{5.4\%--68.4\%} at \mbox{20--30} steps after it, with all intervals above zero. Thus, on every simulated dataset, \mbox{20--30} steps after a change, removing the slow correction raises the error far more at step 40 than at step 1, and removing the fast correction raises it more at step 1 than~at~step~40.

\begin{figure*}[t]
\centering
\includegraphics[width=\textwidth]{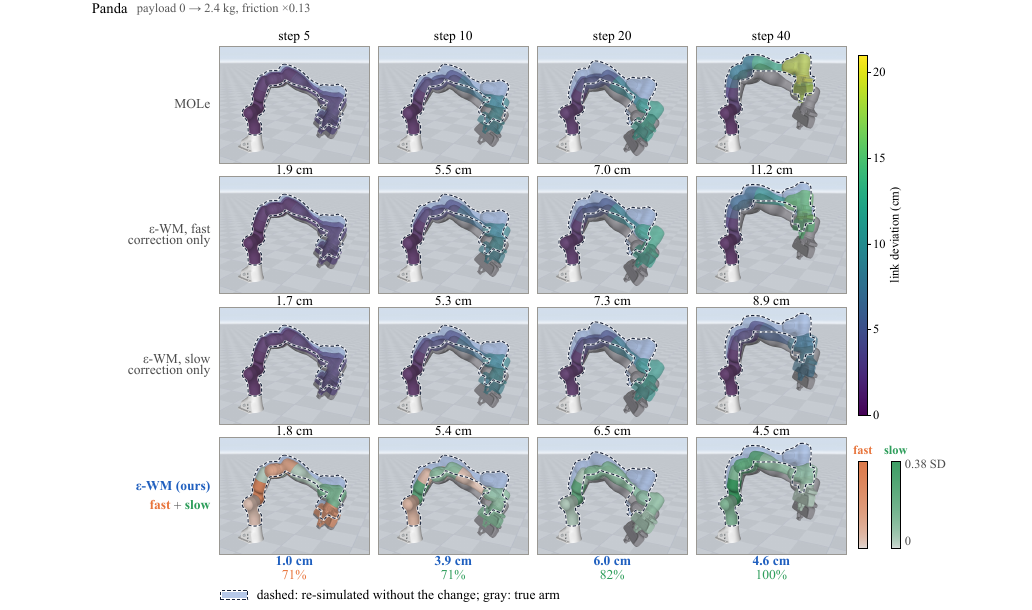}
\caption{The Panda example of Fig.~5 of the main paper with one of \ewm{}'s corrections removed at inference. Rows: MOLe, \ewm{} with only the fast correction and \ewm{} with only the slow correction (colored by link deviation, mean below), and \ewm{} (each link colored by the correction that moves its predicted state more, with color strength showing the size of that effect, 0 to 0.38 training standard deviations (SD); below: mean link deviation and the larger correction's share of the two corrections' squared effect on the whole arm). The colors show which correction moves the prediction more, not which removes more error, and averaged over all changes the two corrections act on largely the same parts (\cref{sec:supp-E5-body}). Dashed: arm re-simulated without the change. Gray: true arm.}
\label{fig:E5a-single-ex}
\end{figure*}

\Cref{fig:E5a-single-ex} illustrates the step-40 part of this on the Panda example of Fig.~5 of the main paper (selection in \cref{sec:supp-mech}). With only the fast correction, which decays during the open-loop rollout, \ewm{}'s predicted arm moves toward the arm re-simulated without the change, as MOLe's does. At step 40, it is 2.7\,cm from that arm, and its hand is 14.9\,cm above the true one (MOLe 18.5\,cm, \ewm{} 7.4\,cm). With only one correction, the arm is farther from the true arm than \ewm{}'s at steps 5 and 10 (fast only 1.7 and 5.3\,cm, slow only 1.8 and 5.4\,cm, \ewm{} 1.0 and 3.9\,cm), and with only the slow correction it ends as close to the true arm as \ewm{}'s (4.5 and 4.6\,cm~at~step~40).

On average, the strongest compared models fall short even of \ewm{} with only the fast correction. From $d=0$, when the change reaches predicted step 1, to $d=4$, their multi-step mean and step-40 errors fall by 49.2\% and 15.9\% on average over the five datasets, those of \ewm{} with only the fast correction by 57.5\% and 29.5\%, and \ewm{}'s by 59.2\% and 35.9\%. The compared models' drops are thus smaller than those of \ewm{} with only the fast correction, by 8.4 [6.3, 10.5] and 13.6 [2.2, 20.8] percentage points (computed before rounding). Twenty to thirty steps after the change, the compared models' step-40 error is 68.0\% [57.7, 78.5] above \ewm{}'s and that of \ewm{} with only the fast correction 31.8\% above (bottom row of \cref{fig:E5a-single}). The compared models' step-40 error is then above that of \ewm{} with only the fast correction on all five datasets, by \mbox{18.9\%--36.6\%}, while at step 1 the compared models' error is 29.3\% [24.1, 34.4] above \ewm{}'s on average (not drawn). Thus, after a change, the compared models fall farther behind \ewm{} at step 40 than at step 1 on average, and at step 40 \ewm{} relies on its slow correction and routing memory (\cref{fig:E5a-memories}b).

\subsubsection{How the online memories turn around}
\label{sec:supp-E5-turn}

\begin{figure*}[t]
\centering
\includegraphics[width=\textwidth]{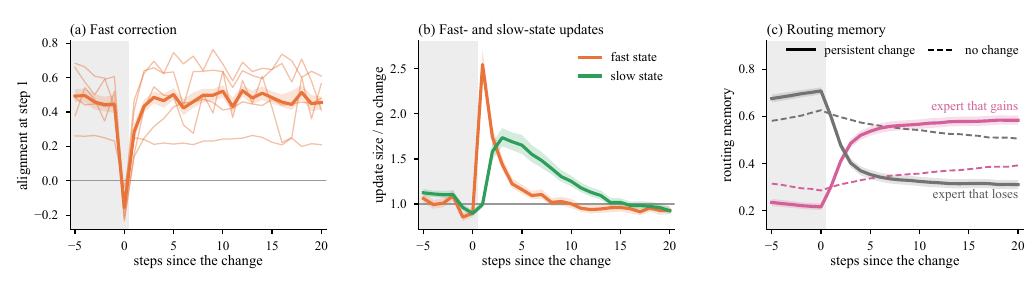}
\caption{How \ewm{}'s online memories turn around after a persistent change, as means over the five simulated datasets (thin lines in (a): the datasets). Gray: the change is not yet in the context. (a)~Alignment of the fast correction at predicted step 1: the cosine between the change it makes to the prediction and the error of the prediction without it, computed over all changes together (above 0: it moves the prediction toward the true state). (b,\,c)~Online memories in the context of the window 48 steps after the change, where $d$ counts the transitions after the change that the online memories have read, and the window whose context ends right before the change is the reference without a change. (b)~Size of the fast- and slow-state updates, averaged over the internal experts with their expert weights and divided by the size at the same context step in the reference. (c)~Routing memory of the internal expert that gains weight by the end of the context and of the internal expert that loses weight (solid), and of two internal experts chosen in the same way in the reference (dashed). Bands: 95\% intervals over the changes (a) or the test episodes (b, c), and the seeds.}
\label{fig:E5a-turn}
\end{figure*}

Before the context shows a change, \ewm{}'s online memories still fit the old dynamics, so after the change these online memories have to turn around. \Cref{fig:E5a-turn} follows the online memories through the context steps right after the change, on average over the five datasets. The online memories are read from the window that starts its prediction 48 steps after the change, whose context starts 16 steps before the change. The window whose context ends right before the change serves as a reference without a change. When the change has reached predicted step 1 but not yet the context ($d=0$), the fast correction still fits the old dynamics. Its alignment with the error it should remove (a cosine) falls on average from 0.45 at $d=-1$ to $-0.16$ at $d=0$ (\cref{fig:E5a-turn}a). In the update that reads the change ($d=1$), the fast states, averaged over the internal experts with their expert weights, move, depending on the dataset, 2.2--3.0 times as far as at the same step of the reference (\cref{fig:E5a-turn}b). Within 1--5 context steps, the fast correction regains 90\% of its later alignment. Before the change, the routing memory gives on average 0.71 to the internal expert that will lose weight and 0.22 to the internal expert that will gain weight. It completes half of its later movement within 2 steps on every dataset, and the two internal experts swap places after 2--5 steps (\cref{fig:E5a-turn}c). In the reference without a change, two internal experts chosen in the same way swap places only after 29--43 steps. The slow-state updates grow more slowly than the fast-state updates and are largest 3--5 steps after the change. Thus, right at the change, \ewm{}'s online memories still follow the old dynamics, and reading the new context turns these online memories around: the fast state within a few steps, the routing memory within 2--5 steps and the slow state over the following steps.

\subsubsection{Persistent changes and transient disturbances}
\label{sec:supp-E5-transient}

\begin{figure*}[t]
\centering
\includegraphics[width=\textwidth]{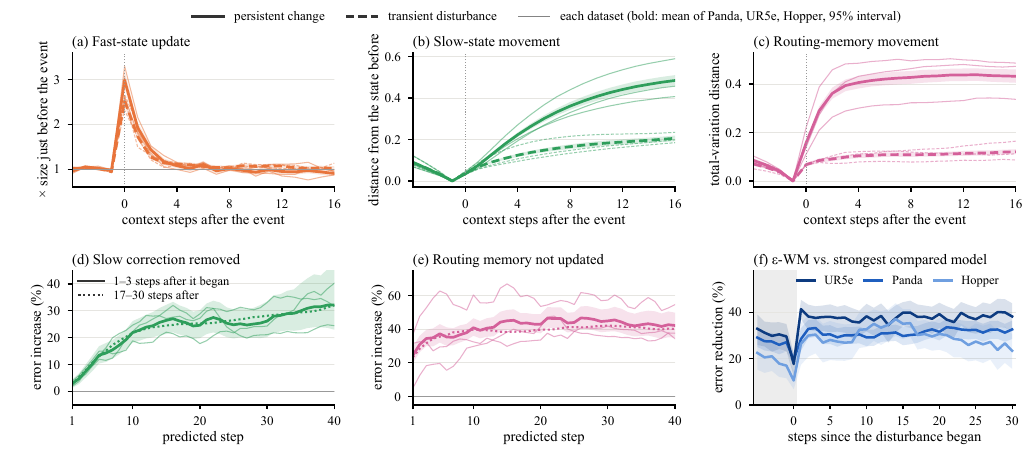}
\caption{Persistent changes and transient disturbances on Panda, UR5e and Hopper (per dataset 288 sudden persistent changes and 515--573 transient disturbances; 5 seeds, 4 on UR5e). (a--c)~\ewm{}'s online memories while it reads the context, aligned at the event (context step 0 is the update that reads the event, $d=1$ in \cref{fig:E5a-turn,fig:E5b-size}). (a)~Size of the fast-state update relative to its mean over the four updates before the event. (b)~Distance of the slow states from their values one step before the event, averaged over the internal experts with their expert weights at that step. (c)~Total-variation distance of the routing memory from its value one step before the event. Solid: persistent change. Dashed: transient disturbance. Bold: mean of the three datasets, with 95\% intervals. Thin: each dataset. (d,\,e)~Windows whose context ends 1--3 steps after a disturbance began (solid: mean of the three datasets with 95\% intervals; thin: each dataset) or 17--30 steps after it (dotted): error increase at each predicted step when the slow correction is removed at inference (d) or when the prediction errors no longer update the routing memory (e). (f)~Reduction of \ewm{}'s multi-step mean error over the strongest compared model (MOLe on Panda and Hopper, ReBAL on UR5e) in the windows of these disturbances, by the steps since the disturbance began, with 95\% intervals. Gray: the disturbance is not yet in the context. The 95\% intervals resample the events and the seeds.}
\label{fig:E5b-transient}
\end{figure*}

Right after a new change appears, it is hard to tell from a transient disturbance (Sec.~1 of the main paper). If \ewm{} tells the two apart, a disturbance should not overwrite what its slow state and routing memory have learned. On Panda, UR5e and Hopper, the three datasets with transient disturbances, we therefore align \ewm{}'s memory updates in the context at the event (a persistent change or a transient disturbance). Per dataset, we compare 288 sudden persistent changes (without the small drift steps) with 515--573 transient disturbances (impacts and pulses of friction or motor strength). None of the windows used for these disturbances has a persistent change in its context or prediction (\cref{fig:E5b-transient}).

In the update that reads the event, the fast-state update rises to 2.6--3.3 times its mean size over the four updates before the event for a persistent change and to 2.3--2.7 times for a disturbance, and four steps later it is back at \mbox{0.99--1.15} times that size (\cref{fig:E5b-transient}a). Only a persistent change makes the slow state and the routing memory keep moving clearly faster than before over the next few steps. Over the four updates after a persistent change, the slow state moves 1.4--1.9 times as far as over the four updates before it, but only \mbox{0.9--1.0} times as far after a disturbance (\cref{fig:E5b-transient}b). Four steps after the event, the routing memory is at a total-variation distance of 0.30--0.49 from its value before a persistent change, but of only 0.08--0.12 after a disturbance, little more than the 0.06--0.10 it moves over the same number of updates before the disturbance (\cref{fig:E5b-transient}c). Impacts show that this difference does not simply follow the size of the error. An impact raises \ewm{}'s one-step prediction error 1.6--9.7 times as much as a persistent change, yet it moves the routing memory less, 0.37--0.73 times as far at the event step and 0.17--0.32 times as far by the fourth step. Individual examples can mix the two kinds of event, as in the UR5e example of Fig.~5a of the main paper, where a motor-strength pulse spans the change and an impact follows 2 steps later. The averages above are aligned at one kind of~event~at~a~time.

What the slow state and the routing memory learned before a disturbance should therefore remain useful after it. In windows whose context ends 1--3 steps after a disturbance began, removing the slow correction at inference raises the multi-step mean error by 14.3\%--18.6\%, and averaged over the three datasets it raises the error at every one of the 40 predicted steps. When the prediction errors no longer update the routing memory at inference, as in the routing test of Fig.~6a of the main paper, the error rises by 26.0\%--52.6\%. Both increases are about as large as 17--30 steps after the disturbance began, 15.4\%--19.8\% and 25.3\%--49.4\% (\cref{fig:E5b-transient}d,\,e). For every prediction start from 5 steps before to 30 steps after the disturbance began, \ewm{}'s multi-step mean error is below that of the strongest compared model (MOLe on Panda and Hopper, ReBAL on UR5e; \cref{fig:E5b-transient}f). One to four steps after a disturbance began, it is 28\%--39\% lower, about the same as 17--30 steps after the disturbance began (28\%--38\%). Thus, on Panda, UR5e and Hopper, \ewm{}'s slow state and routing memory keep moving clearly faster only when a change persists. A transient disturbance is taken up by the fast state, whose updates are back close to their earlier size within four steps, and it does not overwrite what the slow state and the routing memory have learned.

\FloatBarrier
\subsubsection{Where on the robot the corrections act}
\label{sec:supp-E5-body}

To see where on the robot the fast and slow corrections act, we remove one of the two corrections at inference and measure how much the prediction of each state dimension changes. We call this change the effect of that correction. As in \cref{sec:supp-E2}, every state dimension belongs to one body part. The share of a part is the sum of the squared effects on its state dimensions divided by the same sum over all state dimensions, with both sums taken over all changes; the share is then averaged over the seeds.

\begin{figure*}[t]
\centering
\includegraphics[width=\textwidth]{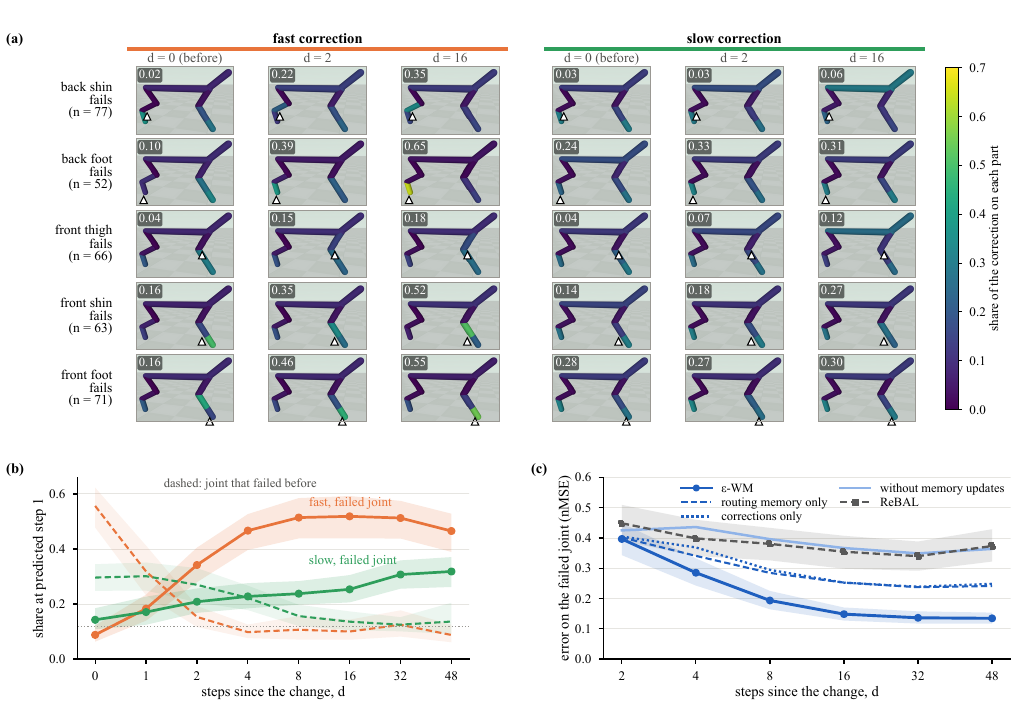}
\caption{Where \ewm{}'s corrections act after a leg joint fails (JSHC, all 329 persistent changes, 5 seeds). (a)~Share of the fast and of the slow correction on each body part at predicted step 1, summed over all changes in which the marked joint (triangle) fails; the number is this joint's share. Before the change is in the context ($d=0$), and 2 and 16 steps after it. (b)~Share on the failed joint (solid) and on the joint that failed before (dashed), with 95\% intervals. Dotted: even spread over the 17 state dimensions (0.12). (c)~Multi-step mean error per state dimension on the failed joint from 2 steps after the change: \ewm{}, \ewm{} with only its routing memory or only its two corrections, \ewm{} without memory updates (routing memory kept at its start value, no corrections), and ReBAL. Bands: 95\% intervals for \ewm{} and ReBAL. In (b) and (c), the horizontal axis is not to scale. Intervals resample the changes and the seeds.}
\label{fig:E5b-jshc}
\end{figure*}

On JSHC, we know which joint has failed, so we start with it (\cref{fig:E5b-jshc}). When a leg joint fails, the fast correction soon moves to it. Summed over all 329 changes, the fast correction's share on the joint's two state dimensions at predicted step 1 rises from 9\% before the change is in the context to 34\% two steps and 51\% eight steps after it, while an even spread would give 12\%. Its share on the joint that failed before drops from 56\% to 15\% within two steps. The slow correction moves later. Eight steps after the change, the fast correction's share on the failed joint has already reached the level it has 32--48 steps after the change, while the slow correction's share has gone only about half of the way there from its value before the change (\cref{fig:E5b-jshc}a,\,b).

\ewm{}'s lead over ReBAL is also concentrated on the failed joint. Four or more steps after the change, 54\% of \ewm{}'s lead over ReBAL in the multi-step mean error lies on the 2 of 17 state dimensions of the failed joint. On this joint, \ewm{} without memory updates (routing memory kept at its start value, no corrections, as in Fig.~5d of the main paper) is about as accurate as ReBAL, with a multi-step mean error per state dimension of 0.38 against 0.37. The routing memory alone and the two corrections alone each remove about 0.10 of this error, and all three together bring it down to 0.18, about half of ReBAL's (\cref{fig:E5b-jshc}c). From four steps after the change on, ReBAL's error on this joint stays between 0.34 and 0.40, while \ewm{}'s falls to 0.15 after 16 steps and to 0.13 after 48 steps.

\begin{figure*}[t]
\centering
\includegraphics[width=\textwidth]{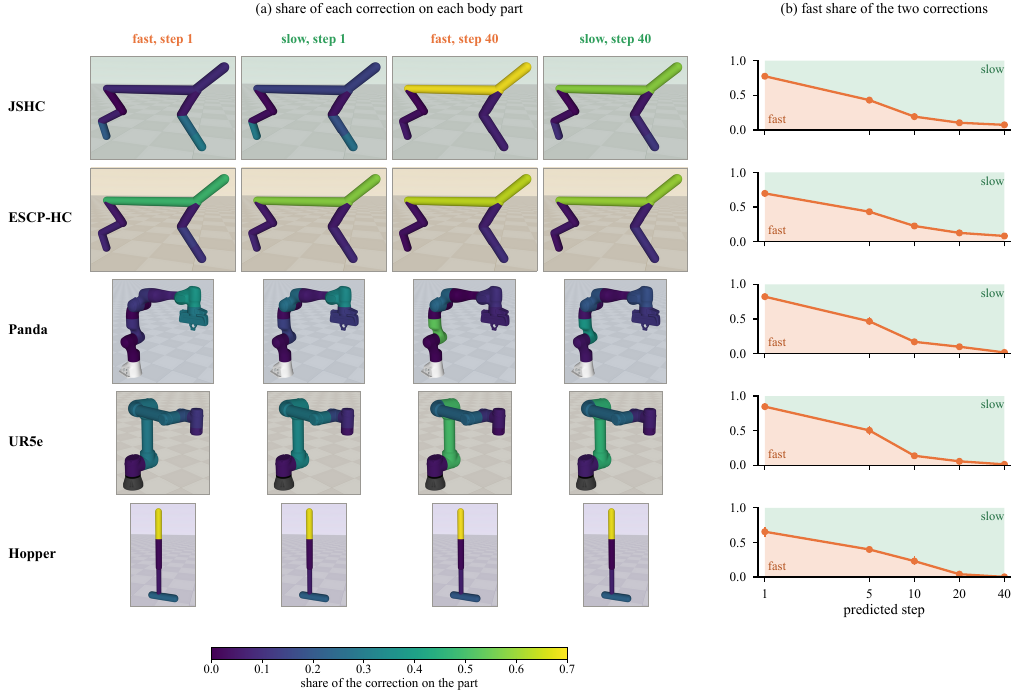}
\caption{Where the corrections act on the five simulated datasets (all persistent changes, Panda, UR5e and Hopper without the small drift steps; windows with the change in the context; 5 seeds, 4 on UR5e). (a)~Share of the fast or the slow correction on each body part at predicted steps 1 and 40. Each render shows where one correction acts, not how large it is, and the shares of one correction sum to 1 over the parts. (b)~Fast correction's part of the two corrections' combined effect (squared, summed over all state dimensions) at predicted steps 1, 5, 10, 20 and 40, with 95\% intervals that resample the changes and the seeds; the rest is the slow correction.}
\label{fig:E5b-robots}
\end{figure*}

On the other datasets, the corrections also concentrate on a few parts (\cref{fig:E5b-robots}a; windows with the change in the context; on Panda, UR5e and Hopper, without the small drift steps). On both arms, the two corrections together act mainly on the pitch joints that carry the arm's weight, 84\% on the shoulder lift, elbow and wrist 1 joints of the UR5e and 77\% on joints 2, 4 and 6 of the Panda. On ESCP-HC, the 5 of 17 state dimensions of the torso take 56\%, and on Hopper, the 5 of 11 state dimensions of the torso take 69\%. The fast and the slow correction act on largely the same parts. Weighted over the whole prediction, their shares differ by less than 0.06 on 26 of the 31 parts of the five datasets. What differs is the timing (\cref{fig:E5b-robots}b). At predicted step 1, the fast correction makes up 66\%--85\% of the two corrections' combined effect (squared), at step 10, 14\%--23\%, and at step 40, at most 8\%, on each of the five datasets.

\begin{figure*}[t]
\centering
\includegraphics[width=\textwidth]{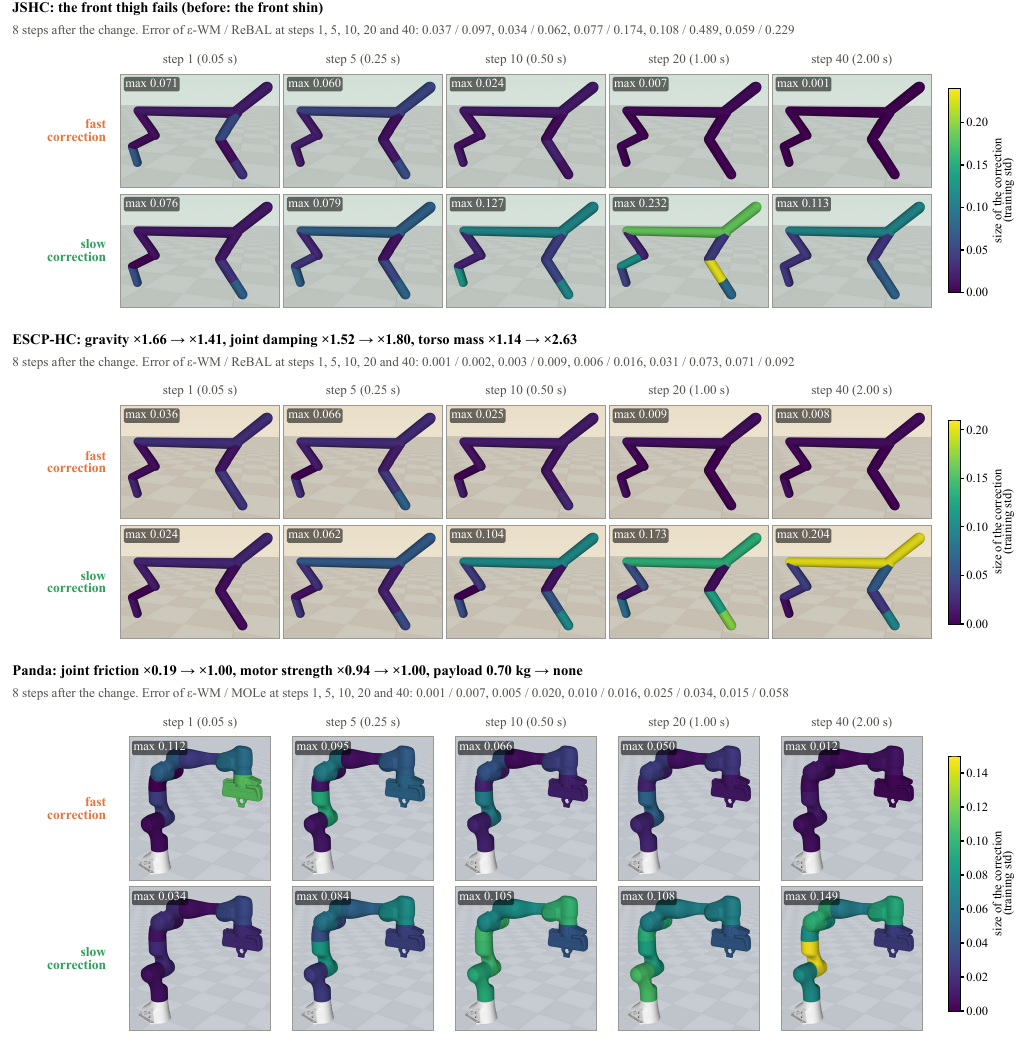}
\caption{Examples on JSHC, ESCP-HC and Panda. Each example is one test window whose context ends 8 steps after a persistent change. The robot is drawn in a fixed pose, and each part is colored by the size of the effect of the fast correction (top row) or the slow correction (bottom row) on its predicted state at predicted steps 1, 5, 10, 20 and 40 (root mean square over the part's state dimensions, in training standard deviations; one color scale per example; max: largest part value in the frame). The title gives the change, and the line below it the per-step errors of \ewm{} and of the strongest compared model (both from the same training seed) at the five steps. Each example is chosen among the windows in which \ewm{}'s error is the lower one at all five steps (\cref{sec:supp-E5-body}).}
\label{fig:E5b-ex1}
\end{figure*}

\begin{figure*}[t]
\centering
\includegraphics[width=\textwidth]{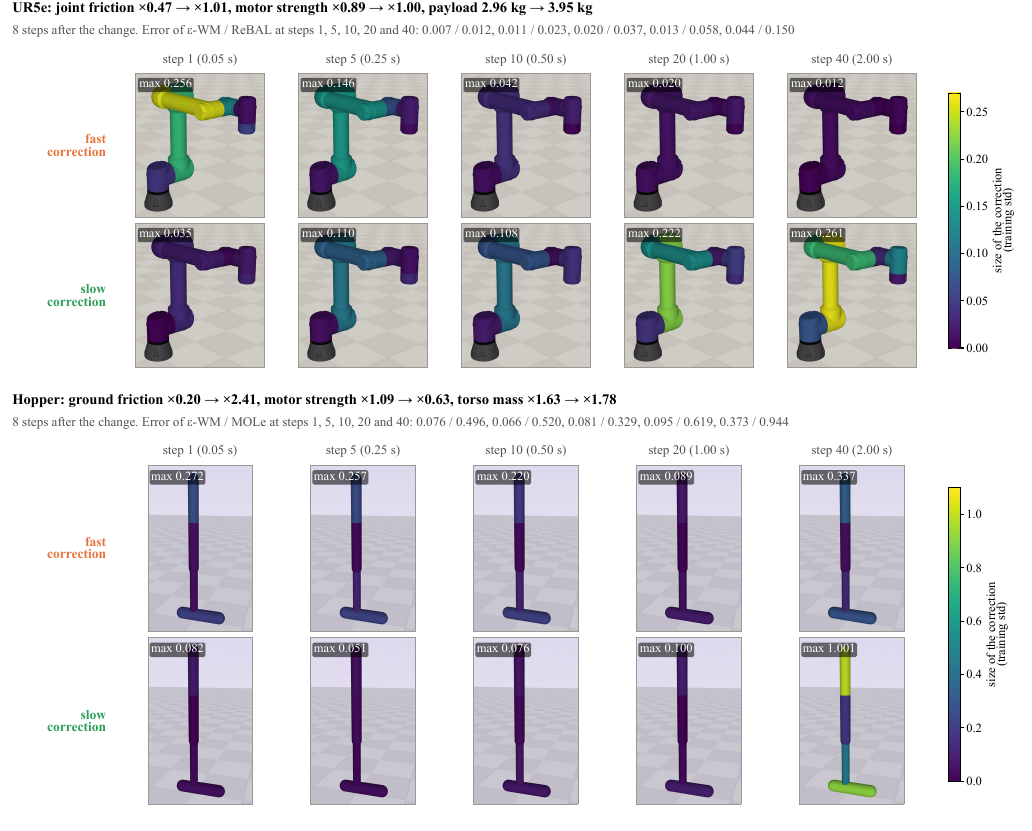}
\caption{Examples on UR5e and Hopper, drawn as in \cref{fig:E5b-ex1}.}
\label{fig:E5b-ex2}
\end{figure*}

\Cref{fig:E5b-ex1,fig:E5b-ex2} show one example per simulated dataset, each a test window whose context ends 8 steps after a persistent change. Each is an illustration chosen by a fixed rule from one training seed. Of the changes of \cref{sec:supp-E5-measures}, we keep those whose window has a lower error for \ewm{} than for the strongest compared model at all five shown steps (1, 5, 10, 20 and 40), and take the one at the middle rank of \ewm{}'s relative lead in the multi-step mean error. Each part is colored by the size of the effect of the fast correction (top row) or the slow correction (bottom row) on its predicted state. As the prediction is rolled out, this effect at a step also contains the differences that the correction made at earlier predicted steps, which can spread to other parts and grow. On Hopper, for example, the fast correction's effect is larger at step 40 than at step 20, although the fast correction has decayed by then. In all five examples, the slow correction has the larger effect at steps 20 and 40. In the JSHC example, the failed front thigh is not the part with the largest effect of either correction at any shown step. Of the five joints that can fail, the front thigh is the one on which the fast correction concentrates least after a change (\cref{fig:E5b-jshc}a).

Thus, in simulation, the fast and the slow correction act on largely the same body parts, and what differs is when each correction acts. The fast correction mainly serves the early predicted steps, and the slow correction persists to the end of the prediction. On JSHC, summed over all changes, the slow correction moves to the failed joint later than the fast one, and four or more steps after the change, \ewm{}'s lead over ReBAL lies mainly on this joint and comes from its routing memory and its two corrections.

\FloatBarrier
\subsubsection{Size of the change}
\label{sec:supp-E5-size}

\ewm{} updates its fast and slow states from the prediction errors, and the prediction errors also adjust its routing memory. A larger change, which should cause larger prediction errors, should therefore also cause larger updates. To test this, we use all 1,545 persistent changes of ESCP-HC, Panda, UR5e and Hopper (including the small drift steps of Panda, UR5e and Hopper). The size of a change combines the changes of all physical parameters. Factors are taken as logarithms and the payload in kilograms, each parameter's change is divided by that parameter's standard deviation over the values before and after these changes, and the size is the square root of the sum of their squares. A JSHC change switches the failed joint and has no size, so JSHC is left out of this size analysis.

\begin{figure*}[t]
\centering
\includegraphics[width=\textwidth]{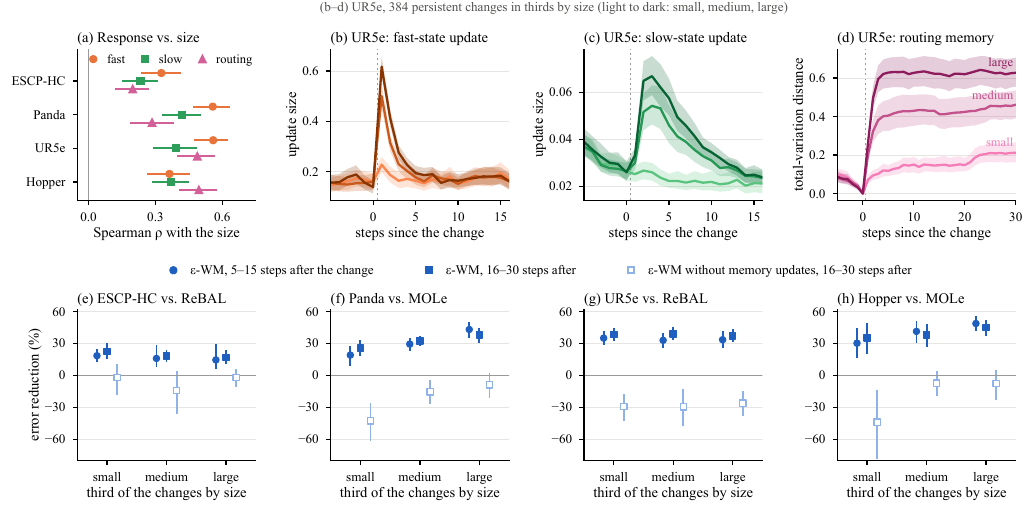}
\caption{\ewm{}'s response grows with the size of the change. (a)~Spearman rank correlation between the size of a persistent change and the fast-state update in the 4 context steps after it, the slow-state updates in the 16 steps after it and the movement of the routing memory 16--30 steps after it, with 95\% intervals (393 changes on ESCP-HC, 384 on each of the other three datasets). (b--d)~UR5e, the 384 persistent changes in thirds by size (light to dark: small, medium, large): size of the fast- and slow-state updates, and total-variation distance of the routing memory from its value before the change, by the context steps since the change. Mean over the 4 UR5e seeds with 95\% intervals; dashed vertical line: the change enters the context. (e--h)~Reduction of the multi-step mean error over the strongest compared model in each third of the changes by size, for \ewm{} 5--15 and 16--30 steps after the change and for \ewm{} without memory updates (routing memory kept at its start value, no corrections) 16--30 steps after it, with 95\% intervals.}
\label{fig:E5b-size}
\end{figure*}

The size of a change is positively correlated with the fast-state update in the 4 context steps after it (Spearman rank correlation 0.33--0.56), with the slow-state updates in the 16 steps after it (0.23--0.42) and with how far the routing memory has moved 16--30 steps after it (0.20--0.50). All 95\% intervals lie above zero (\cref{fig:E5b-size}a). On Panda, UR5e and Hopper without the small drift steps, the correlations are 0.21--0.37, again with all intervals above zero. \Cref{fig:E5b-size}b--d show this on UR5e, whose 384 changes are split into thirds by size. After a large change, the fast-state update that reads the change is about three times its size at the same context step without a change (0.62 against 0.21). The slow-state updates peak two steps later at about twice their size without a change (0.067 against 0.032). These references without a change are not drawn in \cref{fig:E5b-size}. The routing memory has moved by a total-variation distance of 0.59 three steps after the change and by 0.63 thirty steps after it, against 0.10 and 0.21 after a small change.

\begin{figure*}[t]
\centering
\includegraphics[width=\textwidth]{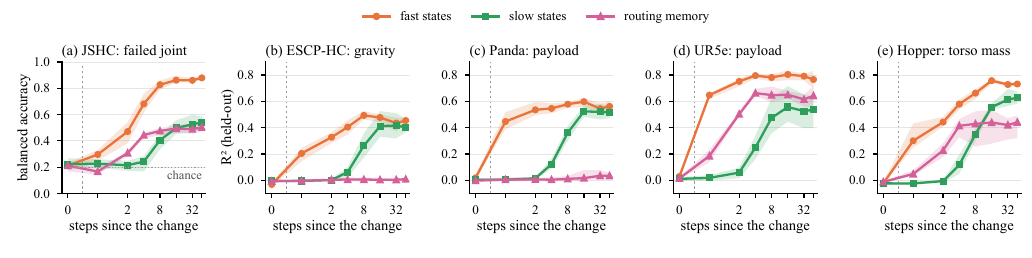}
\caption{What \ewm{}'s online memories hold after a persistent change: linear read-out of the parameter's value after the change from the fast states (3$\times$32 numbers), the slow states (3$\times$3) or the routing memory (3) at the prediction start, by the steps since the change. Ridge regression with 5-fold cross-validation over test episodes, scored on held-out episodes; $R^2$, or balanced accuracy for JSHC's failed joint (dotted: chance, 0.2). Mean over the seeds (5, and 4 on UR5e); bands: range over seeds. Dashed vertical line: the change enters the context; the horizontal axis is not to scale (linear up to 2 steps, logarithmic beyond). All persistent changes of each dataset.}
\label{fig:E5b-readout}
\end{figure*}

The online memories also come to hold the new parameter values (\cref{fig:E5b-readout}). On the test windows, a linear ridge read-out maps the fast states, the slow states or the routing memory at the prediction start to the value of the parameter after the change. It is fit with 5-fold cross-validation over episodes and scored only on held-out episodes. On JSHC, it reads out the failed joint and is scored by balanced accuracy (chance 0.2). Unlike the probes of \cref{sec:supp-E3}, which are fit on training and validation windows, this read-out is fit on the test windows, separately for each number of steps since the change. Two context steps after a change, the new value can already be read from the fast states, with $R^2$ of 0.33--0.75 for gravity on ESCP-HC, the payload on Panda and UR5e and the torso mass on Hopper, and a balanced accuracy of 0.47 on JSHC. The slow states hardly hold it yet, with $R^2$ of at most 0.06 and a balanced accuracy of 0.21 on JSHC, about chance. The slow states take it up over the next steps, with $R^2$ of 0.06--0.25 four steps after the change, 0.26--0.47 after eight steps and 0.41--0.56 after sixteen (JSHC 0.40 and 0.50 after eight and sixteen steps). Even the routing memory, only 3 numbers, takes up the new value of the UR5e payload, the Hopper torso mass and the JSHC failed joint within 4 steps. For the UR5e payload, its $R^2$ rises from 0.01 before the change is in the context to 0.66 four steps after it, for the Hopper torso mass from $-0.01$ to 0.41, and its balanced accuracy for the JSHC failed joint from 0.21 to 0.44. The fast states take up the new value earlier and the slow states follow, which matches fast--slow timescale adaptation.

Five to thirty steps after a change, \ewm{} stays ahead after small and large changes alike (\cref{fig:E5b-size}e--h): \ewm{}'s multi-step mean error is 14.5\%--48.7\% below that of the strongest compared model in every third of the changes by size, and its step-40 error (not drawn) is 18.6\%--54.3\% below it, with all 95\% intervals above zero. Sixteen to thirty steps after a change, the lead in the multi-step mean error relies on the online memories. In this error, \ewm{} without memory updates is then not more accurate than that model in any third, while \ewm{}'s error is 16.9\%--44.9\% below that model's in every third. Thus, on these four datasets, larger changes tend to bring larger memory updates, the new parameter values enter the fast states before the slow states, and five to thirty steps after a change, \ewm{} is more accurate than the strongest compared model whether the change is~small~or~large.

\subsubsection{How long each model's correction works}
\label{sec:supp-E5-kept}

A persistent change acts at every predicted step while no new observations arrive, so a model's adaptation must persist to the last step (persistence, Sec.~4.3 of the main paper). We measure this by comparing each model with the same model with its inference-time updates switched off, on the windows of \cref{sec:supp-E5-measures}. For \ewm{}, this is \ewm{} without memory updates (all online memories at their start values). For MOLe, the parameters learned in training are used for every window, without its online updates of the parameters and of its mixture over tasks. For ReBAL, the recurrent state starts from zero at the prediction start, the state with which every training window starts, instead of being computed from the context. With the updates on, the same code reproduces the window predictions of the runs of \cref{sec:supp-E5-measures} bit for bit in the cases we compared \mbox{(ReBAL:} all windows of four seeds of JSHC and ESCP-HC; MOLe: two windows of Hopper). At each predicted step $h$, the share of a model's error that its own updates remove is $s_h=1-\mathrm{nMSE}_h/\mathrm{nMSE}^{\mathrm{off}}_h$, where $\mathrm{nMSE}^{\mathrm{off}}_h$ is the error with the updates switched off, so the share refers to the error without the updates, as in Fig.~5f of the main paper; each window's nMSE is averaged over the seeds and then over the windows. The five datasets are averaged with equal weight, and intervals resample the changes within~each~dataset.

Twenty to thirty steps after a change, the compared models' updates remove 47\% [45, 49] of their error at step 1 and 14\% [12, 17] at step 40, and \ewm{}'s remove 50\% [48, 53] and 33\% [30, 35] (difference at step 40, before rounding: 18 [15, 21] percentage points). With only the fast correction, \ewm{}'s updates remove 49\% and 11\%, with only the slow one 35\% and 32\%, and with the routing memory alone 30\% and 9\%. At step 40, \ewm{}'s updates remove more than the compared model's on every dataset: Panda 9.1 [4.6, 13.7], UR5e 26.0 [22.0, 29.9], JSHC 28.5 [22.3, 34.3], \mbox{ESCP-HC} 20.9 [12.1, 33.0] and Hopper 6.4 [0.3, 12.8] percentage points. The two compared models fall short in different ways. ReBAL (UR5e, JSHC, ESCP-HC) carries its correction in a recurrent state that is updated with its own predictions during the open-loop rollout, and its updates remove \mbox{38\%--69\%} of its error at step 1 but \mbox{1\%--9\%} at step 40. MOLe (Panda, Hopper) carries its correction in adapted parameters that stay fixed during the open-loop rollout. Its updates remove \mbox{32\%--34\%} at step 1 and \mbox{27\%--33\%} at step 40, less than \ewm{}'s updates on the same two datasets (\mbox{53\%--54\%} and \mbox{36\%--39\%}). Right after a change ($d=1$--4), \ewm{}'s updates also remove more at step 40, 22\% against 7\%. At step~1, ReBAL's updates then already remove \mbox{32\%--63\%} of its error, while MOLe's remove only 14\% [7, 20] on Panda and none on Hopper ($-7\%$ [$-29$, 14]), against \ewm{}'s 37\%~and~24\%.

A second measure leads to the same picture: the share of the change's own effect that each prediction keeps. All five datasets are simulated, so this effect can be computed for every window of \cref{sec:supp-E5-measures}. From the recorded state at the prediction start, we re-simulate the 40 predicted steps with the recorded actions, forces and per-step dynamics parameters, and once more with the dynamics parameters from before the change, keeping every transient pulse and impact as recorded. With the recorded parameters, the re-simulation reproduces the recorded future to within $1.8\times10^{-6}$ on Panda and UR5e (in standardized coordinates, largest error over all windows and steps) and exactly on JSHC, ESCP-HC and Hopper, whose episodes we replay from their start with the generators' own code. The difference $\boldsymbol{E}_h=\vz_h-\vz^{\mathrm{cf}}_h$ between the recorded future $\vz_h$ and the counterfactual future $\vz^{\mathrm{cf}}_h$ at predicted step $h$ is the change's effect on the state, and the share of it that a prediction $\hvz_h$ keeps is
\begin{equation}
\kappa_h=\frac{\sum_w\langle \hvz_h-\vz^{\mathrm{cf}}_h,\,\boldsymbol{E}_h\rangle}{\sum_w\lVert \boldsymbol{E}_h\rVert^2},
\label{eq:supp-kept}
\end{equation}
summed over the windows $w$, with each model's predictions averaged over its seeds, in standardized coordinates. A prediction with $\kappa_h=1$ contains the whole effect, and one with $\kappa_h=0$ contains none of it, as a prediction that follows the dynamics from before the change does. A model that does not adapt is expected to lie in between, since it is trained on all dynamics, so what matters is the difference between models. We use the windows 20 to 30 steps after each change, when the online memories have turned around (\cref{sec:supp-E5-turn}). Intervals resample the changes within~each~dataset.

On Panda and UR5e (576 changes), the strongest compared models' predictions keep 88\% of the change's effect at step 1 and 66\% at step 40, and \ewm{}'s keep 93\% and 73\% (differences before rounding: 4 [3, 5] and 7 [4, 10] percentage points). \ewm{} with only the fast correction keeps 91\% and 66\%, and with only the slow one 80\% and 73\%. The fast correction thus keeps the change in the early predicted steps and the slow correction in the late steps. On JSHC, \mbox{ESCP-HC} and Hopper (1,010 changes), the compared models keep 87\% and 57\%, \ewm{} 87\% and 65\% (before rounding, 7 [6, 9] percentage points more at step 40), \ewm{} with only the fast correction 86\% and 62\%, and with only the slow one 81\%~and~65\%.

\FloatBarrier
\FloatBarrier
\subsubsection{What each online memory removes, part by part}
\label{sec:supp-E5-parts}

\begin{figure*}[p]
\centering
\includegraphics[width=\textwidth,height=0.88\textheight,keepaspectratio]{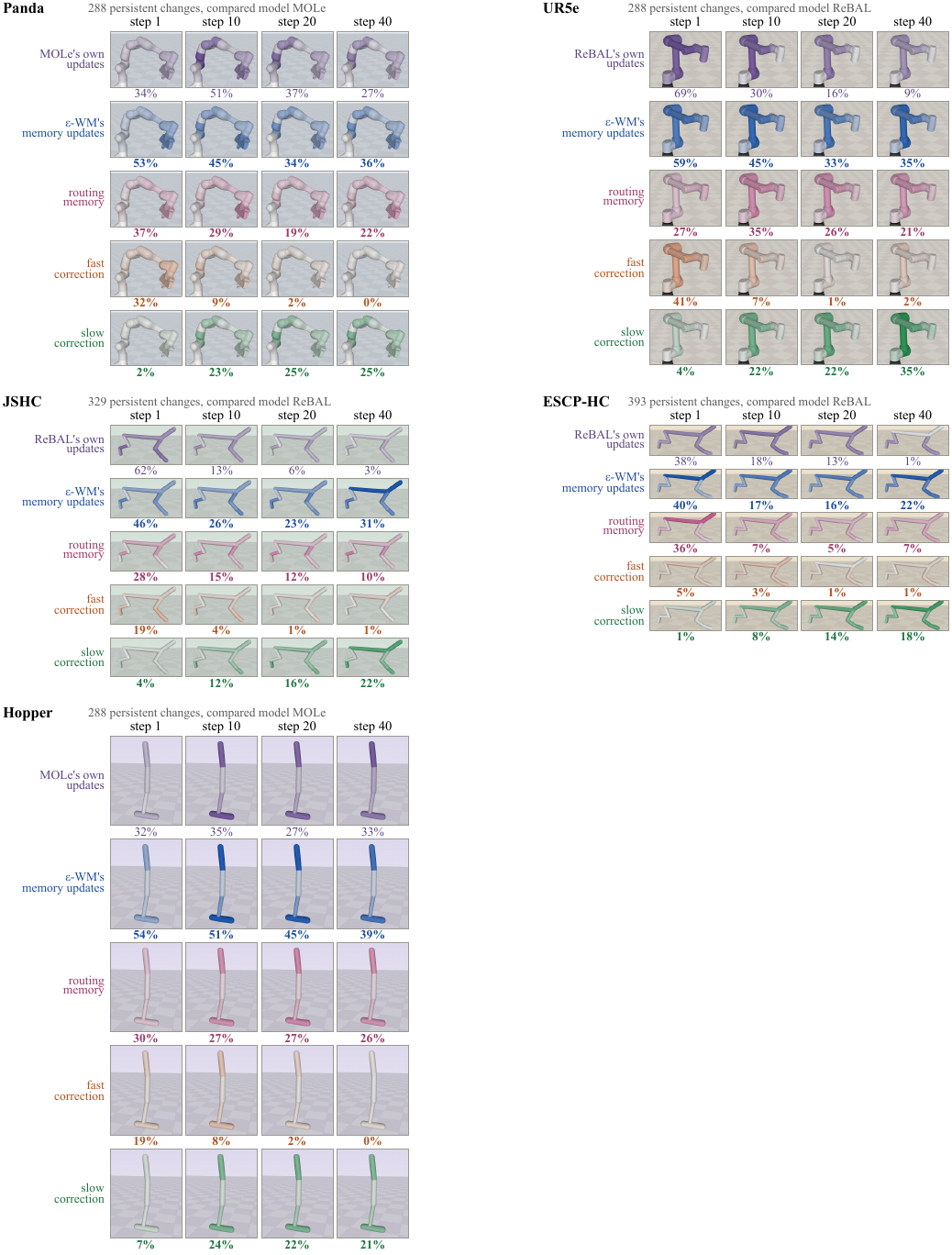}
\caption{What each mechanism removes on each body part, averaged over all persistent changes of the five simulated datasets (windows 20--30 steps after the change; 5 seeds, 4 on UR5e). Rows: the strongest compared model's own inference-time updates; \ewm{}'s memory updates together; its routing memory, fast correction and slow correction, each switched off at inference with the other two kept. Color per part: how much the mechanism lowers the part's error (darker = more, one scale per dataset; gray: no reduction). Number: share of the whole error removed at that step. The robot is drawn in one fixed pose.}
\label{fig:E5-memavg}
\end{figure*}

Figure~5b,\,c of the main paper show, in two examples, on which joints and at which predicted steps each of \ewm{}'s online memories and the compared model's own updates remove error. \Cref{fig:E5-memavg} shows the same averaged over all persistent changes. On the windows 20 to 30 steps after each change of \cref{sec:supp-E5-measures}, we switch off one mechanism at a time at inference: the compared model's own updates (as in \cref{sec:supp-E5-kept}), all of \ewm{}'s memory updates, or one of \ewm{}'s three online memories with the other two kept, as in Fig.~5b,\,c of the main paper. For each body part, the error is the nMSE on the part's state dimensions (a joint's angle and angular velocity; on the legged robots the torso holds the root coordinates), averaged over the seeds of each window and then over the windows. The color shows how much a mechanism lowers this error, and the number below each robot the share of the whole error that it removes. The robot is drawn in one fixed pose, as the colors are averages over all changes.

On all five datasets, the division of labor of the two examples holds on average. The fast correction removes 5\%--41\% of the error at step~1 and at most 2\% at step~40. The slow correction removes at most 7\% at step~1 but 18\%--35\% at step~40; on the arms, it removes most on the joints that carry the load: Panda joints~2 and~4 (42\% and 45\% of their step-40 error) and UR5e's shoulder-lift and elbow joints (53\% and 42\%). The routing memory removes 27\%--37\% at step~1 and still 7\%--26\% at step~40. Together, \ewm{}'s memory updates remove 22\%--39\% at step~40. ReBAL's own updates remove 38\%--69\% at step~1 but 1\%--9\% at step~40, while MOLe's still remove 27\% and 33\% at step~40 on Panda and Hopper, less than \ewm{}'s memory updates there (36\% and 39\%). Thus, averaged over all changes, the fast correction serves the early steps, the slow correction the last ones, and the routing memory every step.

\subsubsection{Online memories held in the open-loop rollout}
\label{sec:supp-E5-held}

\begin{figure*}[t]
\centering
\includegraphics[width=\textwidth]{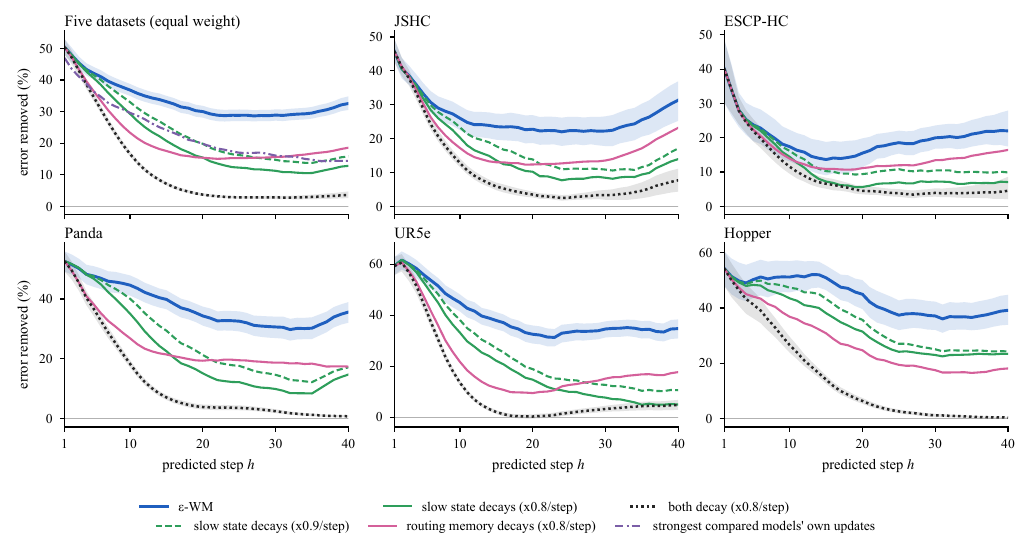}
\caption{Share of \ewm{}'s error removed by its memory updates at each predicted step, 20--30 steps after a persistent change, when the slow state or the routing memory decays during the open-loop rollout ($\times\eta$ per step), or both, next to the strongest compared models' own updates (top-left panel only, as in Fig.~5f of the main paper). \ewm{}: online memories held, as in Sec.~3.5 of the main paper. At step~1 all \ewm{} variants coincide. The slow state decaying by $\eta=0.95$ is given in the text only. Bands (for \ewm{} and for both decaying): 95\% intervals over the changes.}
\label{fig:E5-dose}
\end{figure*}

During the 40 predicted steps, \ewm{}'s fast state decays by a factor of 0.8 per step, while its routing memory and slow state stay fixed (Sec.~3.5 of the main paper). To test whether this is why the error removed by the routing memory and the slow state persists, we let these two online memories decay at inference on the windows 20 to 30 steps after each change of \cref{sec:supp-E5-measures} (intervals resample the changes). At predicted step $h$ ($h=1,\dots,40$), the slow state enters the prediction scaled by $\eta^{h-1}$, and the expert weights $\hvp$ move toward those \ewm{} would have without its routing memory, $\hvp_0$, as $\eta^{h-1}\hvp+(1-\eta^{h-1})\hvp_0$, for $\eta=0.95$, 0.9 and 0.8 (0.8 is the decay factor of the fast state). Nothing is retrained, and with $\eta=1$ the code reproduces \ewm{}'s predictions exactly. \Cref{fig:E5-dose} shows the share of \ewm{}'s error that its memory updates then remove at each step, as in Fig.~5f of~the~main~paper.

At step~1, all variants remove the same share of \ewm{}'s error, 50\% on average, as the decay starts only after it. At the later steps, \ewm{}'s updates remove less the faster its online memories decay (\cref{fig:E5-dose}). At step~40, these updates remove 33\% on average with the online memories held (\cref{sec:supp-E5-kept}), 21\%, 16\% and 13\% [10, 16] when the slow state decays by a factor of 0.95, 0.9 and 0.8 per step, 19\% [16, 21] when the routing memory decays by a factor of 0.8 per step, and 4\% [3, 5] when both do, less than the strongest compared models' own updates (14\%). The drop holds on every dataset: with both online memories decaying, \mbox{0.4\%--7.7\%} remain at step~40, against 22\%--39\% with the online memories held. Thus, \ewm{}'s updates still remove error at the last step because the routing memory and the slow state are held during the open-loop rollout. If the routing memory and the slow state decayed like the fast state, \ewm{} would lose most of its correction by the last step, as ReBAL does.

\clearpage %
\makeatletter\global\@colroom\@colht\global\vsize\@colht\makeatother %

\end{document}